\documentclass[sigconf]{acmart}
\AtBeginDocument{%
  \providecommand\BibTeX{{%
    \normalfont B\kern-0.5em{\scshape i\kern-0.25em b}\kern-0.8em\TeX}}}

\copyrightyear{2021}
\acmYear{2021}
\setcopyright{rightsretained}
\acmConference[ACM CHIL '21]{ACM Conference on Health, Inference,
and Learning}{April 8--10, 2021}{Virtual Event, USA}
\acmBooktitle{ACM Conference on Health, Inference, and Learning (ACM
CHIL '21), April 8--10, 2021, Virtual Event,
USA}\acmDOI{10.1145/3450439.3451860}
\acmISBN{978-1-4503-8359-2/21/04}

\usepackage{enumitem}
\usepackage[utf8]{inputenc} 
\usepackage[T1]{fontenc}    
\usepackage{booktabs}       
\usepackage{amsfonts}       
\usepackage{nicefrac}       
\usepackage{microtype}      
\usepackage{commath}
\usepackage{longtable}
\usepackage[graphicx]{realboxes}
\usepackage{placeins}
\usepackage{textcomp}
\usepackage{dcolumn}
\usepackage{mathtools}
\usepackage{multirow}
\usepackage{algorithm}
\usepackage{stmaryrd}
\usepackage{algorithmic}
\usepackage{afterpage}
\usepackage{float}
\usepackage[load-configurations=version-1]{siunitx} 
\usepackage{hyperref}

\newsavebox\CBox
\def\textBF#1{\sbox\CBox{#1}\resizebox{\wd\CBox}{\ht\CBox}{\textbf{#1}}}

\definecolor{lightblue}{rgb}{0.01, 0.6, 1.0}

\begin{document}

\title[TPC Networks for LoS Prediction in the ICU]{Temporal Pointwise Convolutional Networks for Length of Stay Prediction in the Intensive Care Unit}

\author{Emma Rocheteau}
\orcid{1234-5678-9012}
\affiliation{%
  \institution{University of Cambridge}
  \city{Cambridge}
  \country{UK}
}
\email{ecr38@cam.ac.uk}

\author{Pietro Liò}
\affiliation{%
  \institution{University of Cambridge}
  \city{Cambridge}
  \country{UK}
}
\email{pl219@cam.ac.uk}

\author{Stephanie Hyland}
\affiliation{%
  \institution{Microsoft Research}
  \city{Cambridge}
  \country{UK}
}
\email{stephanie.hyland@microsoft.com}

\begin{abstract}
    The pressure of ever-increasing patient demand and budget restrictions make hospital bed management a daily challenge for clinical staff. Most critical is the efficient allocation of resource-heavy Intensive Care Unit (ICU) beds to the patients who need life support. Central to solving this problem is knowing for how long the current set of ICU patients are likely to stay in the unit. In this work, we propose a new deep learning model based on the combination of temporal convolution and pointwise (1x1) convolution, to solve the length of stay prediction task on the eICU and MIMIC-IV critical care datasets. The model -- which we refer to as Temporal Pointwise Convolution (TPC) -- is specifically designed to mitigate common challenges with Electronic Health Records, such as skewness, irregular sampling and missing data. In doing so, we have achieved significant performance benefits of 18-68\% (metric and dataset dependent) over the commonly used Long-Short Term Memory (LSTM) network, and the multi-head self-attention network known as the Transformer. By adding mortality prediction as a side-task, we can improve performance further still, resulting in a mean absolute deviation of 1.55 days (eICU) and 2.28 days (MIMIC-IV) on predicting remaining length of stay.
\end{abstract}

\ccsdesc[500]{Applied computing~Health informatics}
\ccsdesc[500]{Computing methodologies~Neural networks}
\ccsdesc[500]{Computing methodologies~Multi-task learning}
\ccsdesc[500]{Mathematics of computing~Time series analysis}

\keywords{Patient Outcome Prediction, Length of Stay, Mortality, Intensive Care Unit, Temporal Convolution}

\maketitle

\section{Introduction}

In-patient length of stay (LoS) explains approximately 85-90\% of inter-patient variation in hospital costs in the United States \citep{rapoport2003}. Extended length of stay is associated with increased risk of contracting hospital acquired infections \citep{hassan} and mortality \citep{LAUPLAND2006954}. Hospital bed planning can help to mitigate these risks and improve patient experiences \citep{Blom2015ThePO}. This is particularly important in the intensive care unit (ICU), which has the highest operational costs in the hospital \citep{dahl2012} and a limited supply of specialist staff and resources. 

At present, discharge date estimates are done manually by clinicians, but these rapidly become out-of-date and can be unreliable (for example \citet{Mak2012PhysiciansAT} found that the average error made by clinicians was 3.82 days). Automated systems drawing on the electronic health record (EHR) have the potential to improve forecasting accuracy using state-of-the-art models that can be updated in light of new data. This has efficiency benefits in reducing the administrative burden on clinicians, and the improved accuracy may enable more sophisticated planning strategies e.g.\ scheduling high-risk elective surgeries on days with more availability \citep{gentimis2017}.

In our work, we simulate real-time predictions in retrospective data by updating the patients' remaining ICU length of stay prediction at hourly intervals during their stay using the preceding data from the EHR (similar to \citet{harutyunyan}). When designing both the architecture and pre-processing, we focus on mitigating the effects of non-random missingness due to irregular sampling, sparsity, outliers, skew, and other common biases in EHR data. Our key contributions are:
\begin{enumerate}
    \item A new model -- Temporal Pointwise Convolution (TPC) -- which combines:
    \begin{itemize}
        \item Temporal convolutional layers \citep{Simonyan2016, DBLP:journals/corr/KalchbrennerESO16}, which capture causal dependencies across the time domain.
        \item Pointwise convolutional layers \citep{lin2013network}, which compute higher level features from interactions in the feature domain.
    \end{itemize}
    Our model significantly outperforms the commonly used Long-Short Term Memory (LSTM) network \citep{10.1162/neco.1997.9.8.1735} and the Transformer \citep{46201} by margins of 18-68\%.
    \item We make a case for using the mean-squared logarithmic error (MSLE) loss function to train LoS models, as it deals more naturally with positively-skewed labels.
    \item By adding in-hospital mortality as a side-task, we demonstrate further performance gains in the multitask setting.
    \item We perform several investigations to improve our understanding of the model, including: an extensive ablation study of the model architecture, a post-hoc analysis of feature importances with integrated gradients~\citep{integratedgradients}, and a visualisation to show the model reliability as a function of the time since admission and the predicted remaining LoS.
\end{enumerate}

Additionally, we develop a data processing pipeline for the eICU \citep{Pollard2018} and MIMIC-IV \cite{johnson} databases that is designed to i) mitigate some of the impact of sparsity (for the diagnoses) and missing data (for time series) in the EHR and ii) extract a wide variety of features semi-automatically such that the approach is generalisable to other EHR databases. Our code is available at: \url{https://github.com/EmmaRocheteau/TPC-LoS-prediction}.

\section{Related Work}
\begin{figure}
  \centering
  \includegraphics[width=0.35\textwidth]{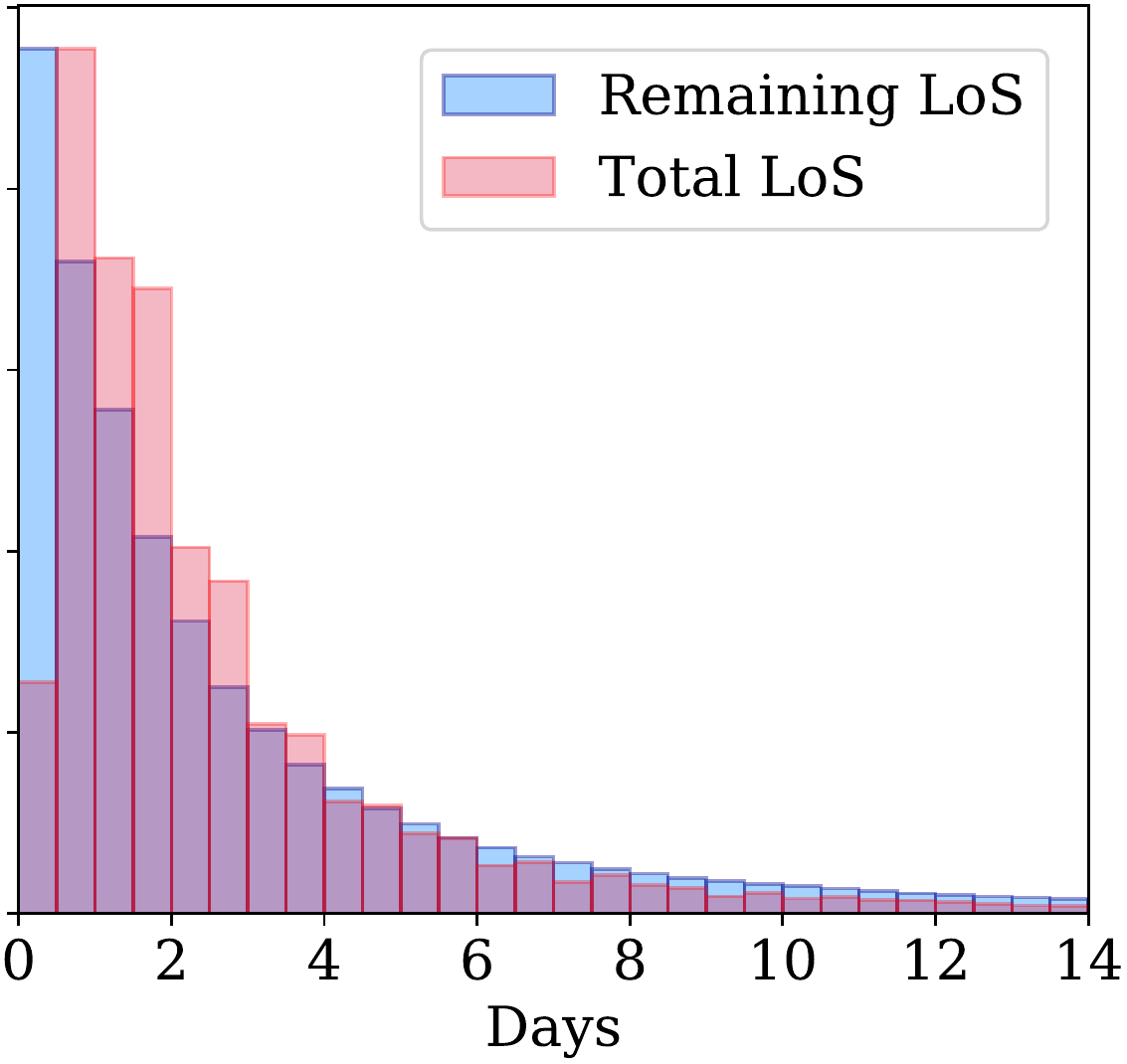}
  \caption{Total and remaining LoS distributions in the eICU dataset. The remaining LoS has a significant positive skew, with mean and median values of 3.47 and 1.67 days respectively. The skew in MIMIC-IV (not shown) is even more pronounced (5.70 and 2.70 days).}
\label{fig:los_dist}
\end{figure}

Despite its importance, LoS prediction has received less attention than mortality prediction. This could be due to its difficulty; LoS depends heavily on operational factors and there is considerable positive skew in its distribution (see Figure~\ref{fig:los_dist}). While it has been addressed as a regression problem (optimised using the mean-squared error (MSE) \citep{purushotham,sheikhalishahi2019benchmarking}), it is often simplified into binary classification (short vs. long stay) \citep{Gong2017PredictingCO,DBLP:journals/corr/abs-1811-12583,Rajkomar2018ScalableAA}, or as a multi-class task \citep{harutyunyan}. This simplification comes at a cost of utility, so we choose to focus on the more challenging regression variant.

Owing to the centrality of time series in the EHR, LSTMs have been by far the most popular model for predicting LoS \citep{harutyunyan,sheikhalishahi2019benchmarking,Rajkomar2018ScalableAA}. This reflects the prominence of LSTMs in other clinical prediction tasks such as predicting in-hospital adverse events including cardiac arrest \citep{Tonekaboni2018PredictionOC} and acute kidney injury \citep{Tomaev2019ACA}, forecasting diagnoses,  medications and interventions \citep{Choi2015DoctorAP,Lipton2015LearningTD,Suresh2017ClinicalIP}, missing-data imputation \citep{Cao2018BRITSBR}, and mortality prediction \citep{Che2018,harutyunyan,Shickel2019DeepSOFAAC}. More recently, the Transformer model \citep{46201} been shown to marginally outperform the LSTM on LoS \citep{2304ed73e858419398e3ee1508af5825} (and it continues to dominate in many other domains \citep{Mousavi2020}). Therefore, the LSTM and the Transformer were chosen as key baselines. 

Temporal convolution models have previously been applied to the task of early disease detection using longitudinal
lab tests \citep{pmlr-v85-oh18a, DBLP:journals/corr/RazavianS15, sontag}, yielding similar results to the LSTM. We highlight two main differences in our work: we introduce a set of pointwise convolutions in parallel, and the temporal convolution filters do not share their parameters between features, allowing the model to optimise processing in spite of heterogeneity in the temporal characteristics. We demonstrate via ablation studies how these design choices contribute substantial improvements to the patient state representation, yielding state-of-the-art results on LoS prediction.


\section{Methods}
\label{methods}
\subsection{Model Overview}
\label{sec:modeloverview}
We want our model to extract both temporal trends and inter-feature relationships in order to capture the patient's clinical state. Consider a patient who is experiencing slowly worsening respiratory symptoms but is otherwise stable. As this patient is unlikely to be weaned from their ventilator in the near future, a clinician might anticipate a long remaining LoS, but how do they come to this conclusion? Intuitively, one of the factors they are evaluating is the trajectory of the patient e.g.\ they may ask themselves ``Is the respiratory rate getting better or deteriorating?''. However, they can obtain a better indication of lung function by combining certain features e.g.\ the PaO$_2$/FiO$_2$ ratio, and then looking at how \emph{these} vary over time. A model should therefore be adept at extracting and combining both intra-feature temporal statistics and inter-feature relationships.

Formally, our task is to predict the remaining LoS at regular timepoints $y_1,\ldots,y_T\in\mathbb{R}_{>0}$ in the patient's ICU stay, up to the discharge time $T$, using the diagnoses ($\mathbf{d}\in\mathbb{R}^{D\times 1}$), static features ($\mathbf{s}\in\mathbb{R}^{S\times 1}$), and time series ($\mathbf{x}_1,\ldots, \mathbf{x}_T \in\mathbb{R}^{F\times 2}$). Initially, for every timepoint $t$, there are two `channels' per time series feature: $F$ feature \emph{values} ($\mathbf{x'}_t\in\mathbb{R}^{F\times 1}$), and their corresponding decay indicators ($\mathbf{x''}_t\in\mathbb{R}^{F\times 1}$). The decay indicators tell the model how recently the observation $\mathbf{x'}_t$ was recorded. They are described in detail in Section~\ref{data}. As we pass through the layers of our model, we repeatedly extract trends and inter-feature relationships using a novel combination of techniques.

\subsection{Temporal Convolution}
Temporal Convolution Networks (TCNs) \citep{Simonyan2016,DBLP:journals/corr/KalchbrennerESO16} are a subclass of convolutional neural networks \citep{Fukushima1980} that convolve over the time dimension. They operate on two key principles: the output is the same length as the input, and there can be no leakage of data from the future. We use stacked TCNs to extract \textit{temporal trends} in our data. Unlike most implementations including \cite{sontag}, we \textit{do not share weights across features} i.e.\ weight sharing is only across time (like in Xception \citep{DBLP:journals/corr/Chollet16a}). This is because our features differ sufficiently in their temporal characteristics to warrant specialised processing. 

We define the temporal convolution operation for the $i_\text{th}$ feature in the $n_\text{th}$ layer as \begin{equation}\label{equation1}
    (f^{n,i} \ast \mathbf{h}^{n,i})(t) = \sum_{j=1}^{k}f^{n,i}[j] \; \mathbf{h}_{t-d(j-1)}^{n,i}
\end{equation}
where $\mathbf{h}_{1:t}^{n,i}\in\mathbb{R}^{C^{n} \times t}$ represents the temporal input to layer $n$ up to timepoint $t$, which contains $C^{n}$ channels per feature\footnote{In the first layer, the input $\mathbf{h}_{1:t}^{n,i}$ is the original data $\mathbf{x}_{1:t}^{n,i}\in\mathbb{R}^{2 \times t}$, so $C^{1}=2$.}.
The convolutional filter $f^{n,i}:\{1,\ldots,k\}\to\mathbb{R}^{Y \times C^{n}}$ is a tensor of $Y\times C^{n} \times k$ parameters per feature. It maps $C^{n}$ input channels into $Y$ output channels while examining $k$ timesteps. The output is therefore $(f^{n,i} \ast \mathbf{h}^{n,i})(t)^{\top}\in\mathbb{R}^{1\times Y}$. 
The dilation factor, $d$, and kernel size, $k$, together determine the temporal receptive field or `timespan' of the filter: $d(k-1) + 1$ hours for a single layer. To ensure that the output is always length $T$, we add left-sided padding of size $d(k-1)$ before every temporal convolution (not shown in equation~\ref{equation1}). The $t-d(j-1)$ term ensures that we only look backwards in time. The receptive field can be increased by stacking multiple TCNs (as in Wavenet \citep{Simonyan2016} and ByteNet \citep{DBLP:journals/corr/KalchbrennerESO16}). We increment the dilation by 1 with each layer i.e.\ $d=n$.

We concatenate the temporal convolution outputs for each feature, $i$ as follows
\begin{equation}
    \underbrace{(f^{n} \ast \overbrace{\mathbf{h}^{n}}^{\mathclap{\text{Temp.\ In.}(1)}})}_{\mathclap{\text{Temp.\ Out.}(2)}}(t) = \bigparallel_{i=1}^{R^{n}}(f^{n,i} \ast \mathbf{h}^{n,i})(t)^{\top}
\end{equation}
We use $\bigparallel$ to denote concatenation i.e.\ $\bigparallel^{A}_{i=1}\mathbf{a}^{i} = \mathbf{a}^{1}\parallel\ldots\parallel\mathbf{a}^{A}$. In our case, the output dimensions are $R^{n} \times Y$, where $R^{n}$ is the number of temporal input features. Throughout this section we label terms with numbers $(1)$, $(2)$ etc.\ corresponding to objects in Figure~\ref{fig:TPC}. We recommend following this alongside the equations.

\subsection{Pointwise Convolution}
\begin{figure}
  \centering
  \begin{minipage}{0.05\textwidth}
    \centering
    (a)
  \end{minipage}
  \begin{minipage}{0.39\textwidth}
      \includegraphics[width=1\textwidth]{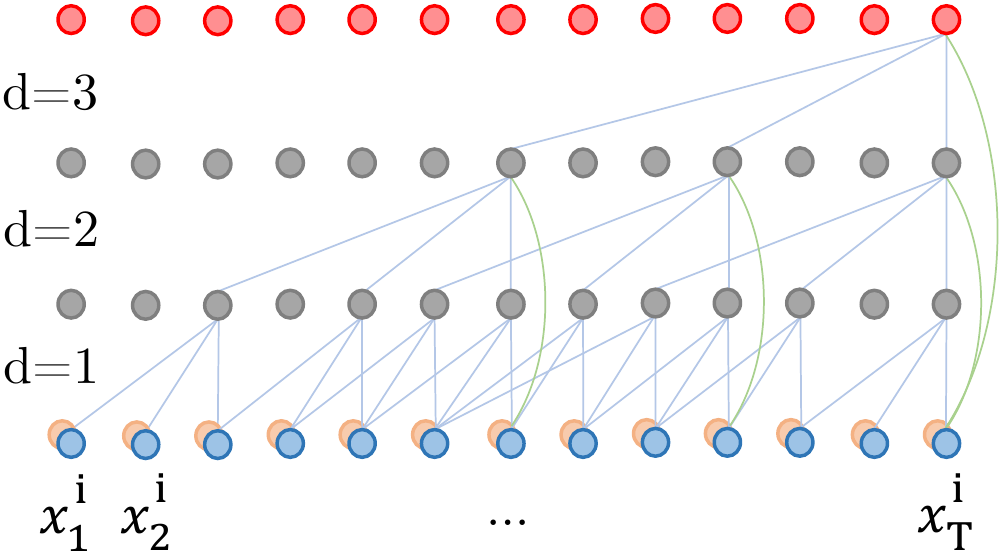}
      \vspace{0.1em}
  \end{minipage}
  \begin{minipage}{0.05\textwidth}
      \centering
    (b)
  \end{minipage}
  \begin{minipage}{0.39\textwidth}
  \includegraphics[width=1\textwidth]{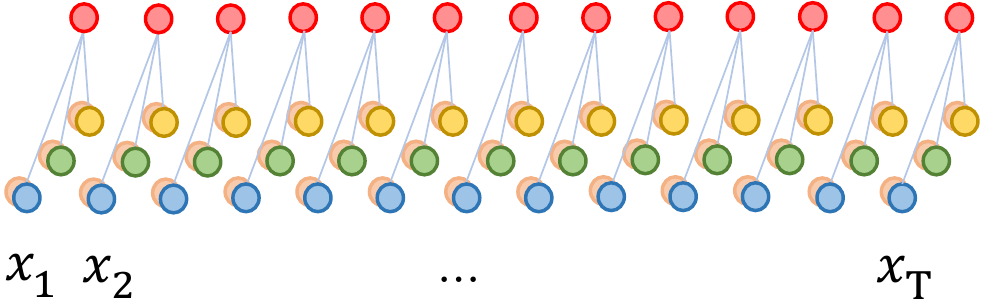}
  \end{minipage}
  \caption{(a) Temporal convolution with skip connections (green lines). Each time series, $i$ (blue dots) and their decay indicators (pale orange dots) are processed with independent parameters. (b) Pointwise convolution. There is no information sharing across time, only across features (blue, green, yellow dots).}
\label{fig:temp_pointwise}
\end{figure}

Pointwise convolution \citep{lin2013network}, also referred to as $1\times1$ convolution, is typically used to reduce the channel dimension when processing images \citep{DBLP:journals/corr/SzegedyLJSRAEVR14}. It can be conceptualised as a fully connected layer, applied separately to each timepoint (shown diagrammatically in Figure~\ref{fig:temp_pointwise}). As in temporal convolution, the weights are shared across all timepoints; however, there is no \textit{information transfer} across time. Instead, information is shared across the \textit{features} to obtain $Z$ interaction features\footnote{We use a wider set of features for pointwise convolution, including static features $\mathbf{s}$ and decay indicators $\mathbf{x''}$ i.e.\ $\mathbf{p}_t^{n} = (\flat(\mathbf{h}_t^{n}) \parallel \mathbf{s} \parallel \mathbf{x''}_t)\in\mathbb{R}^{P^{n}\times 1}$.}, $\mathbf{p}_t^{n} = (\flat(\mathbf{h}_t^{n}) \parallel \mathbf{s} \parallel \mathbf{x''}_t)\in\mathbb{R}^{P^{n}\times 1}$, where $P^{n} = (R^{n} \times C^{n}) + F + S$, and $\flat:A^{d_1 \times d_2 \ldots \times d_n}\to A^{(d_1 \cdot d_2 \ldots \cdot d_n)\times 1}$ is the flatten operation. We define the pointwise convolution operation in the $n_\text{th}$ layer as
\begin{equation}
    \underbrace{(g^{n} \ast \overbrace{\mathbf{p}^{n}}^{\mathclap{\text{Point.\ In.}(4)}})}_{\mathclap{\text{Point.\ Out.}(5)}}(t) = \sum_{i=1}^{P^{n}}g^{n}[i]p_t^{n,i}
\end{equation}
where $g^{n}:\{1,\ldots,P^{n}\}\to\mathbb{R}^{Z\times 1}$ is the pointwise filter, and the resulting convolution produces $Z$ output channels, so $(g^{n} \ast \mathbf{p}^{n})(t)\in\mathbb{R}^{Z\times 1}$.

\subsection{Skip Connections}
\label{skip}
\begin{figure}
\includegraphics[width=0.4782\textwidth]{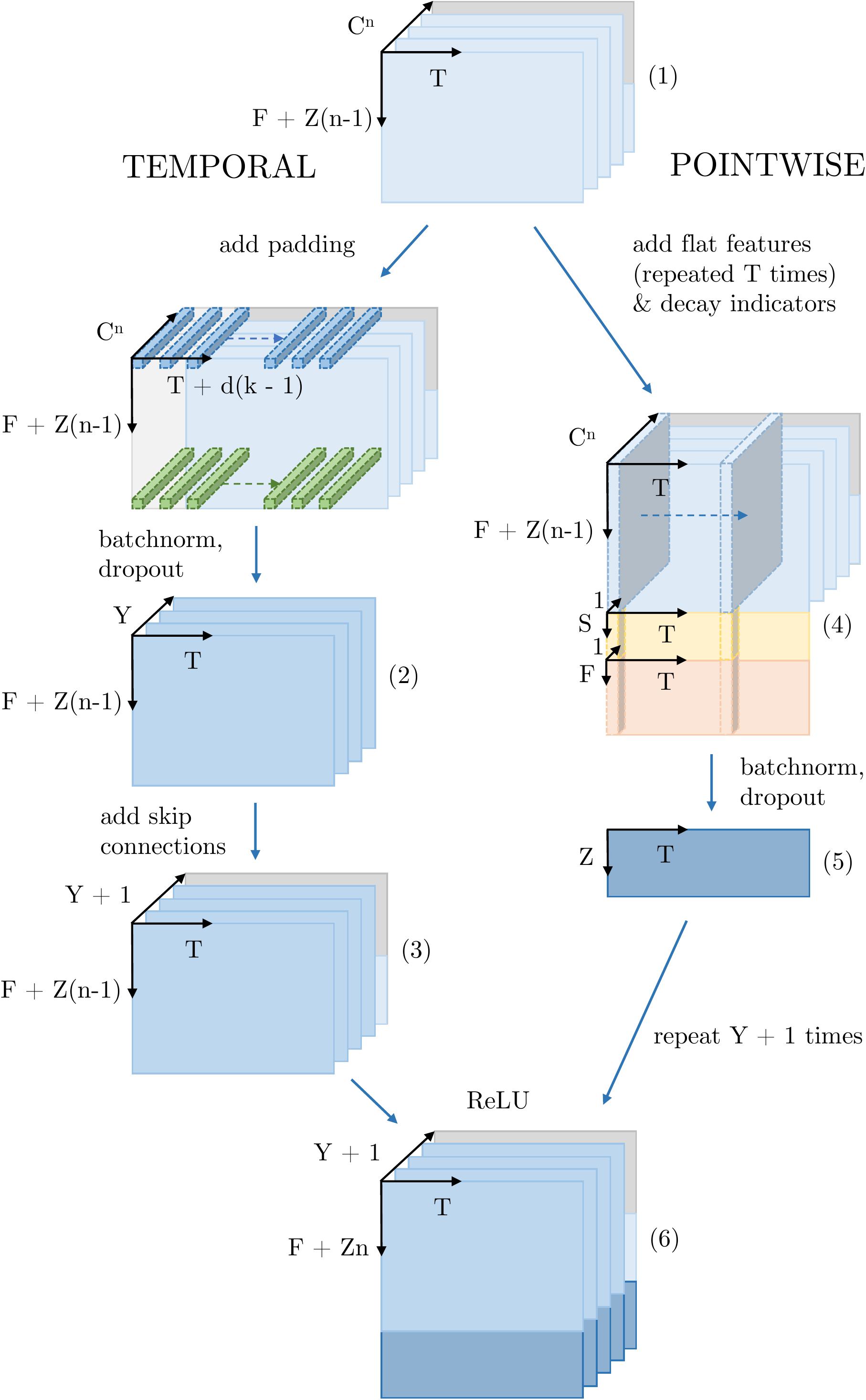}
\caption{The $n_\text{th}$ TPC layer. Left-sided padding (off-white) is added to the temporal side before each feature is processed independently. On the pointwise side, flat features (yellow) and decay indicators (orange) are added before each convolution.} 
\label{fig:TPC}
\end{figure}

We propagate skip connections \citep{DBLP:journals/corr/HeZRS15} to allow each layer to see the original data and the pointwise outputs from previous layers. This helps the network to cope with sparsely sampled data. For example, suppose a particular blood test is taken once per day. In order not to lose temporal resolution, we forward-fill these data (Section~\ref{data}) and convolve with increasingly dilated temporal filters until we find the appropriate width to capture a useful trend. However, if the smaller filters in previous layers (which did not see any useful trend) have polluted the original data by re-weighting, learning will be harder. Therefore, skip connections provide a consistent anchor to the input. They are concatenated (like in DenseNet \citep{densenet}, and are arranged in the shared-source connection formation \citep{wang20188}) as illustrated in Figure~\ref{fig:temp_pointwise}. The skip connections expand the feature dimension, $R^{n}=F+Z(n-1)$, to accommodate the pointwise outputs, and also the channel dimension to fit the original data, $C^{n}=Y+1$. This is best visualised in Figure~\ref{fig:TPC}.

\subsection{Temporal Pointwise Convolution}
Our model -- which we refer to as Temporal Pointwise Convolution (TPC) -- combines temporal and pointwise convolution in parallel. Firstly, the temporal output is combined with the skip connections to form $\mathbf{r}_t^{n}$ (Step 3 in Figure~\ref{fig:TPC}).
\begin{equation}
    \underbrace{\mathbf{r}_t^{n}}_{(3)} = \underbrace{(f^{n} \ast \mathbf{h}_t^{n})}_{\mathclap{\text{Temp.\ Out.}(2)}} \parallel \underbrace{\mathbf{x'}_t \parallel \Big[ \bigparallel_{n'=1}^{n-1} (g^{n'} \ast \mathbf{p}_t^{n'})}_{\mathclap{\text{Skip Connections}}}\Big]
\end{equation}
$\mathbf{r}_t^{n}$ is then concatenated with the pointwise output after it has been broadcast $Y+1$ times. We can therefore define the $n_\text{th}$ TPC layer as
\begin{equation}
    \underbrace{\mathbf{h}_t^{(n+1)}}_{\mathclap{\text{TPC Out.}(6)}} = \sigma\bigg(\underbrace{\mathbf{r}_t^{n}}_{(3)} \parallel \Big[ \bigparallel_{i=1}^{Y+1} \underbrace{(g^{n} \ast \mathbf{p}_t^{n})}_{\mathclap{\text{Point.\ Out.}(5)}}\Big]\bigg)
\end{equation}
where $\sigma$ represents the ReLU activation function. The full model has $N$ TPC layers stacked sequentially. After $N$ layers, the output $\mathbf{h}_t^{N}$ is combined with static features $\mathbf{s}\in\mathbb{R}^{S\times 1}$, and a diagnosis embedding $\mathbf{d^*}\in\mathbb{R}^{D^*\times 1}$. Two pointwise layers are then applied to obtain the final predictions (see Appendix~\ref{sec:fullmodel} for the full details). We use batch normalisation \citep{Ioffe2015} and dropout \citep{Srivastava2014} throughout to regularise the model. 

\subsection{Loss Function}
\begin{figure}
  \centering
  \includegraphics[scale=0.43]{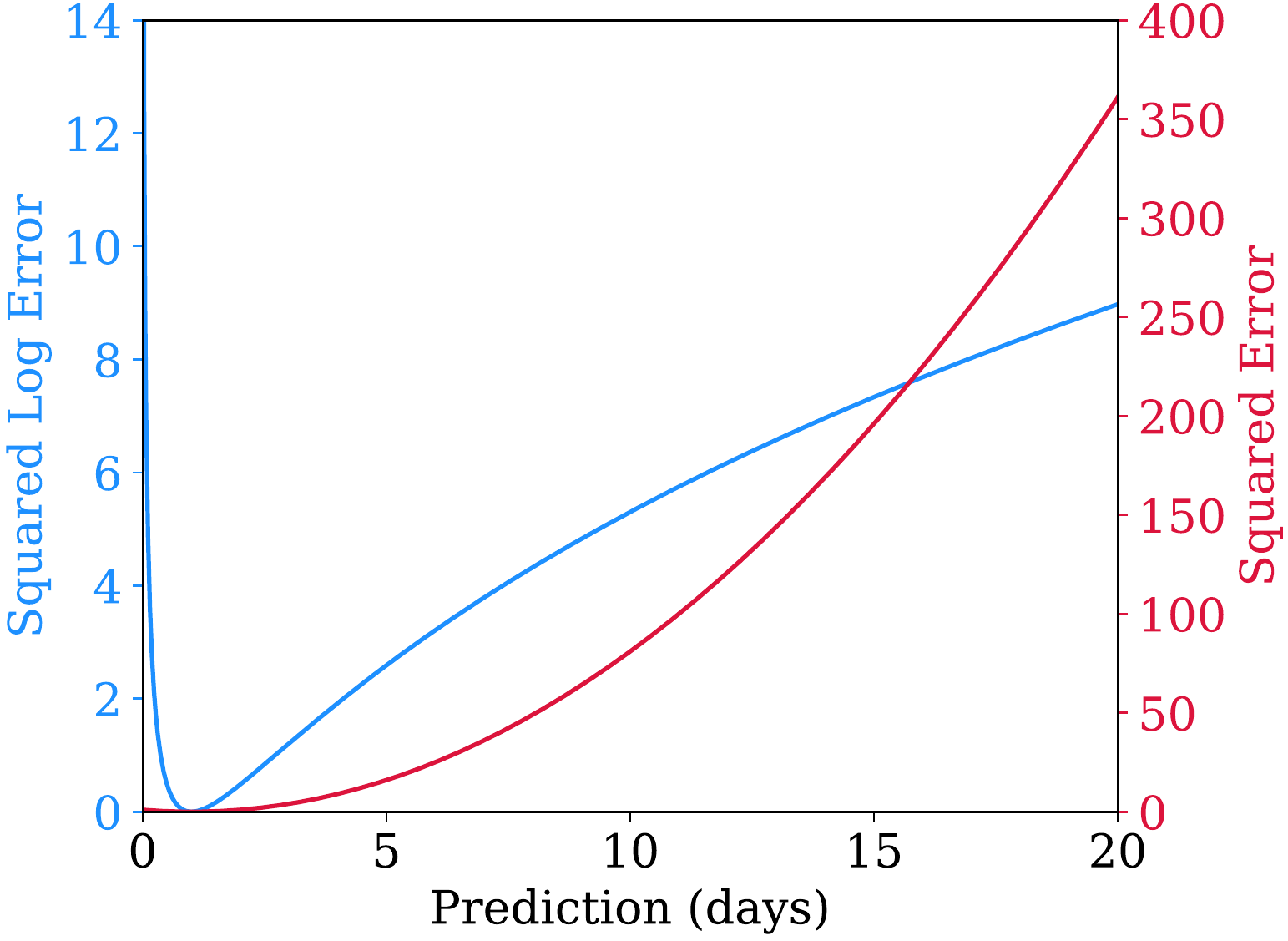}
  \caption{The behaviour of squared logarithmic error (blue) and squared error (red) functions when the true LoS is 1 day.}
 \label{fig:lossfunc}
\end{figure}

The remaining LoS has a positive skew (shown in Figure~\ref{fig:los_dist}) which makes the prediction task more challenging. We address this by replacing the commonly-used mean squared error (MSE) loss with mean squared \textit{log} error (MSLE). 
\begin{equation}
    \mathcal{L} = \frac{1}{T}\sum_{t=1}^T(\log(\hat{y}_t) - \log(y_t))^2
\end{equation}
MSLE penalises \textit{proportional} errors, which is more reasonable when considering an error of e.g. 5 days in the context of a 2-day stay vs.\ a 30-day stay. The difference can be seen in Figure~\ref{fig:lossfunc}. For bed management purposes it is particularly important not to harshly penalise over-predictions -- the model will become overly cautious and regress its predictions towards the mean. This is counter-productive because long stay patients have a disproportionate effect on bed occupancy.

\section{Data}
\label{data}
\subsection{eICU Database}
We use the eICU Collaborative Research Database \citep{Pollard2018}, a multi-centre dataset collated from 208 care centres in the United States, available through PhysioNet \cite{Goldbergere215}. It comprises 200,859 patient unit encounters for 139,367 unique patients admitted to ICUs between 2014 and 2015. 

\begin{figure}
  \centering
  \includegraphics[scale=0.41]{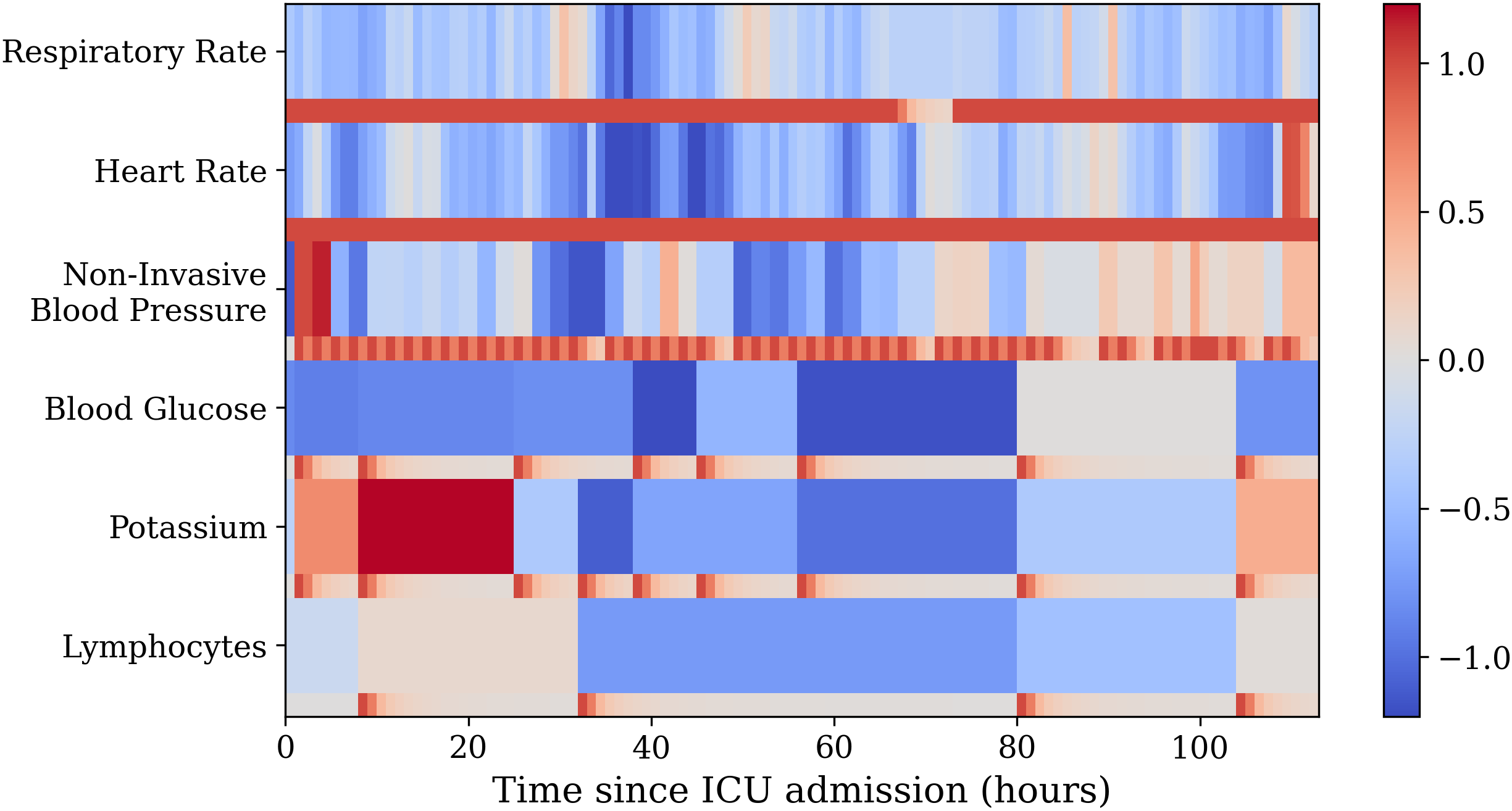}
  \caption{Example data from a patient in eICU (after pre-processing). The colour scale indicates the value of the feature, and the narrow bars show the corresponding decay indicators. Blood glucose, potassium and lymphocytes are from the \textit{lab} table and are sparsely sampled. Non-invasive blood pressure is manually recorded by the nurse every 2 hours, while respiratory rate and heart rate are vital signs that are automatically logged.}
  \label{fig:one_patient}
\end{figure}

We selected all adult patients (\textgreater 18 years) with an ICU LoS of at least 5 hours and at least one recorded observation, resulting in 118,535 unique patients and 146,671 ICU stays. We selected 87 time series from the following tables: \textit{lab}, \textit{nursecharting}, \textit{respiratorycharting}, \textit{vitalperiodic} and \textit{vitalaperiodic}. To be included, variables had to be present in at least 12.5\% of patient stays, or 25\% for \textit{lab} variables. As shown in Figure~\ref{fig:one_patient}, the \textit{lab} variables tend to be sparsely sampled. To help the model cope with this missing data, we forward-filled over the gaps. This is more realistic than interpolation as the clinician would only have the most recent value. We then added `decay indicators' to specify where the data is stale. The decay was calculated as $0.75^j$, where $j$ is the time since the last recording. This is similar in spirit to the masking used by \citet{Che2018}.

We extracted diagnoses from the \textit{pasthistory}, \textit{admissiondx} and \textit{diagnoses} tables, and 17 static features from the \textit{patient}, \textit{apachepatientresult} and \textit{hospital} tables (see Tables~\ref{tab:static} and \ref{tab:timeseries}, and Appendix~\ref{preproc} for the full list of features and further details). 

\subsection{MIMIC-IV Database}

We verify our results on a second dataset, the Medical Information Mart for Intensive Care (MIMIC-IV v0.4) database \cite{johnson}, a de-identified and publicly available EHR dataset from the Beth Israel Deaconess Medical Center containing 69,619 ICU stays from 50,048 patients admitted between 2008 and 2019.

We use the same cohort selection criteria as in eICU to select 69,609 ICU stays from 50,042 patients. We followed the same feature selection process to obtain a short list of 172 time series from the \textit{chartevents} and \textit{labevents}. We manually removed 71 of these from \textit{chartevents} because the variable did not vary over time, or because the distribution was not found to provide useful discrimination between patients (see Table~\ref{tab:timeseriesMIMIC} for the final list of features). We filled the missing data in the same way as in eICU. We extracted 12 flat features from the \textit{icustays}, \textit{admissions}, \textit{patients} and \textit{chartevents} tables (Table~\ref{tab:staticMIMIC}). We did not extract diagnoses from MIMIC-IV because they are not associated with reliable timestamps.

\begin{table}[h]
    \caption{Cohort summaries.}
    \label{tab:cohortsummary}
    \centering
    \begin{tabular}{lll}
        \toprule
        & \textbf{eICU} & \textbf{MIMIC-IV} \\
        \midrule
        Number of patients & 118,535 & 50,042 \\
        \hspace{2em}Train & 82,973 & 35,028 \\
        \hspace{2em}Validation & 17,781 & 7,507 \\
        \hspace{2em}Test & 17,781 & 7,507 \\
        Number of stays & 146,671 & 69,609 \\
        \hspace{2em}Train & 102,749 & 48,848 \\
        \hspace{2em}Validation & 22,033 & 10,497 \\
        \hspace{2em}Test & 21,889 & 10,264 \\\hline
        Gender (\% male) & 54.1\% & 55.8\% \\
        Age (mean) & 63.1 & 64.7 \\
        LoS (mean) & 3.01 & 3.98 \\
        LoS (median) & 1.82 & 2.06 \\
        Remaining LoS (mean) & 3.47 & 5.70 \\
        Remaining LoS (median) & 1.67 & 2.70 \\
        In-hospital mortality & 9.25\% & 11.4\% \\\hline
        Number of input features & 104 & 113 \\
        \hspace{2em}Time series & 87 & 101 \\
        \hspace{2em}Static & 17 & 12 \\
        \bottomrule
    \end{tabular}
\end{table}

\section{Experiments}

In this section, we describe the prediction tasks, baseline models and evaluation metrics. As in \citet{harutyunyan} the training and test data was fixed upfront -- the patients were divided such that 70\% were used for training, 15\% for validation, and 15\% for testing. 

\subsection{Prediction tasks}
\subsubsection{Remaining Length of Stay}
We assign a remaining LoS target to each hour of the stay, beginning at 5 hours and ending when the patient dies or is discharged. We train the models to make a prediction every hour of the stay. We only include the first 14 days of any patient's stay to protect against very long batches which would slow down training. This cut-off applies to <5\% of patient stays, but it does \textit{not} affect their maximum remaining LoS values.

\subsubsection{In-Hospital Mortality}
We also tested the performance of the models on mortality prediction. Unlike LoS, these labels remain static throughout the patient stay. We used the same training procedure as the LoS task i.e.\ one prediction each hour. However, to reflect the approach taken by \citet{purushotham2017benchmark} and \citet{harutyunyan}, we only report the mortality performance once per patient (at 24 hours into the stay). This means that the cohort represented in the mortality metrics in Table~\ref{tab:multitaskresults} is smaller (16,239 of 21,889 test stays in eICU and 8,320 of 10,264 test stays in MIMIC-IV).

\subsubsection{Multitask}
Previous work has found merit in a multitask approach to patient outcome prediction \citep{harutyunyan, sheikhalishahi2019benchmarking}. We investigated whether we would see a similar benefit in the TPC model. When combining the LoS and mortality losses, we applied a relative weighting to the mortality loss -- dictated by a parameter $\alpha$ (which was treated as a hyperparameter). Further information on the hyperparameter search and implementation details is in Appendix~\ref{hyperparamsearch}.

\subsection{Baselines}
\label{baselines}
We include the following baselines in our experiments:
\paragraph{`Mean' and `Median' models (LoS only)} 
These always predict 3.47 and 1.67 days respectively for eICU and 5.70 and 2.70 days for MIMIC-IV (these correspond to the mean and median of the training data). This is to benchmark the level of performance which is achievable `for free' just by predicting in a reasonable range, and to provide points of reference when setting performance expectations for each dataset. 
\paragraph{APACHE-IV values \citep{Zimmerman2006} (eICU only)}
These are generated by a risk assessment scoring model which is evaluated only once per patient at 24 hours. Therefore it cannot be compared directly, but we include it \textit{only as a point of reference} for a widely used clinical model. APACHE-IV is only present in the eICU dataset.
\paragraph{Standard LSTM}
Our standard LSTM is similar to \citet{harutyunyan}. 
\paragraph{Channel-wise LSTM (CW LSTM)}
Again similar to \citet{harutyunyan}, this consists of a set of independent LSTMs that process each feature separately before concatenation (note the similarity with the independent temporal convolutions in the TPC model). 

\begin{table*}
  \caption{Performance of the TPC model compared to baseline models. The loss function in all experiments is MSLE. For the first four metrics, lower is better. The error margins are 95\% confidence intervals (CIs) calculated over 10 runs. These are not present for the mean, median and APACHE-IV models because they are deterministic. The best results are highlighted in blue. If the result is statistically significant on a t-test then it is indicated with stars (*p<0.05, **p<0.001). MAD: mean absolute deviation; MAPE: mean absolute percentage error; MSE: mean squared error; MSLE: mean squared logarithmic error; R$^2$: coefficient of determination, Kappa: Cohen Kappa Score. $^\dagger$Note that the APACHE-IV results (only present in the eICU dataset) cannot be compared directly to the other models (explained in Section \ref{baselines}).}
  \label{tab:results}
  \centering
  \begin{tabular}{p{2cm}|p{2.7cm}|p{1.45cm}p{1.45cm}p{1.3cm}p{1.45cm}p{1.45cm}p{1.45cm}}
    \toprule
        \textbf{Data} & \textbf{Model} & \textbf{MAD} & \textbf{MAPE} & \textbf{MSE} & \textbf{MSLE} & \boldmath{$R^2$} & \textbf{Kappa} \\
    \midrule
        \multirow{7}{*}{eICU} & Mean & {3.21} & {395.7} & {29.5} & {2.87} & {0.00} & {0.00} \\
        & Median & {2.76} & {184.4} & {32.6} & {2.15} & \hspace{-0.32em}{-0.11} & {0.00} \\
        & APACHE-IV$^\dagger$ & {2.54} & {182.1} & {16.6}$^\dagger$ & {1.10} & \hspace{-0.32em}{-0.01} & {0.20} \\
        & LSTM & {2.39$\pm$0.00} & {118.2$\pm$1.1} & {26.9$\pm$0.1} & {1.47$\pm$0.01} & {0.09$\pm$0.00} & {0.28$\pm$0.00} \\
        & CW LSTM & {2.37$\pm$0.00} & {114.5$\pm$0.4} & {26.6$\pm$0.1} & {1.43$\pm$0.00} & {0.10$\pm$0.00} & {0.30$\pm$0.00} \\
        & Transformer & {2.36$\pm$0.00} & {114.1$\pm$0.6} & {26.7$\pm$0.1} & {1.43$\pm$0.00} & {0.09$\pm$0.00} & {0.30$\pm$0.00} \\
        & TPC & {\textBF{\textcolor{blue}{1.78$\pm$0.02}}}$^{\footnotesize{**}}$ & {\textBF{\textcolor{blue}{63.5$\pm$4.3}}}$^{\footnotesize{**}}$ & {\textBF{\textcolor{blue}{21.7$\pm$0.5}}}$^{\footnotesize{**}}$ & {\textBF{\textcolor{blue}{0.70$\pm$0.03}}}$^{\footnotesize{**}}$ & {\textBF{\textcolor{blue}{0.27$\pm$0.02}}}$^{\footnotesize{**}}$ & {\textBF{\textcolor{blue}{0.58$\pm$0.01}}}$^{\footnotesize{**}}$ \\
    \midrule
        \multirow{6}{*}{MIMIC-IV} & Mean & 5.24 & 474.9 & 77.7 & 2.80 & 0.00 & 0.00 \\
        & Median & 4.60 & 216.8 & 86.8 & 2.09 & \hspace{-0.32em}{-0.12} & 0.00 \\
        & LSTM & 3.68$\pm$0.02 & 107.2$\pm$3.1 & 65.7$\pm$0.7 & 1.26$\pm$0.01 & 0.15$\pm$0.01 & 0.43$\pm$0.01 \\
        & CW LSTM & 3.68$\pm$0.02 & 107.0$\pm$1.8 & 66.4$\pm$0.6 & 1.23$\pm$0.01 & 0.15$\pm$0.01 & 0.43$\pm$0.00 \\
        & Transformer & 3.62$\pm$0.02 & 113.8$\pm$1.8 & 63.4$\pm$0.5 & 1.21$\pm$0.01 & 0.18$\pm$0.01 & 0.45$\pm$0.00 \\
        & TPC & \textBF{\textcolor{blue}{2.39$\pm$0.03}}$^{\footnotesize{**}}$ & \textBF{\textcolor{blue}{47.6$\pm$1.4}}$^{\footnotesize{**}}$ & \textBF{\textcolor{blue}{46.3$\pm$1.3}}$^{\footnotesize{**}}$ & \textBF{\textcolor{blue}{0.39$\pm$0.02}}$^{\footnotesize{**}}$ & \textBF{\textcolor{blue}{0.40$\pm$0.02}}$^{\footnotesize{**}}$ & \textBF{\textcolor{blue}{0.78$\pm$0.01}}$^{\footnotesize{**}}$ \\
    \bottomrule
    \end{tabular}
\end{table*}

\paragraph{Transformer} 
This model takes advantage of multi-head self-attention. Like the TPC model, it is not constrained to progress one timestep at a time; however, unlike TPC, it is not able to scale its receptive fields or process features independently. 

\subsection{Evaluation Metrics}
\subsubsection{Length of Stay}We report on 6 LoS metrics: mean absolute deviation (MAD), mean absolute percentage error (MAPE), mean squared error (MSE), mean squared log error (MSLE), coefficient of determination ($R^2$) and Cohen Kappa Score. This is important because bad models can `cheat' particular metrics just by being close to the mean or median value (see Appendix~\ref{evaluationmetrics} for additional discussion on this).

\subsubsection{In-Hospital Mortality}In the mortality and multitask experiments we report the area under the receiver operating characteristic curve (AUROC) and the area under the precision recall curve (AUPRC).

\section{Results}
In this section, we analyse the model in several ways. Firstly, we report overall performance and compare against a set of baselines. Next, we examine the role of the loss function. Finally, we perform a set of ablation studies to find out which components of the model architecture contribute the most to its success.

\subsection{TPC Performance on Length of Stay}

\begin{table*}[hbt!]
  \caption{Ablation studies of the TPC model (performed on the eICU dataset). Unless otherwise specified, the loss function is MSLE. The first subtable compares the effect of the loss function on the TPC model (see Table~\ref{tab:mseresults} in the Appendix for the MSE results of LSTM, CW LSTM and Transformer). The second shows various TPC ablation studies. Results that are not significantly different from the best result are highlighted in light blue. The TPC (MSLE) result has been repeated in each subtable for ease of comparison. WS: weight sharing; "no skip": no skip connections; "no diag.": no diagnoses, "no decay": no decay indicators.}
  \label{tab:ablationresults}
  \centering
  \begin{tabular}{p{2.7cm}|p{1.45cm}p{1.45cm}p{1.3cm}p{1.45cm}p{1.45cm}p{1.45cm}}
    \toprule
        \textbf{Model} & \textbf{MAD} & \textbf{MAPE} & \textbf{MSE} & \textbf{MSLE} & \boldmath{$R^2$} & \textbf{Kappa} \\
    \midrule
        TPC (MSLE) & {\textBF{\textcolor{blue}{1.78$\pm$0.02}}}$^{\footnotesize{**}}$ & {\textBF{\textcolor{blue}{63.5$\pm$4.3}}}$^{\footnotesize{**}}$ & {\textBF{\textcolor{lightblue}{21.7$\pm$0.5}}} & {\textBF{\textcolor{blue}{0.70$\pm$0.03}}}$^{\footnotesize{**}}$ & {\textBF{\textcolor{blue}{0.27$\pm$0.02}}} & {\textBF{\textcolor{blue}{0.58$\pm$0.01}}}$^{\footnotesize{**}}$ \\
        TPC (MSE) & {2.21$\pm$0.02} & {154.3$\pm$10.1} & {\textBF{\textcolor{blue}{21.6$\pm$0.2}}} & {1.80$\pm$0.10} & {\textBF{\textcolor{blue}{0.27$\pm$0.01}}} & {0.47$\pm$0.01} \\
    \midrule
        TPC & {\textBF{\textcolor{lightblue}{1.78$\pm$0.02}}} & {\textBF{\textcolor{blue}{63.5$\pm$3.8}}}$^{\footnotesize{*}}$ & {\textBF{\textcolor{lightblue}{21.8$\pm$0.5}}} & {\textBF{\textcolor{blue}{0.71$\pm$0.03}}}$^{\footnotesize{*}}$ & {\textBF{\textcolor{lightblue}{0.26$\pm$0.02}}} & {\textBF{\textcolor{lightblue}{0.58$\pm$0.01}}} \\
        Point.\ only & {2.68$\pm$0.15} & {137.8$\pm$16.4} & {29.8$\pm$2.9} & {1.60$\pm$0.03} & \hspace{-0.334em}{-0.01$\pm$0.10}\hspace{-0.334em} & {0.38$\pm$0.01} \\
        Temp.\ only & {1.91$\pm$0.01} & {71.2$\pm$1.1} & {23.1$\pm$0.2} & {0.86$\pm$0.01} & {0.22$\pm$0.01} & {0.52$\pm$0.01} \\
        Temp.\ only (WS) & {2.34$\pm$0.01} & {116.0$\pm$1.2} & {26.5$\pm$0.2} & {1.40$\pm$0.01} & {0.10$\pm$0.01} & {0.31$\pm$0.00} \\
        TPC (no skip) & {1.93$\pm$0.01} & {73.9$\pm$1.9} & {23.0$\pm$0.2} & {0.89$\pm$0.01} & {0.22$\pm$0.01} & {0.51$\pm$0.01} \\
        TPC (no diag.) & {\textBF{\textcolor{blue}{1.77$\pm$0.02}}} & {65.6$\pm$4.1} & {\textBF{\textcolor{blue}{21.5$\pm$0.5}}} & {\textBF{\textcolor{blue}{0.71$\pm$0.03}}}$^{\footnotesize{*}}$ & {\textBF{\textcolor{blue}{0.27$\pm$0.02}}} & {\textBF{\textcolor{blue}{0.59$\pm$0.01}}} \\
        TPC (no decay) & {1.84$\pm$0.01} & {64.5$\pm$3.0} & {22.5$\pm$0.3} & {0.77$\pm$0.02} & {0.24$\pm$0.01} & {0.56$\pm$0.01} \\
        Point.\ (no decay) & {2.90$\pm$0.18} & {179.1$\pm$17.4} & {34.2$\pm$4.6} & {1.80$\pm$0.05} & \hspace{-0.334em}{-0.16$\pm$0.16}\hspace{-0.334em} & {0.33$\pm$0.00} \\
    \bottomrule
    \end{tabular}
\end{table*}

\begin{table*}[h]
  \caption{Performance of the TPC model in the multitask setting. We compare the performance of each model on individual tasks (mortality only on the first line; LoS only on the second) to the multitask setting (both LoS and mortality on the third line). The performance of the baseline models are reported in Tables~\ref{tab:eICUmultitaskresults} and \ref{tab:MIMICmultitaskresults}.}
  \label{tab:multitaskresults}
    \centering
    \begin{tabular}{p{1.6cm}|p{1.75cm}p{1.75cm}|p{1.4cm}p{1.4cm}p{1.25cm}p{1.4cm}p{1.4cm}p{1.4cm}}
    \toprule
        & \multicolumn{2}{l|}{\textbf{In-Hospital Mortality}} & \multicolumn{6}{c}{\textbf{Length of Stay}} \\
        \textbf{Data} & \textbf{AUROC} & \textbf{AUPRC} & \textbf{MAD} & \textbf{MAPE} & \textbf{MSE} & \textbf{MSLE} & \boldmath{$R^2$} & \textbf{Kappa} \\
    \midrule
        \multirow{3}{*}{eICU} & \textBF{\textcolor{lightblue}{0.864$\pm$0.001}} & 0.508$\pm$0.005 & -- & -- & -- & -- & -- & -- \\
        & -- & -- & 1.78$\pm$0.02 & 63.5$\pm$3.8 & 21.8$\pm$0.5 & 0.71$\pm$0.03 & 0.26$\pm$0.02 & 0.58$\pm$0.01 \\
        & \textBF{\textcolor{blue}{0.865$\pm$0.002}} & \textBF{\textcolor{blue}{0.523$\pm$0.006}}$^{\footnotesize{**}}$ & \textBF{\textcolor{blue}{1.55$\pm$0.01}}$^{\footnotesize{**}}$ & \textBF{\textcolor{blue}{46.4$\pm$2.6}}$^{\footnotesize{**}}$ & \textBF{\textcolor{blue}{18.7$\pm$0.2}}$^{\footnotesize{**}}$ & \textBF{\textcolor{blue}{0.40$\pm$0.02}}$^{\footnotesize{**}}$ & \textBF{\textcolor{blue}{0.37$\pm$0.01}}$^{\footnotesize{**}}$ & \textBF{\textcolor{blue}{0.70$\pm$0.00}}$^{\footnotesize{**}}$ \\
    \midrule
        \multirow{3}{*}{MIMIC-IV} & 0.905$\pm$0.001 & 0.691$\pm$0.006 & -- & -- & -- & -- & -- & -- \\
        & -- & -- & 2.39$\pm$0.03 & 47.6$\pm$1.4 & 46.3$\pm$1.3 & 0.39$\pm$0.02 & 0.40$\pm$0.02 & 0.78$\pm$0.01 \\
        & \textBF{\textcolor{blue}{0.918$\pm$0.002}}$^{\footnotesize{**}}$ & \textBF{\textcolor{blue}{0.713$\pm$0.007}}$^{\footnotesize{**}}$ & \textBF{\textcolor{blue}{2.28$\pm$0.07}}$^{\footnotesize{*}}$ & \textBF{\textcolor{blue}{32.4$\pm$1.2}}$^{\footnotesize{**}}$ & \textBF{\textcolor{blue}{42.0$\pm$1.2}}$^{\footnotesize{**}}$ & \textBF{\textcolor{blue}{0.19$\pm$0.00}}$^{\footnotesize{**}}$ & \textBF{\textcolor{blue}{0.46$\pm$0.02}}$^{\footnotesize{**}}$ & \textBF{\textcolor{blue}{0.85$\pm$0.00}}$^{\footnotesize{**}}$ \\
    \bottomrule
    \end{tabular}
\end{table*}

\label{tpcperf}
The TPC model outperforms all of the baseline models on every metric on both datasets (Table~\ref{tab:results}) -- particularly those that are more robust to skewness: MAPE, MSLE and Kappa. Discounting APACHE-IV, the best performing \textit{baseline} across both datasets is the Transformer (although the channel-wise LSTM (CW LSTM) is similar on eICU). This is consistent with \citet{harutyunyan} (for CW LSTM) and \citet{2304ed73e858419398e3ee1508af5825} (for Transformers), who found small improvements over standard LSTMs. 

\paragraph{Performance differences between eICU and MIMIC-IV} Although the pattern of results is remarkably similar between eICU and MIMIC-IV, there are notable differences in the magnitudes of the metrics. These differences can be attributed to their LoS distributions -- the positive skew is more severe in MIMIC-IV (Table~\ref{tab:cohortsummary}). This skew has a disproportionate impact on the \emph{absolute} error, which is captured in the MSE and MAD metrics. Interestingly, the Kappa score is higher in MIMIC-IV because the model can assign the longest stay patients to the >8 day bin, whereas eICU has more medium stay patients in the 3-8 day range which need to be precisely placed. The most comparable results are the MSLE and MAPE metrics, both of which penalise the \emph{proportional} error, making them more robust to shifts in the LoS distribution. 

\subsection{Ablation Studies}
To understand the impact of each design choice for the TPC model, we study performance under different ablations on the eICU dataset. The results of these ablations are reported in Table~\ref{tab:ablationresults}.

\subsubsection{MSLE Loss Function}
The first two rows of Table~\ref{tab:ablationresults} show that using the MSLE (rather than MSE) loss function leads to significant improvements in the TPC model, with large performance gains in MAD, MAPE, MSLE and Kappa, while conceding little in terms of MSE and $R^2$. The MSE results for the other models are in Appendix Table~\ref{tab:mseresults}; they show a similar pattern to the TPC model.

\subsubsection{Model Architecture}
The second subtable shows that the temporal-only model is superior to the pointwise-only model, but neither reaches the performance of the TPC model. The temporal-only model performs much better than its weight-sharing variant, which demonstrates the importance of having independent parameters per feature. Note that the temporal-only model with weight sharing is the most similar to the approach taken by \citet{sontag}, and the results are comparable to the LSTM which is consistent with the results presented in the paper. Removing the skip connections reduces performance by 5-25\%. Together the ablation studies demonstrate that the superior performance of the TPC model is the culmination of multiple design decisions.

\subsubsection{Data}
We also tested the models without the diagnoses or decay indicators. Perhaps surprisingly, we found that the exclusion of diagnoses does not seem to harm the model. This could be because the relevant diagnoses for predicting LoS e.g.\ Acute Respiratory Distress Syndrome (ARDS), are discernible from the time series alone e.g.\ PaO2, FiO2, PEEP etc. The decay indicators contribute a small (but statistically significant) benefit. Their contribution is more obvious in the pointwise-only model where all of the metrics see improvements of 5-23\%. This difference is expected since they might reveal some of the temporal structure to the pointwise model e.g.\ reveal links between up-to-date observations and patient deterioration. 

In Appendix~\ref{tsablation} we tested the models without the laboratory tests (which are infrequently sampled) and without the other time series (which tend to be regularly monitored). They indicate that the TPC model is able to exploit disparate EHR time series more successfully than the baselines. They also show that the advantage of the CW LSTM over the standard LSTM is only apparent when the model has to process different types of time series simultaneously.

\subsection{Mortality and Multitask Performance}
We investigated adding in-patient mortality as a side-task to improve LoS prediction. Table~\ref{tab:multitaskresults} shows the TPC performance both on \emph{single-task} mortality prediction, as well as the multi-task setting. We observe first that TPC achieves good performance on mortality alone. Comparing the impact on LoS forecasting in the multi-task setting, we see significant improvements on every metric. Multi-task performance for all baselines is reported in Tables~\ref{tab:eICUmultitaskresults} and \ref{tab:MIMICmultitaskresults} in the Appendix, where the multitask training confers a more modest benefit.

\section{Further Analyses}
In this section, we further explore the performance and behaviour of the TPC model for LoS prediction on the eICU dataset. We test its capacity to exploit smaller datasets, explore which features it uses, and provide a visualisation of the reliability of the model. Finally, we simulate the potential use of the model for bed planning.

\subsection{Training Data Size}
\begin{figure}[h]
  \centering
  \includegraphics[width=0.4\textwidth]{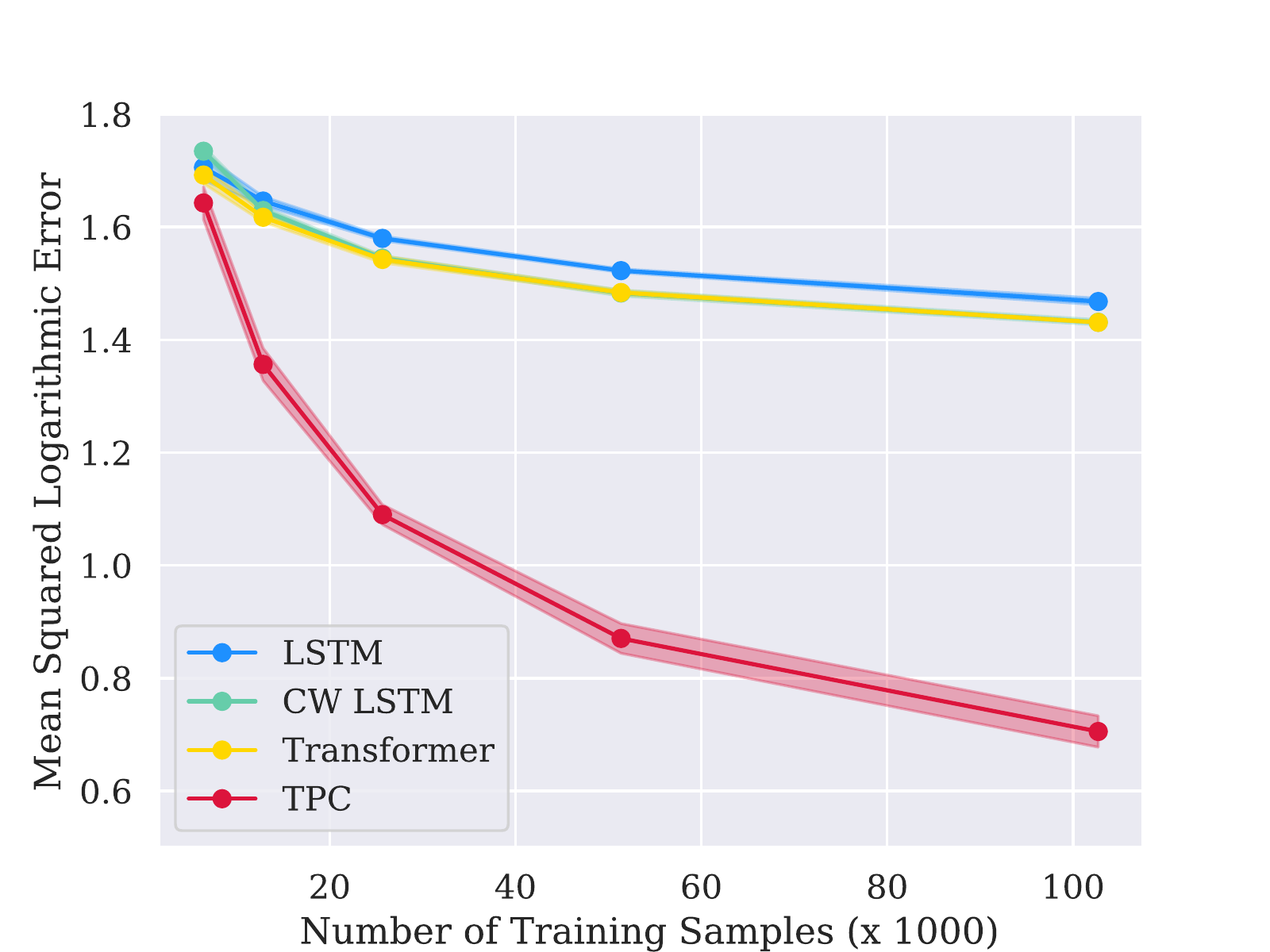}
  \caption{The effect of changing the training data size on the LSTM, CW LSTM, Transformer, and TPC model performance on the eICU dataset. Only the mean squared logarithmic error (MSLE) is shown for clarity, however the other metrics are shown in Table~\ref{tab:extraablationresults}. Note that the performance of the CW LSTM and Transformer models are so similar that the curves are superimposed.}
  \label{fig:datasize}
\end{figure}

The TPC model consistently outperforms the baselines when the training data is small, but we noticed even greater potential for big data. We tested the TPC, LSTM, CW LSTM, and Transformer models with 6.25\%, 12.5\%, 25\%, 50\%, and 100\% of the eICU training data. TPC maintains the best test performance on all data sizes, with an increasing benefit for larger data. Figure~\ref{fig:datasize} shows the effect on MSLE (the full results for all metrics are included in Table~\ref{tab:extraablationresults}).

\subsection{Feature Importance}
\begin{figure}[h]
  \centering
  \hspace{-0.7em}
  \includegraphics[width=0.47\textwidth]{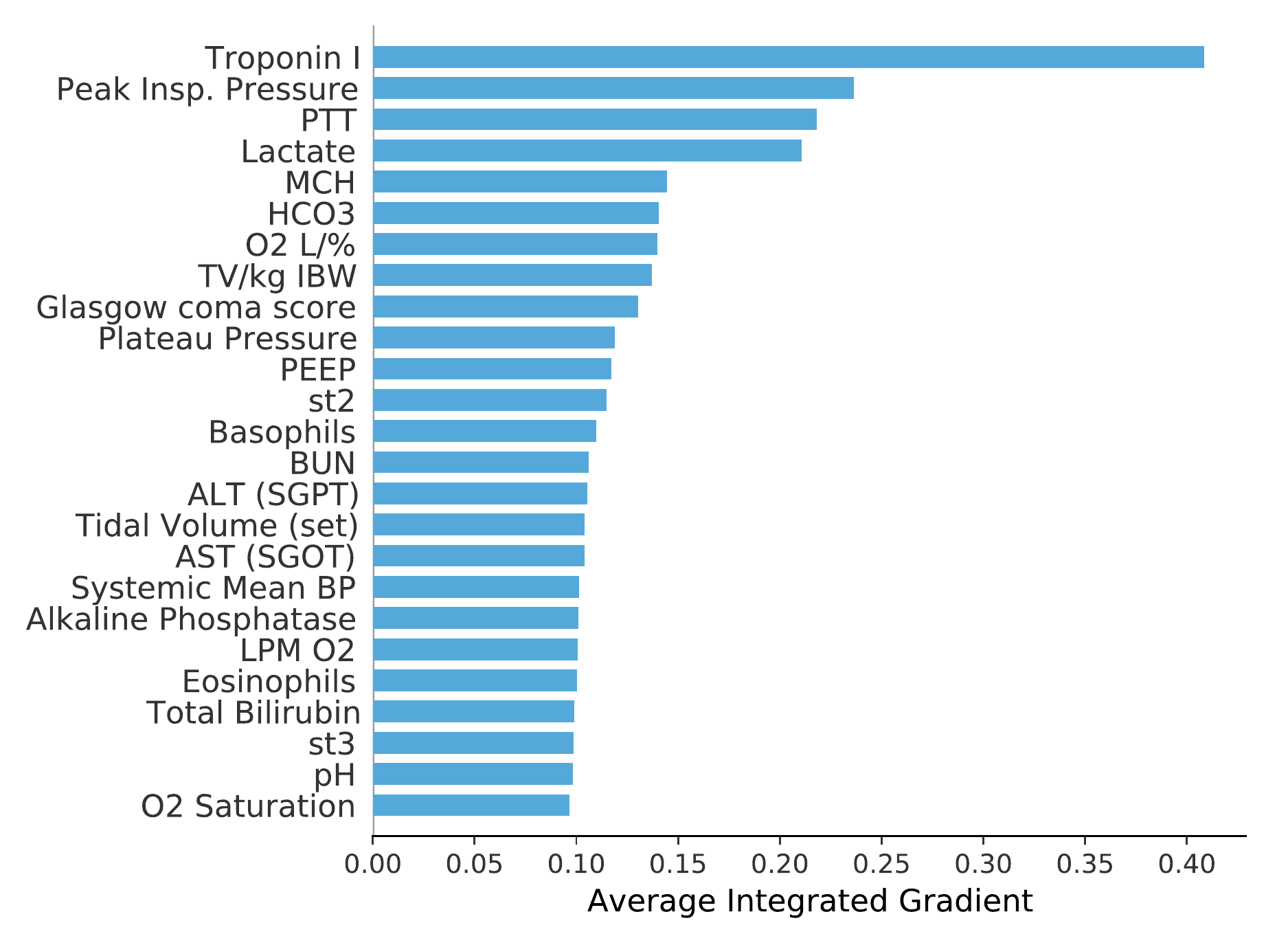}
  \caption{Top 25 most important features to the TPC model in the eICU dataset.}
    \label{fig:featureattr}
\end{figure}

We used the integrated gradients method~\citep{integratedgradients} to calculate feature attributions for the LoS estimates in the eICU dataset. This method computes the importance scores $\phi_i^{IG}$ by accumulating gradients interpolated between a baseline input $\textbf{b}$ (intended to represent the absence of data) and the current input $\textbf{x}$:
\begin{equation}
    \phi_i^{IG}(\psi, \textbf{x}, \textbf{b}) = \overbrace{(\textbf{x}_i - \textbf{b}_i)}^{\text{diff. from baseline}} \times \int_{\alpha=0}^{1}\overbrace{\frac{\delta \psi (\textbf{b} + \alpha(\textbf{x} - \textbf{b}))}{\delta \textbf{x}_i}}^{\text{acc. local grad.}}d \alpha
\end{equation}
where the TPC model is represented as $\psi$. We use the mean feature values as our baseline input vector. We take the absolute attribution values when a single LoS prediction is made for each patient at 24 hours. We aggregate by taking the mean along the time dimension and then the patient dimension to obtain Figure~\ref{fig:featureattr}. The background and intuition behind the method is explained clearly by \citet{sturmfels2020visualizing}. 

\label{sec:featimportance}

Analysing Figure~\ref{fig:featureattr}, we note that the top features are all strong indicators of organ failure: troponin I is a specific biomarker of myocardial infarction; peak inspiratory pressure, O2 L/\%, TV/kg IBW, plateau pressure, PEEP and tidal volume indicate mechanical ventilation (on account of respiratory failure); PTT, ALT (SGPT), AST (SGOT) and alkaline phosphatase suggest liver disease; and high BUN and bilirubin levels point towards kidney failure. Additionally we see infection markers such as lactate, basophils and eosinophils which could indicate sepsis. Both multi-organ failure and sepsis are known causes of extended LoS in the ICU~\citep{Bohmer2014}.

\subsection{Evaluation by Use-Case}
We have reported aggregate performance metrics indicating strong performance of the TPC model for overall LoS forecasting. In this section, we provide further evaluations tailored to two potential users -- an individual ICU clinician, and a bed manager for the unit.

\subsubsection{Individual-level Reliability}

Although aggregate measures of performance are typically reported, these can mask underlying variability in model performance. Such variability can undermine trust or result in unsafe application of systems~\cite{Sendak2020}. In this section, we think of a clinician who wishes to interpret the prediction of the system for an \emph{individual} patient. We break down the aggregate performance metrics based on factors which will be readily-available at the time of the prediction. Specifically, we visualise the MAPE (chosen for its interpretability) as a function of the time since admission and the \emph{predicted} remaining LoS.

\begin{figure}[h]
  \centering
  \includegraphics[width=0.41\textwidth]{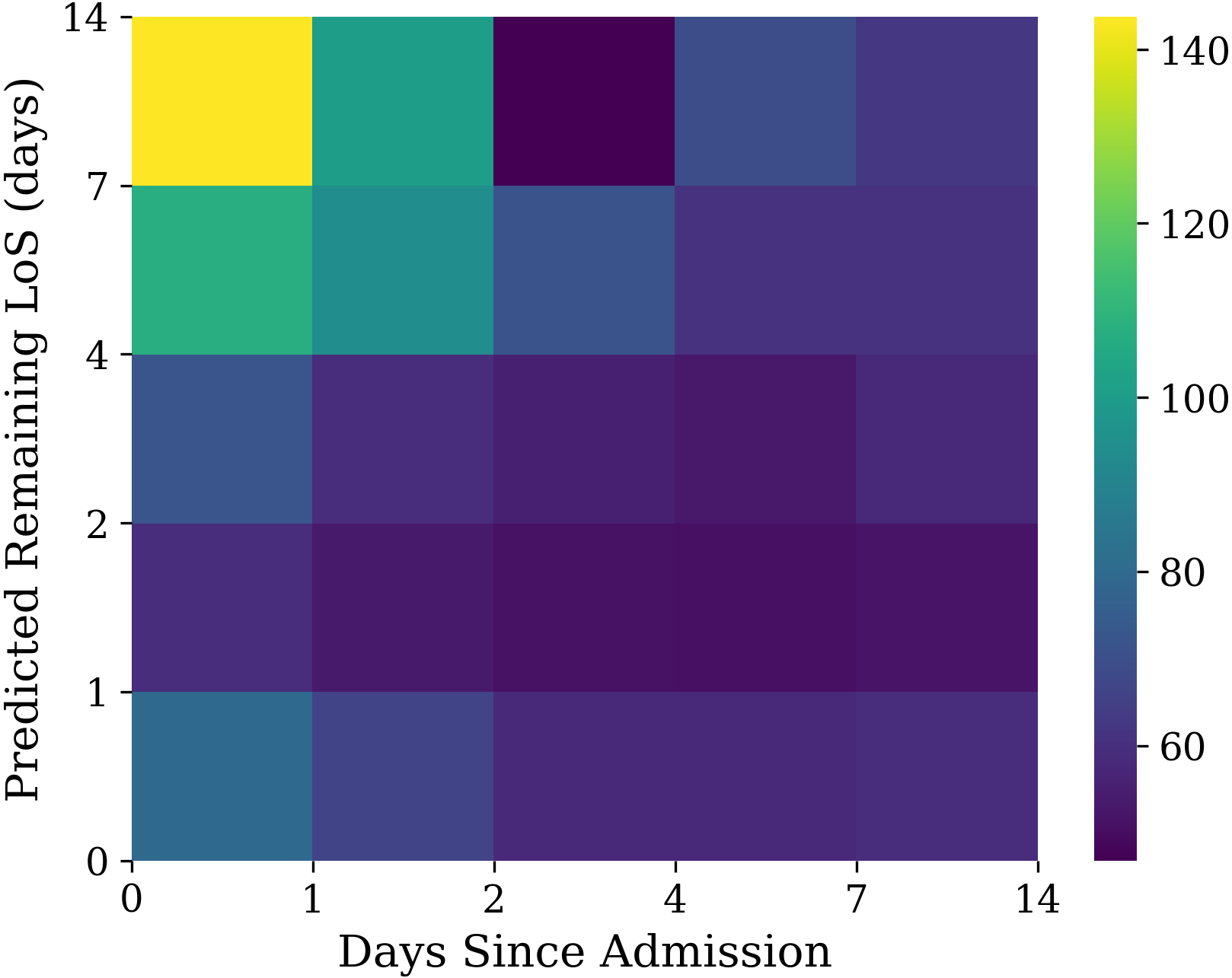}
  \caption{Mean absolute percentage error as a function of days since admission and predicted remaining LoS on the eICU dataset.}
    \label{fig:reliability}
\end{figure}

Figure~\ref{fig:reliability} shows an example for the TPC model on eICU. We can see that high predicted remaining LoS on the \emph{first} day of a patient's stay can be quite unreliable, with performance rapidly improving over time. Additional investigation revealed these initial predictions to be \emph{under}-predictions, indicating that it is challenging to accurately forecast \emph{very long} LoS for patients on their first day. The long tail of LoS in the dataset reflects the abundance of short-stay patients. The model therefore seems to wait for 1-2 days of data to justify a long LoS prediction. The system can therefore be equipped with instructions indicating that a high predicted remaining LoS on the \emph{first} day should not be acted upon. This could complement information provided on a model card~\cite{Mitchell2019ModelCF,Sendak2020}.

\subsubsection{ICU-level Bed Management}
\begin{figure}[h]
  \centering
  \includegraphics[width=0.45\textwidth]{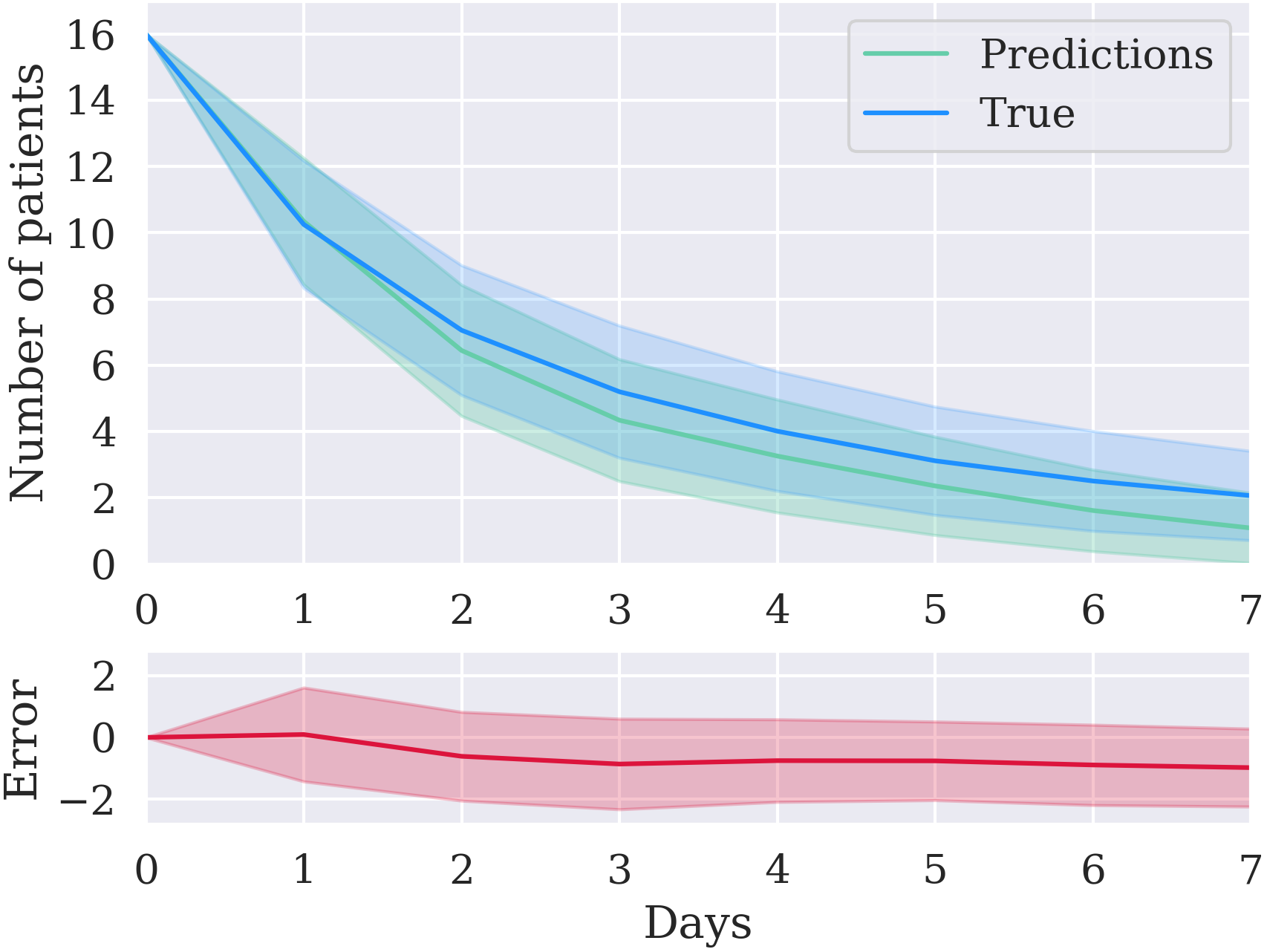}
  \caption{ICU simulation. We show the number of patients remaining in the ICU over time from an initial cohort of 16 random eICU patients from 500 simulations. The shaded regions show the standard deviation across the runs. `Error' is calculated from `True' minus `Predictions'.}
\label{fig:icusim}
\end{figure}

From the perspective of a bed manager, \emph{aggregate} performance of the model is important: an over-prediction for one patient could be offset by an under-prediction for another, resulting in the same net bed availability. To investigate this, we performed a simulation study. We ran 500 ICU simulations by randomly selecting 16 examples from the eICU test set to form a `virtual cohort'. The number 16 was chosen because US hospitals have, on average, 24 ICU beds \citep{doi:10.1164/rccm.201409-1746OC} with an occupancy rate of 68\%~\citep{Halpern2015}. Figure~\ref{fig:icusim} shows the number of patients remaining in the ICU (of the selected cohort; we do not visualise incoming ICU admissions) using their true remaining LoS (blue). We compute the error (red) between the predictions (green) and true values. The model is well calibrated when predicting patients who are going to stay for at least 1 day. After this, the model tends to under-predict the occupancy by approximately 0.8 patients, corresponding to a small bias towards under-estimating the remaining LoS. 

\section{Discussion}
\label{discussion}
We have shown that the TPC model outperforms all baseline models in all task settings (LoS, mortality or multitask) on both the eICU and MIMIC-IV datasets. To explain the success of TPC, we start by examining the parallel architectures in the TPC model. Each component has been designed to extract different information: trends from the temporal convolutions and inter-feature relationships from the pointwise convolutions. The eICU ablation studies reveal that the temporal element is more important, but we stress that their contributions are complementary since the best performance is achieved when they are used together.

Next, we highlight that the temporal-only model far outperforms its most direct comparison, the CW LSTM, on all metrics. Theoretically, they are well matched because they both have feature-specific parameters but are restricted from learning cross-feature interactions. To begin to explain this, we consider how the information flows through the model. The temporal-only model can directly step across large time gaps, whereas the CW LSTM is forced to progress one timestep at a time. This gives the CW LSTM the harder task of remembering information across a noisy EHR with distracting signals of varying frequency. In addition, the temporal-only model can tune its receptive fields for improved processing of each feature thanks to the skip connections (which are not present in the CW LSTM).


The difference in performance between the temporal-only model with and without weight sharing provides strong evidence that assigning independent parameters to each feature is important. Some EHR time series are irregularly and sparsely sampled, and can exhibit considerable variability in the temporal frequencies within the underlying data (evident in Figure~\ref{fig:one_patient}). This presents a challenge for any model, especially if it is constrained to learn one set of parameters to suit all features. The relative success of the CW LSTM over the standard LSTM when processing \textit{disparate} time series -- but not similar -- also lends weight to this theory.

However, the assignment of independent parameters to each feature does not explain all the successes of TPC e.g.\ the TPC model can process disparate time series and gain more marginal performance than the CW LSTM (Table~\ref{tab:additionalresults}). We need to consider that \textit{periodicity} is a key property of EHR data -- this is true in both the sampling patterns and in the underlying biology e.g.\ medication schedules, sleep cycles, meals etc. The temporal component of the TPC model is the only architecture with an inherent periodic structure (from the stacked temporal filters) which makes it much easier to learn EHR trends. By comparison, a single attention head in the Transformer model does not look at timepoints a fixed distance apart, but can take an arbitrary form. This is a strength for natural language processing, given the variety of sentence structures possible, but it does not help the Transformer to process EHRs.

Additionally, we have shown that the TPC model outperforms baselines on in-hospital mortality both as a standalone task and in combination with LoS. The performance on both mortality and LoS is significantly better in the multitask setting (this is consistent with past works \citep{harutyunyan, sheikhalishahi2019benchmarking}) because multitask learning helps to regularise the model and reduce the chance of overfitting \citep{multitaskoverfitting}. Adding further tasks may be a valid strategy to improve LoS performance.

Finally, we reiterate that using MSLE loss instead of MSE greatly mitigates for positive skew in the LoS task, and this benefit is not model-specific (all of the baselines perform better with MSLE -- see Table~\ref{tab:mseresults}). This demonstrates that careful consideration of the task -- as well as the data and model -- is an important step towards building useful tools in healthcare.

\subsection{Limitations and Future Work}

Our work has several limitations. We know that LoS is heavily influenced by operational factors, and clinical practices can change over time \citep{axel}. Capacity to maintain performance over time is an important consideration before a system could be used in practice. In future work, it would be instructive to test how quickly the models become out-of-date by reserving more recent data as a test set \citep{DBLP:journals/corr/abs-1811-12583}. 
Although we have included a large set of baselines, we acknowledge that a more exhaustive comparison could be performed, for example comparing by Gaussian Processes~\citep{DBLP:journals/corr/PrasadCCDE17} or ODE-RNNs~\citep{Rubanova2019LatentOF,Brouwer2019GRUODEBayesCM} for handling irregularly sampled time-series.
Finally, although we have motivated our study by bed management, this work describes a methodological proof of concept and does not constitute a real clinical system. Prospective study and integration into a real-world EHR is necessary to demonstrate real-world benefit, both of which pose their own challenges \citep{Rajkomar2018ScalableAA, sepsiswatch}.



In future work, we would like to investigate why the TPC model gains more from the multitask setting than the other models. It seems likely that it is related to additional regularisation provided by the mortality task, but further investigation is needed to confirm our speculations.

\section{Conclusion}
We have proposed and evaluated a new deep learning architecture, which we call `Temporal Pointwise Convolution' (TPC). TPC combines temporal convolutional layers with pointwise convolutions to extract temporal and inter-feature information. We have shown that the TPC model is well-equipped to analyse EHR time series containing missingness, differing frequencies and sparse sampling. We believe that the following four aspects contribute the most to its success:
\begin{enumerate}
    \item The combination of two complementary architectures that are able to extract different features, both of which are important.
    \item The ability to step over large time gaps.
    \item The capacity to specialise processing to each feature (including the freedom to select the receptive field size for each).
    \item The rigid spacing of the temporal filters, making it easy to derive trends.
\end{enumerate}
From a clinical perspective, we have contributed to the advancement of LoS prediction models, a prerequisite for automated bed management tools. Improving the practice of bed management promises cost reduction \citep{Halpern2015} and better resource allocation \citep{Mathews2015} worldwide. From a computational perspective, we have provided key insights for retrospective EHR studies, particularly where LSTMs are the currently model of choice. In the broader context of machine learning for healthcare we have demonstrated that careful consideration of the complexities of health data is necessary to gain state-of-the-art performance in these tasks.

\section*{Acknowledgements}
The authors would like to thank Alex Campbell, Petar Veli\v{c}kovi\'{c}, and Ari Ercole for helpful discussions and advice. We would also like to thank Louis-Pascal Xhonneux, Seyon Sivarajah, Rudolf Cardinal, Jacob Deasy, Paul Scherer, and Katharina Kohler for their help in reviewing the manuscript. Finally we thank the Armstrong Fund, the Frank Edward Elmore Fund, and the School of Clinical Medicine at the University of Cambridge for their generous funding.

\bibliographystyle{ACM-Reference-Format}
\bibliography{references}


\begin{thebibliography}{67}


\ifx \showCODEN    \undefined \def \showCODEN     #1{\unskip}     \fi
\ifx \showDOI      \undefined \def \showDOI       #1{#1}\fi
\ifx \showISBNx    \undefined \def \showISBNx     #1{\unskip}     \fi
\ifx \showISBNxiii \undefined \def \showISBNxiii  #1{\unskip}     \fi
\ifx \showISSN     \undefined \def \showISSN      #1{\unskip}     \fi
\ifx \showLCCN     \undefined \def \showLCCN      #1{\unskip}     \fi
\ifx \shownote     \undefined \def \shownote      #1{#1}          \fi
\ifx \showarticletitle \undefined \def \showarticletitle #1{#1}   \fi
\ifx \showURL      \undefined \def \showURL       {\relax}        \fi
\providecommand\bibfield[2]{#2}
\providecommand\bibinfo[2]{#2}
\providecommand\natexlab[1]{#1}
\providecommand\showeprint[2][]{arXiv:#2}

\bibitem[\protect\citeauthoryear{Blom, Erwander, Gustafsson, Landin-Olsson,
  Jonsson, and Ivarsson}{Blom et~al\mbox{.}}{2015}]%
        {Blom2015ThePO}
\bibfield{author}{\bibinfo{person}{Mathias~C Blom}, \bibinfo{person}{Karin
  Erwander}, \bibinfo{person}{Lars Gustafsson}, \bibinfo{person}{Mona
  Landin-Olsson}, \bibinfo{person}{Fredrik Jonsson}, {and}
  \bibinfo{person}{Kjell Ivarsson}.} \bibinfo{year}{2015}\natexlab{}.
\newblock \showarticletitle{{The probability of readmission within 30 days of
  hospital discharge is positively associated with inpatient bed occupancy at
  discharge – a retrospective cohort study}}.
\newblock \bibinfo{journal}{\emph{BMC Emergency Medicine}}
  \bibinfo{volume}{15}, \bibinfo{number}{1} (\bibinfo{year}{2015}),
  \bibinfo{pages}{37}.
\newblock
\showISSN{1471-227X}
\urldef\tempurl%
\url{https://doi.org/10.1186/s12873-015-0067-9}
\showDOI{\tempurl}


\bibitem[\protect\citeauthoryear{B{\"{o}}hmer, Just, Lefering, Paffrath,
  Bouillon, Joppich, Wappler, and Gerbershagen}{B{\"{o}}hmer
  et~al\mbox{.}}{2014}]%
        {Bohmer2014}
\bibfield{author}{\bibinfo{person}{Andreas~B B{\"{o}}hmer},
  \bibinfo{person}{Katja~S Just}, \bibinfo{person}{Rolf Lefering},
  \bibinfo{person}{Thomas Paffrath}, \bibinfo{person}{Bertil Bouillon},
  \bibinfo{person}{Robin Joppich}, \bibinfo{person}{Frank Wappler}, {and}
  \bibinfo{person}{Mark~U Gerbershagen}.} \bibinfo{year}{2014}\natexlab{}.
\newblock \showarticletitle{{Factors influencing lengths of stay in the
  intensive care unit for surviving trauma patients: a retrospective analysis
  of 30,157 cases}}.
\newblock \bibinfo{journal}{\emph{Critical care (London, England)}}
  \bibinfo{volume}{18}, \bibinfo{number}{4} (\bibinfo{year}{2014}),
  \bibinfo{pages}{R143--R143}.
\newblock


\bibitem[\protect\citeauthoryear{Cao, Wang, Li, Zhou, Li, and Li}{Cao
  et~al\mbox{.}}{2018}]%
        {Cao2018BRITSBR}
\bibfield{author}{\bibinfo{person}{Wei Cao}, \bibinfo{person}{Dong Wang},
  \bibinfo{person}{Jian Li}, \bibinfo{person}{Hao Zhou}, \bibinfo{person}{Lei
  Li}, {and} \bibinfo{person}{Yitan Li}.} \bibinfo{year}{2018}\natexlab{}.
\newblock \showarticletitle{BRITS: Bidirectional Recurrent Imputation for Time
  Series}. In \bibinfo{booktitle}{\emph{Advances in Neural Information
  Processing Systems}}, Vol.~\bibinfo{volume}{31}. \bibinfo{publisher}{Curran
  Associates, Inc.}
\newblock


\bibitem[\protect\citeauthoryear{Che, Purushotham, Cho, Sontag, and Liu}{Che
  et~al\mbox{.}}{2018}]%
        {Che2018}
\bibfield{author}{\bibinfo{person}{Zhengping Che}, \bibinfo{person}{Sanjay
  Purushotham}, \bibinfo{person}{Kyunghyun Cho}, \bibinfo{person}{David
  Sontag}, {and} \bibinfo{person}{Yan Liu}.} \bibinfo{year}{2018}\natexlab{}.
\newblock \showarticletitle{{Recurrent Neural Networks for Multivariate Time
  Series with Missing Values}}.
\newblock \bibinfo{journal}{\emph{Scientific Reports}} \bibinfo{volume}{8},
  \bibinfo{number}{1} (\bibinfo{year}{2018}), \bibinfo{pages}{6085}.
\newblock


\bibitem[\protect\citeauthoryear{Choi, Bahadori, Schuetz, Stewart, and
  Sun}{Choi et~al\mbox{.}}{2015}]%
        {Choi2015DoctorAP}
\bibfield{author}{\bibinfo{person}{Edward Choi}, \bibinfo{person}{Mohammad~Taha
  Bahadori}, \bibinfo{person}{Andy Schuetz}, \bibinfo{person}{Walter~F.
  Stewart}, {and} \bibinfo{person}{Jimeng Sun}.}
  \bibinfo{year}{2015}\natexlab{}.
\newblock \showarticletitle{{Doctor AI: Predicting Clinical Events via
  Recurrent Neural Networks}}.
\newblock \bibinfo{journal}{\emph{JMLR workshop and conference proceedings}}
  \bibinfo{volume}{56} (\bibinfo{year}{2015}), \bibinfo{pages}{301--318}.
\newblock


\bibitem[\protect\citeauthoryear{Chollet}{Chollet}{2017}]%
        {DBLP:journals/corr/Chollet16a}
\bibfield{author}{\bibinfo{person}{F. Chollet}.}
  \bibinfo{year}{2017}\natexlab{}.
\newblock \showarticletitle{Xception: Deep Learning with Depthwise Separable
  Convolutions}.
\newblock \bibinfo{journal}{\emph{2017 IEEE Conference on Computer Vision and
  Pattern Recognition (CVPR)}} (\bibinfo{year}{2017}),
  \bibinfo{pages}{1800--1807}.
\newblock


\bibitem[\protect\citeauthoryear{Cohen}{Cohen}{1960}]%
        {doi:10.1177/001316446002000104}
\bibfield{author}{\bibinfo{person}{Jacob Cohen}.}
  \bibinfo{year}{1960}\natexlab{}.
\newblock \showarticletitle{{A Coefficient of Agreement for Nominal Scales}}.
\newblock \bibinfo{journal}{\emph{Educational and Psychological Measurement}}
  \bibinfo{volume}{20}, \bibinfo{number}{1} (\bibinfo{year}{1960}),
  \bibinfo{pages}{37--46}.
\newblock


\bibitem[\protect\citeauthoryear{Dahl, G~Wojtal, Breslow, Holl, Huguez, Stone,
  and Korpi}{Dahl et~al\mbox{.}}{2012}]%
        {dahl2012}
\bibfield{author}{\bibinfo{person}{Deborah Dahl}, \bibinfo{person}{Greg
  G~Wojtal}, \bibinfo{person}{Michael Breslow}, \bibinfo{person}{Randy Holl},
  \bibinfo{person}{Debra Huguez}, \bibinfo{person}{David Stone}, {and}
  \bibinfo{person}{Gloria Korpi}.} \bibinfo{year}{2012}\natexlab{}.
\newblock \showarticletitle{{The High Cost of Low‐Acuity ICU Outliers}}.
\newblock \bibinfo{journal}{\emph{Journal of healthcare management / American
  College of Healthcare Executives}}  \bibinfo{volume}{57}
  (\bibinfo{year}{2012}), \bibinfo{pages}{421--434}.
\newblock


\bibitem[\protect\citeauthoryear{De~Brouwer, Simm, Arany, and
  Moreau}{De~Brouwer et~al\mbox{.}}{2019}]%
        {Brouwer2019GRUODEBayesCM}
\bibfield{author}{\bibinfo{person}{Edward De~Brouwer}, \bibinfo{person}{Jaak
  Simm}, \bibinfo{person}{Adam Arany}, {and} \bibinfo{person}{Yves Moreau}.}
  \bibinfo{year}{2019}\natexlab{}.
\newblock \showarticletitle{GRU-ODE-Bayes: Continuous Modeling of
  Sporadically-Observed Time Series}. In \bibinfo{booktitle}{\emph{Advances in
  Neural Information Processing Systems}}, Vol.~\bibinfo{volume}{32}.
  \bibinfo{publisher}{Curran Associates, Inc.}
\newblock


\bibitem[\protect\citeauthoryear{De~Silva, MacDonald, Paterson, Sikdar, and
  Cochrane}{De~Silva et~al\mbox{.}}{2011}]%
        {10.1016/j.cmpb.2011.01.002}
\bibfield{author}{\bibinfo{person}{Thuppahi~Sisira De~Silva},
  \bibinfo{person}{Don MacDonald}, \bibinfo{person}{Grace Paterson},
  \bibinfo{person}{Khokan~C. Sikdar}, {and} \bibinfo{person}{Bonnie Cochrane}.}
  \bibinfo{year}{2011}\natexlab{}.
\newblock \showarticletitle{{Systematized Nomenclature of Medicine Clinical
  Terms (SNOMED CT) to Represent Computed Tomography Procedures}}.
\newblock \bibinfo{journal}{\emph{Comput. Methods Prog. Biomed.}}
  \bibinfo{volume}{101}, \bibinfo{number}{3} (\bibinfo{year}{2011}),
  \bibinfo{pages}{324–329}.
\newblock


\bibitem[\protect\citeauthoryear{Elixhauser, Steiner, and Palmer}{Elixhauser
  et~al\mbox{.}}{2015}]%
        {ccs}
\bibfield{author}{\bibinfo{person}{A Elixhauser}, \bibinfo{person}{C Steiner},
  {and} \bibinfo{person}{L Palmer}.} \bibinfo{year}{2015}\natexlab{}.
\newblock \bibinfo{title}{{Clinical Classifications Software}}.
\newblock
\newblock


\bibitem[\protect\citeauthoryear{Fukushima}{Fukushima}{1980}]%
        {Fukushima1980}
\bibfield{author}{\bibinfo{person}{Kunihiko Fukushima}.}
  \bibinfo{year}{1980}\natexlab{}.
\newblock \showarticletitle{{Neocognitron: A Self-Organizing Neural Network
  Model for a Mechanism of Pattern Recognition Unaffected by Shift in
  Position}}.
\newblock \bibinfo{journal}{\emph{Biological Cybernetics}}
  \bibinfo{volume}{36}, \bibinfo{number}{4} (\bibinfo{year}{1980}),
  \bibinfo{pages}{193--202}.
\newblock


\bibitem[\protect\citeauthoryear{{Gentimis}, {Alnaser}, {Durante}, {Cook}, and
  {Steele}}{{Gentimis} et~al\mbox{.}}{2017}]%
        {gentimis2017}
\bibfield{author}{\bibinfo{person}{T. {Gentimis}}, \bibinfo{person}{A.~J.
  {Alnaser}}, \bibinfo{person}{A. {Durante}}, \bibinfo{person}{K. {Cook}},
  {and} \bibinfo{person}{R. {Steele}}.} \bibinfo{year}{2017}\natexlab{}.
\newblock \showarticletitle{{Predicting Hospital Length of Stay Using Neural
  Networks on MIMIC III Data}}. In \bibinfo{booktitle}{\emph{2017 IEEE 15th
  Intl Conf on Dependable, Autonomic and Secure Computing}}.
  \bibinfo{pages}{1194--1201}.
\newblock


\bibitem[\protect\citeauthoryear{Goldberger, Amaral, Glass,
  et~al\mbox{.}}{Goldberger et~al\mbox{.}}{2000}]%
        {Goldbergere215}
\bibfield{author}{\bibinfo{person}{A.~L. Goldberger}, \bibinfo{person}{L.~A.~N.
  Amaral}, \bibinfo{person}{L. Glass}, {et~al\mbox{.}}}
  \bibinfo{year}{2000}\natexlab{}.
\newblock \showarticletitle{{PhysioBank, PhysioToolkit, and PhysioNet:
  Components of a New Research Resource for Complex Physiologic Signals}}.
\newblock \bibinfo{journal}{\emph{Circulation}} \bibinfo{volume}{101},
  \bibinfo{number}{23} (\bibinfo{year}{2000}), \bibinfo{pages}{e215--e220}.
\newblock


\bibitem[\protect\citeauthoryear{Gong, Naumann, Szolovits, and Guttag}{Gong
  et~al\mbox{.}}{2017}]%
        {Gong2017PredictingCO}
\bibfield{author}{\bibinfo{person}{Jen~J. Gong}, \bibinfo{person}{Tristan
  Naumann}, \bibinfo{person}{Peter Szolovits}, {and} \bibinfo{person}{John~V.
  Guttag}.} \bibinfo{year}{2017}\natexlab{}.
\newblock \showarticletitle{{Predicting Clinical Outcomes Across Changing
  Electronic Health Record Systems}}. In \bibinfo{booktitle}{\emph{KDD}}.
\newblock


\bibitem[\protect\citeauthoryear{G{\"{u}}l{\c{c}}ehre, Moczulski, Denil, and
  Bengio}{G{\"{u}}l{\c{c}}ehre et~al\mbox{.}}{2016}]%
        {DBLP:journals/corr/GulcehreMDB16}
\bibfield{author}{\bibinfo{person}{{\c{C}}aglar G{\"{u}}l{\c{c}}ehre},
  \bibinfo{person}{Marcin Moczulski}, \bibinfo{person}{Misha Denil}, {and}
  \bibinfo{person}{Yoshua Bengio}.} \bibinfo{year}{2016}\natexlab{}.
\newblock \showarticletitle{{Noisy Activation Functions}}.
\newblock \bibinfo{journal}{\emph{CoRR}}  \bibinfo{volume}{abs/1603.00391}
  (\bibinfo{year}{2016}).
\newblock
\showeprint[arxiv]{1603.00391}


\bibitem[\protect\citeauthoryear{Halpern and Pastores}{Halpern and
  Pastores}{2015}]%
        {Halpern2015}
\bibfield{author}{\bibinfo{person}{Neil~A Halpern} {and}
  \bibinfo{person}{Stephen~M Pastores}.} \bibinfo{year}{2015}\natexlab{}.
\newblock \showarticletitle{{Critical Care Medicine Beds, Use, Occupancy, and
  Costs in the United States: A Methodological Review}}.
\newblock \bibinfo{journal}{\emph{Critical care medicine}}
  \bibinfo{volume}{43}, \bibinfo{number}{11} (\bibinfo{year}{2015}),
  \bibinfo{pages}{2452--2459}.
\newblock


\bibitem[\protect\citeauthoryear{{Harutyunyan}, {Khachatrian}, {Kale}, {Ver
  Steeg}, and {Galstyan}}{{Harutyunyan} et~al\mbox{.}}{2019}]%
        {harutyunyan}
\bibfield{author}{\bibinfo{person}{Hrayr {Harutyunyan}}, \bibinfo{person}{Hrant
  {Khachatrian}}, \bibinfo{person}{David~C. {Kale}}, \bibinfo{person}{Greg {Ver
  Steeg}}, {and} \bibinfo{person}{Aram {Galstyan}}.}
  \bibinfo{year}{2019}\natexlab{}.
\newblock \showarticletitle{{{Multitask Learning and Benchmarking with Clinical
  Time Series Data}}}.
\newblock \bibinfo{journal}{\emph{Scientific Data}} \bibinfo{volume}{6},
  \bibinfo{number}{96} (\bibinfo{year}{2019}).
\newblock


\bibitem[\protect\citeauthoryear{Hassan, Tuckman, Patrick, Kountz, and
  Kohn}{Hassan et~al\mbox{.}}{2010}]%
        {hassan}
\bibfield{author}{\bibinfo{person}{Mahmud Hassan}, \bibinfo{person}{Howard
  Tuckman}, \bibinfo{person}{Robert Patrick}, \bibinfo{person}{David Kountz},
  {and} \bibinfo{person}{Jennifer Kohn}.} \bibinfo{year}{2010}\natexlab{}.
\newblock \showarticletitle{{Hospital Length of Stay and Probability of
  Acquiring Infection}}.
\newblock \bibinfo{journal}{\emph{International Journal of Pharmaceutical and
  Healthcare Marketing}}  \bibinfo{volume}{4} (\bibinfo{year}{2010}),
  \bibinfo{pages}{324--338}.
\newblock


\bibitem[\protect\citeauthoryear{He, Zhang, Ren, and Sun}{He
  et~al\mbox{.}}{2015}]%
        {DBLP:journals/corr/HeZRS15}
\bibfield{author}{\bibinfo{person}{Kaiming He}, \bibinfo{person}{Xiangyu
  Zhang}, \bibinfo{person}{Shaoqing Ren}, {and} \bibinfo{person}{Jian Sun}.}
  \bibinfo{year}{2015}\natexlab{}.
\newblock \showarticletitle{{Deep Residual Learning for Image Recognition}}.
\newblock \bibinfo{journal}{\emph{CoRR}}  \bibinfo{volume}{abs/1512.03385}
  (\bibinfo{year}{2015}).
\newblock
\showeprint[arxiv]{1512.03385}


\bibitem[\protect\citeauthoryear{Hochreiter and Schmidhuber}{Hochreiter and
  Schmidhuber}{1997}]%
        {10.1162/neco.1997.9.8.1735}
\bibfield{author}{\bibinfo{person}{Sepp Hochreiter} {and}
  \bibinfo{person}{J{\"u}rgen Schmidhuber}.} \bibinfo{year}{1997}\natexlab{}.
\newblock \showarticletitle{{Long Short-Term Memory}}.
\newblock \bibinfo{journal}{\emph{Neural computation}} \bibinfo{volume}{9},
  \bibinfo{number}{8} (\bibinfo{year}{1997}), \bibinfo{pages}{1735--1780}.
\newblock


\bibitem[\protect\citeauthoryear{Huang, Liu, van~der Maaten, and
  Weinberger}{Huang et~al\mbox{.}}{2017}]%
        {densenet}
\bibfield{author}{\bibinfo{person}{Gao Huang}, \bibinfo{person}{Zhuang Liu},
  \bibinfo{person}{Laurens van~der Maaten}, {and} \bibinfo{person}{Kilian~Q
  Weinberger}.} \bibinfo{year}{2017}\natexlab{}.
\newblock \showarticletitle{Densely Connected Convolutional Networks}. In
  \bibinfo{booktitle}{\emph{Proceedings of the IEEE Conference on Computer
  Vision and Pattern Recognition}}.
\newblock


\bibitem[\protect\citeauthoryear{Ioffe and Szegedy}{Ioffe and Szegedy}{2015}]%
        {Ioffe2015}
\bibfield{author}{\bibinfo{person}{Sergey Ioffe} {and}
  \bibinfo{person}{Christian Szegedy}.} \bibinfo{year}{2015}\natexlab{}.
\newblock \showarticletitle{Batch Normalization: Accelerating Deep Network
  Training by Reducing Internal Covariate Shift}. In
  \bibinfo{booktitle}{\emph{Proceedings of the 32nd International Conference on
  International Conference on Machine Learning - Volume 37}} (Lille, France)
  \emph{(\bibinfo{series}{ICML'15})}. \bibinfo{publisher}{JMLR},
  \bibinfo{pages}{448--456}.
\newblock


\bibitem[\protect\citeauthoryear{Johnson, Bulgarelli, Pollard, Horng, Celi, and
  Mark}{Johnson et~al\mbox{.}}{2020}]%
        {johnson}
\bibfield{author}{\bibinfo{person}{A. Johnson}, \bibinfo{person}{L.
  Bulgarelli}, \bibinfo{person}{T. Pollard}, \bibinfo{person}{S. Horng},
  \bibinfo{person}{L.~A. Celi}, {and} \bibinfo{person}{R. Mark}.}
  \bibinfo{year}{2020}\natexlab{}.
\newblock \bibinfo{title}{{Medical Information Mart for Intensive Care IV}}.
\newblock \bibinfo{howpublished}{\url{https://doi.org/10.13026/a3wn-hq05}}.
\newblock


\bibitem[\protect\citeauthoryear{Kalchbrenner, Espeholt, Simonyan, van~den
  Oord, Graves, and Kavukcuoglu}{Kalchbrenner et~al\mbox{.}}{2016}]%
        {DBLP:journals/corr/KalchbrennerESO16}
\bibfield{author}{\bibinfo{person}{Nal Kalchbrenner}, \bibinfo{person}{Lasse
  Espeholt}, \bibinfo{person}{Karen Simonyan}, \bibinfo{person}{A{\"{a}}ron
  van~den Oord}, \bibinfo{person}{Alex Graves}, {and} \bibinfo{person}{Koray
  Kavukcuoglu}.} \bibinfo{year}{2016}\natexlab{}.
\newblock \showarticletitle{{Neural Machine Translation in Linear Time}}.
\newblock \bibinfo{journal}{\emph{CoRR}}  \bibinfo{volume}{abs/1610.10099}
  (\bibinfo{year}{2016}).
\newblock
\showeprint[arxiv]{1610.10099}


\bibitem[\protect\citeauthoryear{Kalra, Fisher, and Axelrod}{Kalra
  et~al\mbox{.}}{2010}]%
        {axel}
\bibfield{author}{\bibinfo{person}{Amit~D. Kalra}, \bibinfo{person}{Robert~S
  Fisher}, {and} \bibinfo{person}{Peter Axelrod}.}
  \bibinfo{year}{2010}\natexlab{}.
\newblock \showarticletitle{{Decreased Length of Stay and Cumulative
  Hospitalized Days despite Increased Patient Admissions and Readmissions in an
  Area of Urban Poverty.}}
\newblock \bibinfo{journal}{\emph{J Gen Intern Med.}} \bibinfo{volume}{25},
  \bibinfo{number}{9} (\bibinfo{year}{2010}), \bibinfo{pages}{920--935}.
\newblock


\bibitem[\protect\citeauthoryear{Kingma and Ba}{Kingma and Ba}{2014}]%
        {KingmaB14}
\bibfield{author}{\bibinfo{person}{Diederik~P. Kingma} {and}
  \bibinfo{person}{Jimmy Ba}.} \bibinfo{year}{2014}\natexlab{}.
\newblock \showarticletitle{Adam: A Method for Stochastic Optimization}.
\newblock \bibinfo{journal}{\emph{CoRR}}  \bibinfo{volume}{abs/1412.6980}
  (\bibinfo{year}{2014}).
\newblock


\bibitem[\protect\citeauthoryear{Laupland, Kirkpatrick, Kortbeek, and
  Zuege}{Laupland et~al\mbox{.}}{2006}]%
        {LAUPLAND2006954}
\bibfield{author}{\bibinfo{person}{Kevin~B. Laupland},
  \bibinfo{person}{Andrew~W. Kirkpatrick}, \bibinfo{person}{John~B. Kortbeek},
  {and} \bibinfo{person}{Danny~J. Zuege}.} \bibinfo{year}{2006}\natexlab{}.
\newblock \showarticletitle{{Long-term Mortality Outcome Associated With
  Prolonged Admission to the ICU}}.
\newblock \bibinfo{journal}{\emph{Chest}} \bibinfo{volume}{129},
  \bibinfo{number}{4} (\bibinfo{year}{2006}), \bibinfo{pages}{954 -- 959}.
\newblock


\bibitem[\protect\citeauthoryear{Lin, Chen, and Yan}{Lin et~al\mbox{.}}{2013}]%
        {lin2013network}
\bibfield{author}{\bibinfo{person}{Min Lin}, \bibinfo{person}{Qiang Chen},
  {and} \bibinfo{person}{Shuicheng Yan}.} \bibinfo{year}{2013}\natexlab{}.
\newblock \bibinfo{title}{{Network In Network}}.
\newblock
\newblock
\showeprint[arxiv]{1312.4400}~[cs.NE]


\bibitem[\protect\citeauthoryear{Lipton, Kale, Elkan, and Wetzel}{Lipton
  et~al\mbox{.}}{2015}]%
        {Lipton2015LearningTD}
\bibfield{author}{\bibinfo{person}{Zachary~Chase Lipton},
  \bibinfo{person}{David~C. Kale}, \bibinfo{person}{Charles Elkan}, {and}
  \bibinfo{person}{Randall~C. Wetzel}.} \bibinfo{year}{2015}\natexlab{}.
\newblock \showarticletitle{{Learning to Diagnose with LSTM Recurrent Neural
  Networks}}.
\newblock \bibinfo{journal}{\emph{CoRR}}  \bibinfo{volume}{abs/1511.03677}
  (\bibinfo{year}{2015}).
\newblock


\bibitem[\protect\citeauthoryear{Mak, Grant, McKenzie, and Mccabe}{Mak
  et~al\mbox{.}}{2012}]%
        {Mak2012PhysiciansAT}
\bibfield{author}{\bibinfo{person}{Gregory Mak}, \bibinfo{person}{William~D.
  Grant}, \bibinfo{person}{James~C McKenzie}, {and} \bibinfo{person}{John~B.
  Mccabe}.} \bibinfo{year}{2012}\natexlab{}.
\newblock \showarticletitle{{Physicians' Ability to Predict Hospital Length of
  Stay for Patients Admitted to the Hospital from the Emergency Department}}.
  In \bibinfo{booktitle}{\emph{Emergency medicine international}}.
\newblock


\bibitem[\protect\citeauthoryear{Mathews and Long}{Mathews and Long}{2015}]%
        {Mathews2015}
\bibfield{author}{\bibinfo{person}{Kusum~S Mathews} {and}
  \bibinfo{person}{Elisa~F Long}.} \bibinfo{year}{2015}\natexlab{}.
\newblock \showarticletitle{{A Conceptual Framework for Improving Critical Care
  Patient Flow and Bed Use}}.
\newblock \bibinfo{journal}{\emph{Annals of the American Thoracic Society}}
  \bibinfo{volume}{12}, \bibinfo{number}{6} (\bibinfo{year}{2015}),
  \bibinfo{pages}{886--894}.
\newblock


\bibitem[\protect\citeauthoryear{Mitchell, Wu, Zaldivar, Barnes, Vasserman,
  Hutchinson, Spitzer, Raji, and Gebru}{Mitchell et~al\mbox{.}}{2019}]%
        {Mitchell2019ModelCF}
\bibfield{author}{\bibinfo{person}{Margaret Mitchell}, \bibinfo{person}{Simone
  Wu}, \bibinfo{person}{A. Zaldivar}, \bibinfo{person}{P. Barnes},
  \bibinfo{person}{Lucy Vasserman}, \bibinfo{person}{B. Hutchinson},
  \bibinfo{person}{Elena Spitzer}, \bibinfo{person}{Inioluwa~Deborah Raji},
  {and} \bibinfo{person}{Timnit Gebru}.} \bibinfo{year}{2019}\natexlab{}.
\newblock \showarticletitle{Model Cards for Model Reporting}.
\newblock \bibinfo{journal}{\emph{Proceedings of the Conference on Fairness,
  Accountability, and Transparency}} (\bibinfo{year}{2019}).
\newblock


\bibitem[\protect\citeauthoryear{Mousavi, Ellsworth, Zhu, Chuang, and
  Beroza}{Mousavi et~al\mbox{.}}{2020}]%
        {Mousavi2020}
\bibfield{author}{\bibinfo{person}{S~Mostafa Mousavi},
  \bibinfo{person}{William~L Ellsworth}, \bibinfo{person}{Weiqiang Zhu},
  \bibinfo{person}{Lindsay~Y Chuang}, {and} \bibinfo{person}{Gregory~C
  Beroza}.} \bibinfo{year}{2020}\natexlab{}.
\newblock \showarticletitle{{Earthquake transformer—an attentive
  deep-learning model for simultaneous earthquake detection and phase
  picking}}.
\newblock \bibinfo{journal}{\emph{Nature Communications}} \bibinfo{volume}{11},
  \bibinfo{number}{1} (\bibinfo{year}{2020}), \bibinfo{pages}{3952}.
\newblock
\showISSN{2041-1723}
\urldef\tempurl%
\url{https://doi.org/10.1038/s41467-020-17591-w}
\showDOI{\tempurl}


\bibitem[\protect\citeauthoryear{Nestor, McDermott, Chauhan, Naumann, Hughes,
  Goldenberg, and Ghassemi}{Nestor et~al\mbox{.}}{2018}]%
        {DBLP:journals/corr/abs-1811-12583}
\bibfield{author}{\bibinfo{person}{Bret Nestor}, \bibinfo{person}{Matthew B.~A.
  McDermott}, \bibinfo{person}{Geeticka Chauhan}, \bibinfo{person}{Tristan
  Naumann}, \bibinfo{person}{Michael~C. Hughes}, \bibinfo{person}{Anna
  Goldenberg}, {and} \bibinfo{person}{Marzyeh Ghassemi}.}
  \bibinfo{year}{2018}\natexlab{}.
\newblock \showarticletitle{{Rethinking Clinical Prediction: Why Machine
  Learning must Consider Year of Care and Feature Aggregation}}.
\newblock \bibinfo{journal}{\emph{CoRR}}  \bibinfo{volume}{abs/1811.12583}
  (\bibinfo{year}{2018}).
\newblock
\showeprint[arxiv]{1811.12583}


\bibitem[\protect\citeauthoryear{{NHS Digital}}{{NHS Digital}}{2019}]%
        {Digital2019}
\bibfield{author}{\bibinfo{person}{{NHS Digital}}.}
  \bibinfo{year}{2019}\natexlab{}.
\newblock \bibinfo{title}{{DCB0084: OPCS-4.9 Requirements Specification}}.
\newblock
\newblock


\bibitem[\protect\citeauthoryear{Oh, Wang, and Wiens}{Oh et~al\mbox{.}}{2018}]%
        {pmlr-v85-oh18a}
\bibfield{author}{\bibinfo{person}{Jeeheh Oh}, \bibinfo{person}{Jiaxuan Wang},
  {and} \bibinfo{person}{Jenna Wiens}.} \bibinfo{year}{2018}\natexlab{}.
\newblock \showarticletitle{Learning to Exploit Invariances in Clinical
  Time-Series Data using Sequence Transformer Networks}. In
  \bibinfo{booktitle}{\emph{Proceedings of the 3rd Machine Learning for
  Healthcare Conference}} \emph{(\bibinfo{series}{Proceedings of Machine
  Learning Research}, Vol.~\bibinfo{volume}{85})},
  \bibfield{editor}{\bibinfo{person}{Finale Doshi-Velez}, \bibinfo{person}{Jim
  Fackler}, \bibinfo{person}{Ken Jung}, \bibinfo{person}{David Kale},
  \bibinfo{person}{Rajesh Ranganath}, \bibinfo{person}{Byron Wallace}, {and}
  \bibinfo{person}{Jenna Wiens}} (Eds.). \bibinfo{publisher}{PMLR},
  \bibinfo{address}{Palo Alto, California}, \bibinfo{pages}{332--347}.
\newblock
\urldef\tempurl%
\url{http://proceedings.mlr.press/v85/oh18a.html}
\showURL{%
\tempurl}


\bibitem[\protect\citeauthoryear{Paszke, Gross, Massa, et~al\mbox{.}}{Paszke
  et~al\mbox{.}}{2019}]%
        {NEURIPS2019_9015}
\bibfield{author}{\bibinfo{person}{Adam Paszke}, \bibinfo{person}{Sam Gross},
  \bibinfo{person}{Francisco Massa}, {et~al\mbox{.}}}
  \bibinfo{year}{2019}\natexlab{}.
\newblock \showarticletitle{{PyTorch: An Imperative Style, High-Performance
  Deep Learning Library}}.
\newblock In \bibinfo{booktitle}{\emph{Advances in Neural Information
  Processing Systems 32}}. \bibinfo{publisher}{Curran Associates, Inc.},
  \bibinfo{pages}{8024--8035}.
\newblock


\bibitem[\protect\citeauthoryear{Pollard, Johnson, Raffa, Celi, Mark, and
  Badawi}{Pollard et~al\mbox{.}}{2018}]%
        {Pollard2018}
\bibfield{author}{\bibinfo{person}{Tom~J Pollard}, \bibinfo{person}{Alistair
  E~W Johnson}, \bibinfo{person}{Jesse~D Raffa}, \bibinfo{person}{Leo~A Celi},
  \bibinfo{person}{Roger~G Mark}, {and} \bibinfo{person}{Omar Badawi}.}
  \bibinfo{year}{2018}\natexlab{}.
\newblock \showarticletitle{{The eICU Collaborative Research Database, A Freely
  Available Multi-Center Database for Critical Care Research}}.
\newblock \bibinfo{journal}{\emph{Scientific Data}} \bibinfo{volume}{5},
  \bibinfo{number}{1} (\bibinfo{year}{2018}), \bibinfo{pages}{180178}.
\newblock


\bibitem[\protect\citeauthoryear{Prasad, Cheng, Chivers, Draugelis, and
  Engelhardt}{Prasad et~al\mbox{.}}{2017}]%
        {DBLP:journals/corr/PrasadCCDE17}
\bibfield{author}{\bibinfo{person}{Niranjani Prasad},
  \bibinfo{person}{Li{-}Fang Cheng}, \bibinfo{person}{Corey Chivers},
  \bibinfo{person}{Michael Draugelis}, {and} \bibinfo{person}{Barbara~E.
  Engelhardt}.} \bibinfo{year}{2017}\natexlab{}.
\newblock \showarticletitle{{A Reinforcement Learning Approach to Weaning of
  Mechanical Ventilation in Intensive Care Units}}.
\newblock \bibinfo{journal}{\emph{CoRR}}  \bibinfo{volume}{abs/1704.06300}
  (\bibinfo{year}{2017}).
\newblock
\showeprint[arxiv]{1704.06300}


\bibitem[\protect\citeauthoryear{Purushotham, Meng, Che, and Liu}{Purushotham
  et~al\mbox{.}}{2017}]%
        {purushotham2017benchmark}
\bibfield{author}{\bibinfo{person}{Sanjay Purushotham},
  \bibinfo{person}{Chuizheng Meng}, \bibinfo{person}{Zhengping Che}, {and}
  \bibinfo{person}{Yan Liu}.} \bibinfo{year}{2017}\natexlab{}.
\newblock \bibinfo{title}{Benchmark of Deep Learning Models on Large Healthcare
  MIMIC Datasets}.
\newblock
\newblock
\showeprint[arxiv]{1710.08531}~[cs.LG]


\bibitem[\protect\citeauthoryear{Purushotham, Meng, Che, and Liu}{Purushotham
  et~al\mbox{.}}{2018}]%
        {purushotham}
\bibfield{author}{\bibinfo{person}{S. Purushotham}, \bibinfo{person}{C. Meng},
  \bibinfo{person}{Z. Che}, {and} \bibinfo{person}{Y. Liu}.}
  \bibinfo{year}{2018}\natexlab{}.
\newblock \showarticletitle{{Benchmarking Deep Learning Models on Large
  Healthcare Datasets.}}
\newblock \bibinfo{journal}{\emph{Journal of Biomedical Informatics}}
  \bibinfo{volume}{83} (\bibinfo{year}{2018}), \bibinfo{pages}{112--134}.
\newblock


\bibitem[\protect\citeauthoryear{Rajkomar, Oren, Chen, et~al\mbox{.}}{Rajkomar
  et~al\mbox{.}}{2018}]%
        {Rajkomar2018ScalableAA}
\bibfield{author}{\bibinfo{person}{Alvin Rajkomar}, \bibinfo{person}{Eyal
  Oren}, \bibinfo{person}{Kai Chen}, {et~al\mbox{.}}}
  \bibinfo{year}{2018}\natexlab{}.
\newblock \showarticletitle{{Scalable and Accurate Deep Learning with
  Electronic Health Records}}. In \bibinfo{booktitle}{\emph{npj Digital
  Medicine}}.
\newblock


\bibitem[\protect\citeauthoryear{Rapoport, Teres, Zhao, and Lemeshow}{Rapoport
  et~al\mbox{.}}{2003}]%
        {rapoport2003}
\bibfield{author}{\bibinfo{person}{John Rapoport}, \bibinfo{person}{Daniel
  Teres}, \bibinfo{person}{Yonggang Zhao}, {and} \bibinfo{person}{Stanley
  Lemeshow}.} \bibinfo{year}{2003}\natexlab{}.
\newblock \showarticletitle{{Length of Stay Data as a Guide to Hospital
  Economic Performance for ICU Patients}}.
\newblock \bibinfo{journal}{\emph{Medical Care}}  \bibinfo{volume}{41}
  (\bibinfo{year}{2003}), \bibinfo{pages}{386--397}.
\newblock


\bibitem[\protect\citeauthoryear{Razavian, Marcus, and Sontag}{Razavian
  et~al\mbox{.}}{2016}]%
        {sontag}
\bibfield{author}{\bibinfo{person}{Narges Razavian}, \bibinfo{person}{Jake
  Marcus}, {and} \bibinfo{person}{David Sontag}.}
  \bibinfo{year}{2016}\natexlab{}.
\newblock \showarticletitle{Multi-task Prediction of Disease Onsets from
  Longitudinal Laboratory Tests} \emph{(\bibinfo{series}{Proceedings of Machine
  Learning Research}, Vol.~\bibinfo{volume}{56})},
  \bibfield{editor}{\bibinfo{person}{Finale Doshi-Velez}, \bibinfo{person}{Jim
  Fackler}, \bibinfo{person}{David Kale}, \bibinfo{person}{Byron Wallace},
  {and} \bibinfo{person}{Jenna Wiens}} (Eds.). \bibinfo{publisher}{PMLR},
  \bibinfo{address}{Northeastern University, Boston, MA, USA},
  \bibinfo{pages}{73--100}.
\newblock
\urldef\tempurl%
\url{http://proceedings.mlr.press/v56/Razavian16.html}
\showURL{%
\tempurl}


\bibitem[\protect\citeauthoryear{Razavian and Sontag}{Razavian and
  Sontag}{2015}]%
        {DBLP:journals/corr/RazavianS15}
\bibfield{author}{\bibinfo{person}{Narges Razavian} {and}
  \bibinfo{person}{David~A. Sontag}.} \bibinfo{year}{2015}\natexlab{}.
\newblock \showarticletitle{Temporal Convolutional Neural Networks for
  Diagnosis from Lab Tests}.
\newblock \bibinfo{journal}{\emph{CoRR}}  \bibinfo{volume}{abs/1511.07938}
  (\bibinfo{year}{2015}).
\newblock
\showeprint[arxiv]{1511.07938}
\urldef\tempurl%
\url{http://arxiv.org/abs/1511.07938}
\showURL{%
\tempurl}


\bibitem[\protect\citeauthoryear{Rubanova, Chen, and Duvenaud}{Rubanova
  et~al\mbox{.}}{2019}]%
        {Rubanova2019LatentOF}
\bibfield{author}{\bibinfo{person}{Yulia Rubanova}, \bibinfo{person}{Ricky
  T.~Q. Chen}, {and} \bibinfo{person}{David Duvenaud}.}
  \bibinfo{year}{2019}\natexlab{}.
\newblock \showarticletitle{Latent ODEs for Irregularly-Sampled Time Series}.
\newblock \bibinfo{journal}{\emph{NeurIPS}}  \bibinfo{volume}{abs/1907.03907}
  (\bibinfo{year}{2019}).
\newblock


\bibitem[\protect\citeauthoryear{Ruder}{Ruder}{2017}]%
        {multitaskoverfitting}
\bibfield{author}{\bibinfo{person}{Sebastian Ruder}.}
  \bibinfo{year}{2017}\natexlab{}.
\newblock \showarticletitle{An Overview of Multi-Task Learning in Deep Neural
  Networks}.
\newblock \bibinfo{journal}{\emph{CoRR}}  \bibinfo{volume}{abs/1706.05098}
  (\bibinfo{year}{2017}).
\newblock
\showeprint[arxiv]{1706.05098}
\urldef\tempurl%
\url{http://arxiv.org/abs/1706.05098}
\showURL{%
\tempurl}


\bibitem[\protect\citeauthoryear{Sendak, Elish, Gao, Futoma, Ratliff, Nichols,
  Bedoya, Balu, and O'Brien}{Sendak et~al\mbox{.}}{2020a}]%
        {sepsiswatch}
\bibfield{author}{\bibinfo{person}{Mark Sendak},
  \bibinfo{person}{Madeleine~Clare Elish}, \bibinfo{person}{Michael Gao},
  \bibinfo{person}{Joseph Futoma}, \bibinfo{person}{William Ratliff},
  \bibinfo{person}{Marshall Nichols}, \bibinfo{person}{Armando Bedoya},
  \bibinfo{person}{Suresh Balu}, {and} \bibinfo{person}{Cara O'Brien}.}
  \bibinfo{year}{2020}\natexlab{a}.
\newblock \showarticletitle{"The Human Body is a Black Box": Supporting
  Clinical Decision-Making with Deep Learning}. In
  \bibinfo{booktitle}{\emph{Proceedings of the 2020 Conference on Fairness,
  Accountability, and Transparency}} (Barcelona, Spain)
  \emph{(\bibinfo{series}{FAT* '20})}. \bibinfo{publisher}{Association for
  Computing Machinery}, \bibinfo{address}{New York, NY, USA},
  \bibinfo{pages}{99–109}.
\newblock
\showISBNx{9781450369367}
\urldef\tempurl%
\url{https://doi.org/10.1145/3351095.3372827}
\showDOI{\tempurl}


\bibitem[\protect\citeauthoryear{Sendak, Gao, Brajer, and Balu}{Sendak
  et~al\mbox{.}}{2020b}]%
        {Sendak2020}
\bibfield{author}{\bibinfo{person}{Mark~P Sendak}, \bibinfo{person}{Michael
  Gao}, \bibinfo{person}{Nathan Brajer}, {and} \bibinfo{person}{Suresh Balu}.}
  \bibinfo{year}{2020}\natexlab{b}.
\newblock \showarticletitle{{Presenting machine learning model information to
  clinical end users with model facts labels}}.
\newblock \bibinfo{journal}{\emph{npj Digital Medicine}} \bibinfo{volume}{3},
  \bibinfo{number}{1} (\bibinfo{year}{2020}), \bibinfo{pages}{41}.
\newblock
\showISSN{2398-6352}
\urldef\tempurl%
\url{https://doi.org/10.1038/s41746-020-0253-3}
\showDOI{\tempurl}


\bibitem[\protect\citeauthoryear{Sheikhalishahi, Balaraman, and
  Osmani}{Sheikhalishahi et~al\mbox{.}}{2019}]%
        {sheikhalishahi2019benchmarking}
\bibfield{author}{\bibinfo{person}{Seyedmostafa Sheikhalishahi},
  \bibinfo{person}{Vevake Balaraman}, {and} \bibinfo{person}{Venet Osmani}.}
  \bibinfo{year}{2019}\natexlab{}.
\newblock \bibinfo{title}{{Benchmarking Machine Learning Models on eICU
  Critical Care Dataset}}.
\newblock
\newblock
\showeprint[arxiv]{1910.00964}~[cs.LG]


\bibitem[\protect\citeauthoryear{Shickel, Loftus, Adhikari, Ozrazgat-Baslanti,
  Bihorac, and Rashidi}{Shickel et~al\mbox{.}}{2019}]%
        {Shickel2019DeepSOFAAC}
\bibfield{author}{\bibinfo{person}{Benjamin Shickel}, \bibinfo{person}{Tyler~J.
  Loftus}, \bibinfo{person}{Lasith Adhikari}, \bibinfo{person}{Tezcan
  Ozrazgat-Baslanti}, \bibinfo{person}{Azra Bihorac}, {and}
  \bibinfo{person}{Parisa Rashidi}.} \bibinfo{year}{2019}\natexlab{}.
\newblock \showarticletitle{{DeepSOFA: A Continuous Acuity Score for Critically
  Ill Patients using Clinically Interpretable Deep Learning}}. In
  \bibinfo{booktitle}{\emph{Scientific Reports}}.
\newblock


\bibitem[\protect\citeauthoryear{Song, Rajan, Thiagarajan, and Spanias}{Song
  et~al\mbox{.}}{2018}]%
        {2304ed73e858419398e3ee1508af5825}
\bibfield{author}{\bibinfo{person}{Huan Song}, \bibinfo{person}{Deepta Rajan},
  \bibinfo{person}{{Jayaraman J.} Thiagarajan}, {and} \bibinfo{person}{Andreas
  Spanias}.} \bibinfo{year}{2018}\natexlab{}.
\newblock \showarticletitle{{Attend and Diagnose: Clinical Time Series Analysis
  using Attention Models}}. In \bibinfo{booktitle}{\emph{32nd AAAI Conference
  on Artificial Intelligence, AAAI 2018}}. \bibinfo{pages}{4091--4098}.
\newblock


\bibitem[\protect\citeauthoryear{Srivastava, Hinton, Krizhevsky, Sutskever, and
  Salakhutdinov}{Srivastava et~al\mbox{.}}{2014}]%
        {Srivastava2014}
\bibfield{author}{\bibinfo{person}{Nitish Srivastava},
  \bibinfo{person}{Geoffrey Hinton}, \bibinfo{person}{Alex Krizhevsky},
  \bibinfo{person}{Ilya Sutskever}, {and} \bibinfo{person}{Ruslan
  Salakhutdinov}.} \bibinfo{year}{2014}\natexlab{}.
\newblock \showarticletitle{{Dropout: A Simple Way to Prevent Neural Networks
  from Overfitting}}.
\newblock \bibinfo{journal}{\emph{Journal of Machine Learning Research}}
  \bibinfo{volume}{15} (\bibinfo{year}{2014}), \bibinfo{pages}{1929--1958}.
\newblock


\bibitem[\protect\citeauthoryear{Sturmfels, Lundberg, and Lee}{Sturmfels
  et~al\mbox{.}}{2020}]%
        {sturmfels2020visualizing}
\bibfield{author}{\bibinfo{person}{Pascal Sturmfels}, \bibinfo{person}{Scott
  Lundberg}, {and} \bibinfo{person}{Su-In Lee}.}
  \bibinfo{year}{2020}\natexlab{}.
\newblock \showarticletitle{Visualizing the Impact of Feature Attribution
  Baselines}.
\newblock \bibinfo{journal}{\emph{Distill}} \bibinfo{volume}{5},
  \bibinfo{number}{1} (\bibinfo{year}{2020}), \bibinfo{pages}{e22}.
\newblock
\urldef\tempurl%
\url{https://distill.pub/2020/attribution-baselines/}
\showURL{%
\tempurl}


\bibitem[\protect\citeauthoryear{Sundararajan, Taly, and Yan}{Sundararajan
  et~al\mbox{.}}{2017}]%
        {integratedgradients}
\bibfield{author}{\bibinfo{person}{Mukund Sundararajan}, \bibinfo{person}{Ankur
  Taly}, {and} \bibinfo{person}{Qiqi Yan}.} \bibinfo{year}{2017}\natexlab{}.
\newblock \showarticletitle{Axiomatic Attribution for Deep Networks}. In
  \bibinfo{booktitle}{\emph{Proceedings of the 34th International Conference on
  Machine Learning - Volume 70}} (Sydney, NSW, Australia)
  \emph{(\bibinfo{series}{ICML'17})}. \bibinfo{publisher}{JMLR.org},
  \bibinfo{pages}{3319–3328}.
\newblock


\bibitem[\protect\citeauthoryear{Suresh, Hunt, Johnson, Celi, Szolovits, and
  Ghassemi}{Suresh et~al\mbox{.}}{2017}]%
        {Suresh2017ClinicalIP}
\bibfield{author}{\bibinfo{person}{Harini Suresh}, \bibinfo{person}{Nathan
  Hunt}, \bibinfo{person}{Alistair E.~W. Johnson}, \bibinfo{person}{Leo~Anthony
  Celi}, \bibinfo{person}{Peter Szolovits}, {and} \bibinfo{person}{Marzyeh
  Ghassemi}.} \bibinfo{year}{2017}\natexlab{}.
\newblock \showarticletitle{{Clinical Intervention Prediction and Understanding
  with Deep Neural Networks}}. In \bibinfo{booktitle}{\emph{MLHC}}.
\newblock


\bibitem[\protect\citeauthoryear{Szegedy, Liu, Jia, Sermanet, Reed, Anguelov,
  Erhan, Vanhoucke, and Rabinovich}{Szegedy et~al\mbox{.}}{2014}]%
        {DBLP:journals/corr/SzegedyLJSRAEVR14}
\bibfield{author}{\bibinfo{person}{Christian Szegedy}, \bibinfo{person}{Wei
  Liu}, \bibinfo{person}{Yangqing Jia}, \bibinfo{person}{Pierre Sermanet},
  \bibinfo{person}{Scott~E. Reed}, \bibinfo{person}{Dragomir Anguelov},
  \bibinfo{person}{Dumitru Erhan}, \bibinfo{person}{Vincent Vanhoucke}, {and}
  \bibinfo{person}{Andrew Rabinovich}.} \bibinfo{year}{2014}\natexlab{}.
\newblock \showarticletitle{{Going Deeper with Convolutions}}.
\newblock \bibinfo{journal}{\emph{CoRR}}  \bibinfo{volume}{abs/1409.4842}
  (\bibinfo{year}{2014}).
\newblock
\showeprint[arxiv]{1409.4842}


\bibitem[\protect\citeauthoryear{Toma{\v{s}}ev, Glorot, Rae,
  et~al\mbox{.}}{Toma{\v{s}}ev et~al\mbox{.}}{2019}]%
        {Tomaev2019ACA}
\bibfield{author}{\bibinfo{person}{Nenad Toma{\v{s}}ev},
  \bibinfo{person}{Xavier Glorot}, \bibinfo{person}{Jack~W Rae},
  {et~al\mbox{.}}} \bibinfo{year}{2019}\natexlab{}.
\newblock \showarticletitle{{A Clinically Applicable Approach to Continuous
  Prediction of Future Acute Kidney Injury}}.
\newblock \bibinfo{journal}{\emph{Nature}} \bibinfo{volume}{572},
  \bibinfo{number}{7767} (\bibinfo{year}{2019}), \bibinfo{pages}{116--119}.
\newblock


\bibitem[\protect\citeauthoryear{Tonekaboni, Mazwi, Laussen, Eytan, Greer,
  Goodfellow, Goodwin, Brudno, and Goldenberg}{Tonekaboni
  et~al\mbox{.}}{2018}]%
        {Tonekaboni2018PredictionOC}
\bibfield{author}{\bibinfo{person}{Sana Tonekaboni}, \bibinfo{person}{Mjaye
  Mazwi}, \bibinfo{person}{Peter Laussen}, \bibinfo{person}{Danny Eytan},
  \bibinfo{person}{Robert Greer}, \bibinfo{person}{Sebastian~D. Goodfellow},
  \bibinfo{person}{Andrew Goodwin}, \bibinfo{person}{Michael Brudno}, {and}
  \bibinfo{person}{Anna Goldenberg}.} \bibinfo{year}{2018}\natexlab{}.
\newblock \showarticletitle{{Prediction of Cardiac Arrest from Physiological
  Signals in the Pediatric ICU}}. In \bibinfo{booktitle}{\emph{MLHC}}.
\newblock


\bibitem[\protect\citeauthoryear{van~den Oord, Dieleman, Zen, Simonyan,
  Vinyals, Graves, Kalchbrenner, Senior, and Kavukcuoglu}{van~den Oord
  et~al\mbox{.}}{2016}]%
        {Simonyan2016}
\bibfield{author}{\bibinfo{person}{A{\"{a}}ron van~den Oord},
  \bibinfo{person}{Sander Dieleman}, \bibinfo{person}{Heiga Zen},
  \bibinfo{person}{Karen Simonyan}, \bibinfo{person}{Oriol Vinyals},
  \bibinfo{person}{Alex Graves}, \bibinfo{person}{Nal Kalchbrenner},
  \bibinfo{person}{Andrew~W. Senior}, {and} \bibinfo{person}{Koray
  Kavukcuoglu}.} \bibinfo{year}{2016}\natexlab{}.
\newblock \showarticletitle{{WaveNet: {A} Generative Model for Raw Audio}}.
\newblock \bibinfo{journal}{\emph{CoRR}}  \bibinfo{volume}{abs/1609.03499}
  (\bibinfo{year}{2016}).
\newblock
\showeprint[arxiv]{1609.03499}


\bibitem[\protect\citeauthoryear{Vaswani, Shazeer, Parmar, Uszkoreit, Jones,
  Gomez, Kaiser, and Polosukhin}{Vaswani et~al\mbox{.}}{2017}]%
        {46201}
\bibfield{author}{\bibinfo{person}{Ashish Vaswani}, \bibinfo{person}{Noam
  Shazeer}, \bibinfo{person}{Niki Parmar}, \bibinfo{person}{Jakob Uszkoreit},
  \bibinfo{person}{Llion Jones}, \bibinfo{person}{Aidan~N. Gomez},
  \bibinfo{person}{undefinedukasz Kaiser}, {and} \bibinfo{person}{Illia
  Polosukhin}.} \bibinfo{year}{2017}\natexlab{}.
\newblock \showarticletitle{Attention is All You Need}. In
  \bibinfo{booktitle}{\emph{Proceedings of the 31st International Conference on
  Neural Information Processing Systems}} \emph{(\bibinfo{series}{NIPS’17})}.
  \bibinfo{publisher}{Curran Associates Inc.}, \bibinfo{pages}{6000–6010}.
\newblock


\bibitem[\protect\citeauthoryear{Wallace, Angus, Seymour, Barnato, and
  Kahn}{Wallace et~al\mbox{.}}{2015}]%
        {doi:10.1164/rccm.201409-1746OC}
\bibfield{author}{\bibinfo{person}{David~J. Wallace}, \bibinfo{person}{Derek~C.
  Angus}, \bibinfo{person}{Christopher~W. Seymour}, \bibinfo{person}{Amber~E.
  Barnato}, {and} \bibinfo{person}{Jeremy~M. Kahn}.}
  \bibinfo{year}{2015}\natexlab{}.
\newblock \showarticletitle{{Critical Care Bed Growth in the United States. A
  Comparison of Regional and National Trends}}.
\newblock \bibinfo{journal}{\emph{American Journal of Respiratory and Critical
  Care Medicine}} \bibinfo{volume}{191}, \bibinfo{number}{4}
  (\bibinfo{year}{2015}), \bibinfo{pages}{410--416}.
\newblock
\newblock
\shownote{PMID: 25522054.}


\bibitem[\protect\citeauthoryear{Wang, Yi, Jiang, Jiang, Han, Lu, and Ma}{Wang
  et~al\mbox{.}}{2018}]%
        {wang20188}
\bibfield{author}{\bibinfo{person}{Zhongyuan Wang}, \bibinfo{person}{Peng Yi},
  \bibinfo{person}{Kui Jiang}, \bibinfo{person}{Junjun Jiang},
  \bibinfo{person}{Zhen Han}, \bibinfo{person}{Tao Lu}, {and}
  \bibinfo{person}{Jiayi Ma}.} \bibinfo{year}{2018}\natexlab{}.
\newblock \showarticletitle{Multi-Memory Convolutional Neural Network for Video
  Super-Resolution}.
\newblock \bibinfo{journal}{\emph{IEEE Transactions on Image Processing}}
  \bibinfo{volume}{PP} (\bibinfo{date}{12} \bibinfo{year}{2018}),
  \bibinfo{pages}{1--1}.
\newblock
\urldef\tempurl%
\url{https://doi.org/10.1109/TIP.2018.2887017}
\showDOI{\tempurl}


\bibitem[\protect\citeauthoryear{{World Health Organisation}}{{World Health
  Organisation}}{2011}]%
        {icd10}
\bibfield{author}{\bibinfo{person}{{World Health Organisation}}.}
  \bibinfo{year}{2011}\natexlab{}.
\newblock \bibinfo{booktitle}{\emph{{ICD-10: International Statistical
  Classification of Diseases and Related Health Problems}}}.
  Vol.~\bibinfo{volume}{10th Revision}.
\newblock \bibinfo{publisher}{World Health Organision}.
\newblock


\bibitem[\protect\citeauthoryear{Zimmerer, Petersen, Köhler, Wasserthal,
  Adler, Wirkert, and Ross}{Zimmerer et~al\mbox{.}}{2017}]%
        {trixi2017}
\bibfield{author}{\bibinfo{person}{David Zimmerer}, \bibinfo{person}{Jens
  Petersen}, \bibinfo{person}{Gregor Köhler}, \bibinfo{person}{Jakob
  Wasserthal}, \bibinfo{person}{Tim Adler}, \bibinfo{person}{Sebastian
  Wirkert}, {and} \bibinfo{person}{Tobias Ross}.}
  \bibinfo{year}{2017}\natexlab{}.
\newblock \bibinfo{title}{{trixi - Training and Retrospective Insight
  eXperiment Infrastructure}}.
\newblock \bibinfo{howpublished}{\url{https://github.com/MIC-DKFZ/trixi}}.
\newblock
\urldef\tempurl%
\url{https://doi.org/10.5281/zenodo.1345136}
\showDOI{\tempurl}


\bibitem[\protect\citeauthoryear{Zimmerman, Kramer, McNair, and
  Malila}{Zimmerman et~al\mbox{.}}{2006}]%
        {Zimmerman2006}
\bibfield{author}{\bibinfo{person}{Jack~E Zimmerman}, \bibinfo{person}{Andrew~A
  Kramer}, \bibinfo{person}{Douglas~S McNair}, {and} \bibinfo{person}{Fern~M
  Malila}.} \bibinfo{year}{2006}\natexlab{}.
\newblock \showarticletitle{{Acute Physiology and Chronic Health Evaluation
  (APACHE) IV: Hospital Mortality Assessment for Today's Critically Ill
  Patients}}.
\newblock \bibinfo{journal}{\emph{Read Online: Critical Care Medicine | Society
  of Critical Care Medicine}} \bibinfo{volume}{34}, \bibinfo{number}{5}
  (\bibinfo{year}{2006}).
\newblock


\end{thebibliography}

\appendix

\section{Model Architecture: Further Details}
\label{sec:fullmodel}
\begin{figure}
\centering
\hspace{5em}\includegraphics[width=0.4\textwidth]{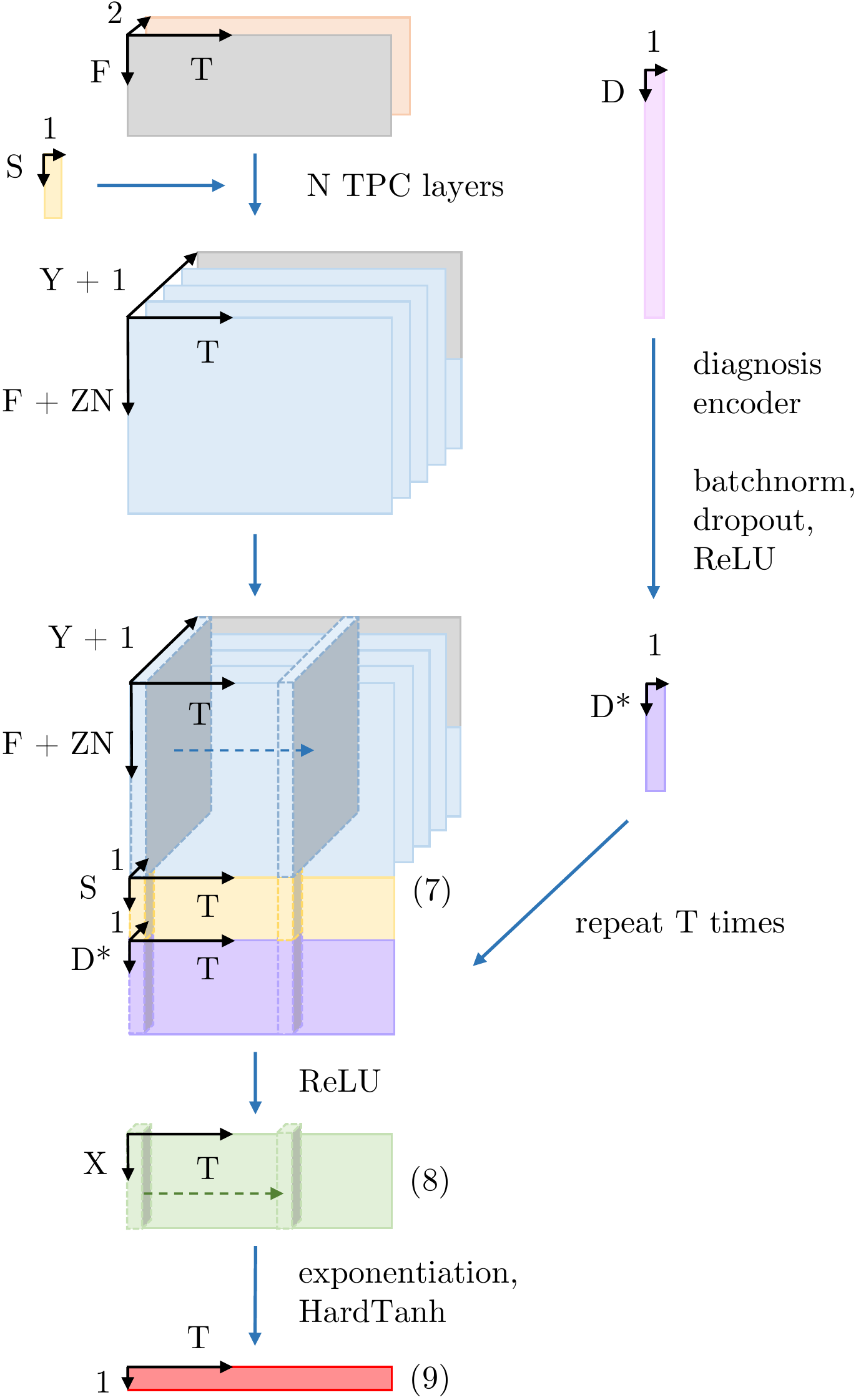}
\caption{The original time series, $\mathbf{x'}$ (grey) and the decay indicators, $\mathbf{x''}$ (orange) are processed by $N$ TPC layers before being combined with a diagnosis embedding $\mathbf{d^*}$ (purple) and static features $\mathbf{s}$ (yellow) along the feature axis. A two-layer pointwise convolution is applied to obtain the final predictions $\mathbf{\hat{y}}$ (red).}
\label{fig:modeloverview}
\end{figure}

After $N$ TPC layers, we apply two further pointwise convolutions to obtain the final predictions. Formally, these final steps (shown in Figure~\ref{fig:modeloverview}) can be written as
\begin{equation}
    \underbrace{\hat{y}_t}_{(9)} = \text{HardTanh}\bigg(\text{exp}\Big(g'' \ast \underbrace{\sigma\Big(g' \ast \overbrace{\Big[\flat(\mathbf{h}_t^{N}) \parallel \mathbf{s} \parallel \mathbf{d^*}\Big]}^{\text{Final Combined In.}(7)}}_{\text{Penultimate Point.\ Out.}(8)}\Big)\Big)\bigg)
\end{equation}
where $B=R^{N}\times(Y+1) + S + D^*$ and the final pointwise filters are $g':\{1,\ldots,B\}\to\mathbb{R}^{X\times 1}$ and $g'':\{1,\ldots,X\}\to\mathbb{R}^{1\times 1}$. Note that if a baseline model were to be used instead of TPC, the output dimensions would be $H$ x $1$ instead of $B$ x $1$, where $H$ is the LSTM hidden size or $d_{model}$ in the Transformer. We apply an exponential function to allow the upstream model to predict $\log(\text{LoS})$ instead of LoS. We hypothesised that this could help to circumvent a common issue seen in previous models (e.g.\ \citet{harutyunyan}, as they struggle to produce predictions over the full dynamic range of length of stays). Finally, we apply a HardTanh function \citep{DBLP:journals/corr/GulcehreMDB16} to clip any predictions that are smaller than 30 minutes or larger than 100 days, which protects against inflated MSLE loss values.
\begin{equation}
        \text{HardTanh}(x) =
        \begin{cases}
            100,          & \text{if } x~\text{\textgreater}~100,\\
            \frac{1}{48},          & \text{if } x~\text{\textless}~\frac{1}{48},\\
            x,         & \text{otherwise.}
        \end{cases}
\end{equation}

\section{Feature Pre-processing}
\label{preproc}
\subsection{Static Features}
We selected 17 static features from eICU (shown in Table~\ref{tab:static}) and 12 from MIMIC-IV (Table~\ref{tab:staticMIMIC}). Discrete and continuous variables were scaled to the interval [-1, 1], using the 5th and 95th percentiles as the boundaries, and absolute cut offs were placed at [-4, 4]. This was to protect against large or erroneous inputs, while avoiding assumptions about the variable distributions. Binary variables were coded as 1 and 0. Categorical variables were converted to one-hot encodings.

\subsection{Diagnoses}
Here we only describe pre-processing for eICU since MIMIC-IV did not contain coded diagnoses with appropriate timestamps.

Like many EHRs, diagnosis coding in eICU is hierarchical. At the lowest level they can be quite specific e.g.\ ``neurologic $|$ disorders of vasculature $|$ stroke $|$ hemorrhagic stroke $|$ subarachnoid hemorrhage $|$ with vasospasm''. To maintain the hierarchical structure within a flat vector, we assigned separate features to each hierarchical level and use binary encoding. This produces a vector of size 4,436 with an average sparsity of 99.5\% (only 0.5\% of the data is positive). We apply a 1\% prevalence cut-off on all these features to reduce the size of the vector to 293 and the average sparsity to 93.3\%. If a disease does not make the cut-off for inclusion, it is still included via any parent classes that do make the cut-off (in the above example we record everything up to ``subarachnoid hemorrhage''). We only included diagnoses that were recorded before the 5th hour in the ICU, to avoid leakage from the future.

Many diagnostic and interventional coding systems are hierarchical in nature: ICD-10 classification \citep{icd10}, Clinical Classifications Software \citep{ccs}, SNOMED CT \citep{10.1016/j.cmpb.2011.01.002} and OPCS Classification of Interventions and Procedures \citep{Digital2019}, so this technique is generalisable to other coding systems present in EHRs. 

\subsection{Time Series}
\label{timeseriespreproc}
For each admission, we extracted 87 time-varying features from eICU (Table~\ref{tab:timeseries}) and 101 from MIMIC-IV (Table~\ref{tab:timeseriesMIMIC}) for each hour of the ICU visit, and up to 24 hours before the ICU visit. The variables were processed in the same manner as the static features. In general, the sampling is very irregular, so the data was re-sampled according to one hour intervals and forward-filled. After forward-filling is complete, any data recorded before the ICU admission is removed. Decay indicators are added as described in Section~\ref{data}.

\begin{table}[h]
    \caption{eICU static features. Age \textgreater 89, Null Height and Null Weight were added as indicator variables to indicate when the age was more than 89 but has been capped, and when the height or weight were missing and have been imputed with the mean value.}
    \label{tab:static}
    \centering
    \begin{tabular}{lll}
        \toprule
        \textbf{Feature} & \textbf{Type} & \textbf{Source Table} \\
        \midrule
        Gender & Binary & \textit{patient} \\
        Age & Discrete & \textit{patient} \\
        Hour of Admission & Discrete & \textit{patient} \\
        Height & Continuous & \textit{patient} \\
        Weight & Continuous & \textit{patient} \\
        Ethnicity & Categorical & \textit{patient} \\
        Unit Type & Categorical & \textit{patient} \\
        Unit Admit Source & Categorical & \textit{patient} \\
        Unit Visit Number & Categorical & \textit{patient} \\
        Unit Stay Type & Categorical & \textit{patient} \\
        Num Beds Category & Categorical & \textit{hospital} \\
        Region & Categorical & \textit{hospital} \\
        Teaching Status & Binary & \textit{hospital} \\
        Physician Speciality & Categorical & \textit{apachepatientresult} \\
        Age \textgreater 89 & Binary & \\
        Null Height & Binary & \\
        Null Weight & Binary & \\
        \bottomrule
    \end{tabular}
\end{table}

\begin{table}[h]
    \caption{MIMIC-IV static features. Age was calculated from the `intime' field in the \textit{icustays} table and `anchor year' in the \textit{patients} table.}
    \label{tab:staticMIMIC}
    \centering
    \begin{tabular}{lll}
        \toprule
        \textbf{Feature} & \textbf{Type} & \textbf{Source Table} \\
        \midrule
        Gender & Binary & \textit{patients} \\
        Ethnicity & Categorical & \textit{admissions} \\
        Admission Location & Categorical & \textit{admissions} \\
        Insurance Type & Categorical & \textit{admissions} \\
        First Careunit & Categorical & \textit{icustays} \\
        Hour of Admission & Discrete & \textit{icustays} \\
        Admission Height & Continuous & \textit{chartevents} \\
        Admission Weight & Continuous & \textit{chartevents} \\
        Eyes & Discrete & \textit{chartevents} \\
        Motor & Discrete & \textit{chartevents} \\
        Verbal & Discrete & \textit{chartevents} \\
        Age & Discrete & \\
        \bottomrule
    \end{tabular}
\end{table}

\section{Hyperparameter Search Methodology and Implementation Details}
\label{hyperparamsearch}

The TPC model and baselines have hyperparameters that can broadly be split into three categories: time series specific, non-time series specific and global parameters (shown in more detail in Tables~\ref{tab:TPChyperparams}, \ref{tab:LSTMhyperparams} and \ref{tab:Transformerhyperparams}). The hyperparameter search ranges have been included in Table~\ref{tab:hyperparamsearch}. 

First, we ran 25 randomly sampled hyperparameter trials on the TPC model to decide the non-time series specific parameters (diagnosis embedding size, final fully connected layer size, batch normalisation strategy, dropout rate and the parameter $\alpha$) keeping all other parameters fixed. These parameters (indicated by stars) remained fixed for all the models which share their non-time series specific architecture (NB. the best value for $\alpha$ was 100 -- not shown in the Tables). 

We then ran 50 hyperparameter trials to optimise the remaining parameters for the TPC, standard LSTM, and Transformer models. To train the channel-wise LSTM and the temporal model with weight sharing, we ran a further 10 trials to re-optimise the hidden size (8 per feature) and number of temporal channels (32 channels shared across all features) respectively. For all other ablation studies and variations of each model, we kept the same hyperparameters where applicable (see Table~\ref{tab:results} for a full list of all of the models). The number of epochs was determined by selecting the best validation performance from a model trained over 50 epochs. This was different for each model. For eICU this was 8 (LSTM), 30 (CW LSTM), 15 (Transformer) and 15 (TPC). For MIMIC-IV this was 8 (LSTM), 20 (CW LSTM), 15 (Transformer) and 10 (TPC). We noted that the best LSTM hyperparameters (Table~\ref{tab:LSTMhyperparams}) were similar to that found in \citet{sheikhalishahi2019benchmarking}. 

\begin{table}
  \caption{The TPC model has 11 hyperparameters (Main Dropout and Batch Normalisation have been repeated in the table because they apply to multiple parts of the model). We allowed the model to optimise a custom dropout rate for the temporal convolutions because they have fewer parameters and might need less regularisation than the rest of the model. The best hyperparameter values are shown in brackets (eICU/MIMIC-IV). Hyperparameters marked with * were fixed across all of the models.}
  \label{tab:TPChyperparams}
  \centering
  \begin{tabular}{ll}
    \toprule
    \multicolumn{2}{c}{\textbf{TPC Specific}}\\
    \textbf{Temporal Specific} & \textbf{Pointwise Specific}\\
    \midrule
    Temp.\ Channels (12/11)&Point.\ Channels (13/5)\\
    Temp.\ Dropout (0.05/0.05)&Main Dropout* (0.45/0)\\
    Kernel Size (4/5)& \\
    \multicolumn{2}{c}{Batch Normalisation* (True/True)}\\
    \multicolumn{2}{c}{No. TPC Layers (9/8)}\\
    \vspace{-0.8em}\\
    \toprule
    \textbf{Non-TPC Specific} & \textbf{Global Parameters}\\
    \midrule
    Diag. Embedding Size* (64/-)&Batch Size (32/8)\\
    Main Dropout* (0.45/0)&Learning Rate (0.00226/\\
    Final FC Layer Size* (17/36)&0.00221)\\
    Batch Normalisation* (True/True)&\\
    \bottomrule
  \end{tabular}
\end{table}

\begin{table}
  \caption{The LSTM model has 9 hyperparameters. We allowed the model to optimise a custom dropout rate for the LSTM layers. Note that batch normalisation is not applicable to the LSTM layers. The best parameters are shown as (eICU/MIMIC-IV).}
  \label{tab:LSTMhyperparams}
  \centering
  \begin{tabular}{ll}
    \toprule
    \textbf{LSTM Specific} & \textbf{Non-LSTM Specific}\\
    \midrule
    Hidden State (128/128)&Diag. Embedding Size* (64/-)\\
    LSTM Dropout (0.2/0.25)&Main Dropout* (0.45/0)\\
    No. LSTM Layers (2/1)&Final FC Layer Size* (17/36)\\
    &Batch Normalisation* (True/True)\\
    \vspace{-0.8em}\\
    \toprule
    \multicolumn{2}{c}{\textbf{Global Parameters}}\\
    \midrule
    \multicolumn{2}{c}{Batch Size (512/32)}\\
    \multicolumn{2}{c}{Learning Rate (0.00129/0.00163)}\\
    \bottomrule
  \end{tabular}
\end{table}

All deep learning methods were implemented in PyTorch \citep{NEURIPS2019_9015} and were optimised using Adam \citep{KingmaB14}. The data (including decay indicators) and the non-time series components of the models were the same as in TPC (Figure~\ref{fig:modeloverview}). We used trixi to structure our experiments and compare different hyperparameter choices \citep{trixi2017}. 

The experiments were performed using resources provided by the Cambridge Tier-2 system operated by the University of Cambridge Research Computing Service (\url{www.hpc.cam.ac.uk}) funded by EPSRC Tier-2 capital grant EP/P020259/1.

\subsection{Transformer}
\label{transformer}

The Transformer is a multi-head self-attention model, originally designed for sequence-to-sequence tasks in natural language processing. It consists of both an encoder and decoder, however we only use the former. Our implementation is the same as the original encoder in \citet{46201}, except that we add temporal masking to impose causality i.e. the current representation can only depend on current or earlier timepoints, and we omit the positional encodings because they were not found to be helpful. This is probably because we already have a feature to indicate the position in the time series (Section~\ref{timeseriespreproc}).

\begin{table}[h]
  \caption{The Transformer model has 12 hyperparameters. We allowed the model to optimise a custom dropout rate for the Transformer layers. The positional encoding hyperparameter is binary; it determines whether or not we used the original positional encodings proposed by \citet{46201}. Note that batch normalisation is not applicable to the Transformer layers (the default implementation uses layer normalisation). The best parameters are shown as (eICU/MIMIC-IV).}
  \label{tab:Transformerhyperparams}
  \centering
  \begin{tabular}{ll}
    \toprule
    \textbf{Transformer Specific} & \textbf{Non-Transformer Specific} \\
    \midrule
    No. Attention Heads (2/1)&Diag. Embedding Size* (64/-)\\
    Feedforward Size (256/64)&Main Dropout* (0.45/0)\\
    $d_{model}$ (16/32)&Final FC Layer Size* (17/36)\\
    Transformer Dropout (0/0.05)& /True)\\
    No. Transformer Layers (6/2)&\\
    \vspace{-0.8em}\\
    \toprule
    \multicolumn{2}{c}{\textbf{Global Parameters}}\\
    \midrule
    \multicolumn{2}{c}{Batch Size (32/64)}\\
    \multicolumn{2}{c}{Learning Rate (0.00017/0.00129)}\\
    \bottomrule
  \end{tabular}
\end{table}

\begin{table}[h]
  \caption{Hyperparameter Search Ranges. We took a random sample from each range and converted to an integer if necessary. For the kernel sizes (not shown in the table) the range was dependent on the number of TPC layers selected (because large kernel sizes combined with a large number of layers can have an inappropriately wide range as the dilation factor increases per layer). In general the range of kernel sizes was around 2-5 (but it could be up to 10 for small numbers of TPC Layers).}
  \label{tab:hyperparamsearch}
  \small
  \centering
  \begin{tabular}{llll}
    \toprule
    \textbf{Hyperparameter} & \textbf{Lower} & \textbf{Upper} & \textbf{Scale}\\
    \midrule
    Batch Size & 4 & 512 & $\log_2$\\
    Dropout Rate (all) & 0 & 0.5 & Linear\\
    Learning Rate & 0.0001 & 0.01 & $\log_{10}$\\
    Batch Normalisation & True & False & \\
    Positional Encoding & True & False & \\
    Diagnosis Embedding Size & 16 & 64 & $\log_2$\\
    Final FC Layer Size & 16 & 64 & $\log_2$\\
    CW LSTM Hidden State Size & 4 & 16 & $\log_2$\\
    Point.\ Channels & 4 & 16 & $\log_2$\\
    Temp.\ Channels & 4 & 16 & $\log_2$\\
    Temp.\ Channels (weight sharing) & 16 & 64 & $\log_2$\\
    LSTM Hidden State Size & 16 & 256 & $\log_2$\\
    $d_{model}$ & 16 & 256 & $\log_2$\\
    Feedforward Size & 16 & 256 & $\log_2$\\
    No. Attention Heads & 2 & 16 & $\log_2$\\
    No. TPC Layers & 1 & 12 & Linear\\
    No. LSTM Layers & 1 & 4 & Linear\\
    No. Transformer Layers & 1 & 10 & Linear\\
    \bottomrule
  \end{tabular}
\end{table}

\section{Evaluation Metrics}
\label{evaluationmetrics}
The metrics we use are: mean absolute deviation (MAD), mean absolute percentage error (MAPE), mean squared error (MSE), mean squared loss error (MSLE), coefficient of determination ($R^2$) and Cohen Kappa Score. We modify the MAPE metric slightly so that very small true LoS values do not produce unbounded MAPE values. We place a 4 hour lower bound on the divisor i.e. 
\begin{equation*}
    \text{Absolute Percentage Error} = \abs{\frac{y_{true} - y_{pred}}{\max{(y_{true}, \frac{4}{24})}}}*100
\end{equation*} 
MAD and MAPE are improved by centering predictions on the median. Likewise, MSE and $R^2$ are bettered by centering predictions around the mean. They are more affected by the skew. MSLE is a good metric for this task, indeed, it is the loss function in most experiments, but is less readily-interpretable than some of the other measures. Cohen's linear weighted Kappa Score \citep{doi:10.1177/001316446002000104} is intended for ordered classification tasks rather than regression, but it can effectively mitigate for skew if the bins are chosen well. It has previously provided useful insights in \citet{harutyunyan}, so we use the same LoS bins: 0-1, 1-2, 2-3, 3-4, 4-5, 5-6, 6-7, 7-8, 8-14, and 14+ days. As a classification measure, it will treat everything falling within the same classification bin as equal, so it is fundamentally a coarser measure than the other metrics.

\begin{table*}[ht]
  \caption{Performance of the TPC model and its baselines when only some of the time series are included (the flat features and diagnoses are still included). The indicator `(labs)' means that only the laboratory tests were included, `(other)' refers to everything except labs: vital signs, nurse observations and machine logged variables. The metric acronyms, colour scheme and confidence interval calculations are described in Table~\ref{tab:results}. The percentage impairment when compared to the complete dataset is shown in grey underneath the absolute values. They are calculated with respect to the best value for the metric: 0 for MAD, MAPE, MSE and MSLE, and 1 for $R^2$ and Kappa. A large percentage impairment means that the model does much better with complete data i.e.\ it has a high `percentage gain' from the combination of both data types compared to the ablation case.}
  \label{tab:additionalresults}
  \centering
  \begin{tabular}{p{3cm}|p{1.4cm}p{1.4cm}p{1.4cm}p{1.4cm}p{1.4cm}p{1.4cm}}
    \toprule
        \textbf{Model} & \textbf{MAD} & \textbf{MAPE} & \textbf{MSE} & \textbf{MSLE} & \boldmath{$R^2$} & \textbf{Kappa} \\
    \midrule
        LSTM & {\textBF{\textcolor{blue}{2.39$\pm$0.00}}} & {\textBF{\textcolor{blue}{118.2$\pm$1.1}}} & {\textBF{\textcolor{blue}{26.9$\pm$0.1}}} & {\textBF{\textcolor{blue}{1.47$\pm$0.01}}} & {\textBF{\textcolor{blue}{0.09$\pm$0.00}}} & {\textBF{\textcolor{blue}{0.28$\pm$0.00}}} \\
        LSTM (labs) & {2.43$\pm$0.00} & {123.8$\pm$1.2} & {27.3$\pm$0.1} & {1.57$\pm$0.00} & {0.08$\pm$0.00} & {0.27$\pm$0.00} \\
        & {\textcolor{gray}{(-1.7\%)}} & {\textcolor{gray}{(-4.7\%)}} & {\textcolor{gray}{(-1.5\%)}} & {\textcolor{gray}{(-6.8\%)}} & {\textcolor{gray}{(-1.1\%)}} & {\textcolor{gray}{(-1.4\%)}} \\
        LSTM (other) & {2.41$\pm$0.00} & {120.2$\pm$0.7} & {27.3$\pm$0.1} & {1.49$\pm$0.00} & {0.07$\pm$0.00} & {0.27$\pm$0.00} \\
        & {\textcolor{gray}{(-0.8\%)}} & {\textcolor{gray}{(-1.7\%)}} & {\textcolor{gray}{(-1.5\%)}} & {\textcolor{gray}{(-1.4\%)}} & {\textcolor{gray}{(-2.2\%)}} & {\textcolor{gray}{(-1.4\%)}} \\
    \midrule
        CW LSTM & {\textBF{\textcolor{blue}{2.37$\pm$0.00}}} & {\textBF{\textcolor{blue}{114.5$\pm$0.4}}} & {\textBF{\textcolor{blue}{26.6$\pm$0.1}}} & {\textBF{\textcolor{blue}{1.43$\pm$0.00}}} & {\textBF{\textcolor{blue}{0.10$\pm$0.00}}} & {\textBF{\textcolor{blue}{0.30$\pm$0.00}}} \\
        CW LSTM (labs) & {2.42$\pm$0.00} & {124.4$\pm$0.7} & {27.0$\pm$0.1} & {1.57$\pm$0.00} & {0.08$\pm$0.00} & {0.28$\pm$0.00} \\
        & {\textcolor{gray}{(-2.1\%)}} & {\textcolor{gray}{(-8.6\%)}} & {\textcolor{gray}{(-1.5\%)}} & {\textcolor{gray}{(-9.8\%)}} & {\textcolor{gray}{(-2.2\%)}} & {\textcolor{gray}{(-2.9\%)}} \\
        CW LSTM (other) & {2.41$\pm$0.00} & {120.6$\pm$0.8} & {27.1$\pm$0.1} & {1.51$\pm$0.00} & {0.08$\pm$0.00} & {0.29$\pm$0.00} \\
        & {\textcolor{gray}{(-1.7\%)}} & {\textcolor{gray}{(-5.3\%)}} & {\textcolor{gray}{(-1.9\%)}} & {\textcolor{gray}{(-5.6\%)}} & {\textcolor{gray}{(-2.2\%)}} & {\textcolor{gray}{(-1.4\%)}} \\
    \midrule
        Transformer & {\textBF{\textcolor{blue}{2.36$\pm$0.00}}} & {\textBF{\textcolor{blue}{114.1$\pm$0.6}}} & {\textBF{\textcolor{blue}{26.7$\pm$0.1}}} & {\textBF{\textcolor{blue}{1.43$\pm$0.00}}} & {\textBF{\textcolor{blue}{0.09$\pm$0.00}}} & {\textBF{\textcolor{blue}{0.30$\pm$0.00}}} \\
        Transformer (labs) & {2.42$\pm$0.00} & {121.0$\pm$0.7} & {27.3$\pm$0.1} & {1.56$\pm$0.00} & {0.07$\pm$0.00} & {0.27$\pm$0.00} \\
        & {\textcolor{gray}{(-2.5\%)}} & {\textcolor{gray}{(-6.0\%)}} & {\textcolor{gray}{(-2.2\%)}} & {\textcolor{gray}{(-9.1\%)}} & {\textcolor{gray}{(-2.2\%)}} & {\textcolor{gray}{(-4.3\%)}} \\
        Transformer (other) & {2.40$\pm$0.00} & {118.3$\pm$0.6} & {27.3$\pm$0.1} & {1.50$\pm$0.00} & {0.07$\pm$0.00} & {0.27$\pm$0.00} \\
        & {\textcolor{gray}{(-1.7\%)}} & {\textcolor{gray}{(-3.7\%)}} & {\textcolor{gray}{(-2.2\%)}} & {\textcolor{gray}{(-4.9\%)}} & {\textcolor{gray}{(-2.2\%)}} & {\textcolor{gray}{(-4.3\%)}} \\
    \midrule
        TPC & {\textBF{\textcolor{blue}{1.78$\pm$0.02}}} & {\textBF{\textcolor{blue}{63.5$\pm$4.3}}} & {\textBF{\textcolor{blue}{21.7$\pm$0.5}}} & {\textBF{\textcolor{blue}{0.70$\pm$0.03}}} & {\textBF{\textcolor{blue}{0.27$\pm$0.02}}} & {\textBF{\textcolor{blue}{0.58$\pm$0.01}}} \\
        TPC (labs) & {1.85$\pm$0.01} & {72.0$\pm$2.2} & {22.5$\pm$0.2} & {0.81$\pm$0.01} & {0.24$\pm$0.01} & {0.55$\pm$0.00} \\
        & {\textcolor{gray}{(-3.9\%)}} & {\textcolor{gray}{(-13.4\%)}} & {\textcolor{gray}{(-3.7\%)}} & {\textcolor{gray}{(-15.7\%)}} & {\textcolor{gray}{(-4.1\%)}} & {\textcolor{gray}{(-7.1\%)}} \\
        TPC (other) & {\textBF{\textcolor{lightblue}{1.81$\pm$0.02}}} & {\textBF{\textcolor{lightblue}{68.5$\pm$4.7}}} & {\textBF{\textcolor{lightblue}{21.8$\pm$0.3}}} & {0.77$\pm$0.03} & {\textBF{\textcolor{lightblue}{0.26$\pm$0.01}}} & {\textBF{\textcolor{lightblue}{0.57$\pm$0.01}}} \\
        & {\textcolor{gray}{(-1.7\%)}} & {\textcolor{gray}{(-7.9\%)}} & {\textcolor{gray}{(-0.5\%)}} & {\textcolor{gray}{(-10.0\%)}} & {\textcolor{gray}{(-1.4\%)}} & {\textcolor{gray}{(-2.4\%)}} \\
    \bottomrule
    \end{tabular}
\end{table*}

To illustrate the importance of using multiple metrics, consider that the mean and median models are in some sense equally poor (neither has learned anything meaningful for our purposes). Nevertheless, the median model is able to better exploit the MAD, MAPE and MSLE metrics, and the mean model fares better with MSE, but the Kappa score betrays them both. A good model will perform well across all of the metrics.

\section{Time Series Ablation}
\label{tsablation}

We performed ablations on the type of time series variable that we include: laboratory tests only (labs), which are infrequently sampled, and all other variables (other) which include vital signs, nurse observations, and automatically recorded variables (e.g.\ from ventilator machines). This shows how well each model can cope with time series exhibiting different periodicity and sampling frequencies. The results are shown in Table~\ref{tab:additionalresults}. 
The TPC model has the largest percentage gain when the labs and other variables are combined (this is synonymous with the greatest percentage impairment in the ablations). Next are the CW LSTM and Transformer, followed by the LSTM. This suggests that the TPC model is best able to exploit EHR time series with different temporal properties. 

When examining the results for LSTM and CW LSTM in more detail, we can see that the CW LSTM only has an advantage when the model has to combine the data types. This supports the hypothesis that the CW LSTM is better able to cope when there are varying frequencies in the data, as it can tailor the processing to each. When the inter-feature variability is small (the same type of time series) they perform similarly.

It is unsurprising that the Transformer does better than the LSTM when combining data types, as it can directly skip over large gaps in time to extract a trend in lab values, while simultaneously attending to recent timepoints for the processing of other variables. 

The TPC is the most successful model; its inherent periodic structure helps it to extract useful information from all of the variables. The CW LSTM and Transformer do not have this in their architectures, making the derivation more obscure. The importance of periodicity is discussed in more detail in Section~\ref{discussion}.

\begin{table*}[ht]
    \caption{The effect of changing the size of the training data on the LSTM, CW LSTM, Transformer, and TPC model performance in the eICU dataset. A hundred percent of the training set represents 102,712 ICU stays, 50\% is 51,356, 25\% is 25,678, 12.5\% is 12,839, and 6.25\% is 6,420 stays.}
    \label{tab:extraablationresults}
    \centering
    \begin{tabular}{p{2.9cm}|p{1.3cm}p{1.3cm}p{1.3cm}p{1.3cm}p{1.3cm}p{1.3cm}}
    \toprule
        \textbf{Model (\% train data)} & \textbf{MAD} & \textbf{MAPE} & \textbf{MSE} & \textbf{MSLE} & \boldmath{$R^2$} & \textbf{Kappa} \\
    \midrule
        LSTM (100) & {2.39$\pm$0.00} & {118.2$\pm$1.1} & {26.9$\pm$0.1} & {1.47$\pm$0.01} & {0.09$\pm$0.00} & {0.28$\pm$0.00} \\
        LSTM (50) & {2.41$\pm$0.01} & {129.9$\pm$1.9} & {26.2$\pm$0.2} & {1.52$\pm$0.00} & {0.11$\pm$0.01} & {0.31$\pm$0.01} \\
        LSTM (25) & {2.44$\pm$0.01} & {126.8$\pm$2.5} & {27.2$\pm$0.3} & {1.58$\pm$0.00} & {0.08$\pm$0.01} & {0.27$\pm$0.01} \\
        LSTM (12.5) & {2.48$\pm$0.01} & {137.4$\pm$3.4} & {27.4$\pm$0.2} & {1.65$\pm$0.01} & {0.07$\pm$0.01} & {0.27$\pm$0.01} \\
        LSTM (6.25) & {2.52$\pm$0.02} & {135.9$\pm$3.3} & {28.0$\pm$0.8} & {1.71$\pm$0.02} & {0.05$\pm$0.03} & {0.26$\pm$0.03} \\
    \midrule
        CW LSTM (100) & {2.37$\pm$0.00} & {114.5$\pm$0.4} & {26.6$\pm$0.1} & {1.43$\pm$0.00} & {0.10$\pm$0.00} & {0.30$\pm$0.00} \\
        CW LSTM (50) & {2.40$\pm$0.01} & {123.4$\pm$0.7} & {26.5$\pm$0.1} & {1.48$\pm$0.01} & {0.10$\pm$0.00} & {0.31$\pm$0.00} \\
        CW LSTM (25) & {2.44$\pm$0.00} & {119.8$\pm$1.3} & {27.2$\pm$0.1} & {1.54$\pm$0.00} & {0.08$\pm$0.00} & {0.29$\pm$0.00} \\
        CW LSTM (12.5) & {2.50$\pm$0.01} & {134.7$\pm$1.5} & {27.7$\pm$0.1} & {1.63$\pm$0.01} & {0.06$\pm$0.00} & {0.28$\pm$0.00} \\
        CW LSTM (6.25) & {2.58$\pm$0.01} & {129.8$\pm$3.5} & {29.0$\pm$0.2} & {1.73$\pm$0.01} & {0.02$\pm$0.01} & {0.25$\pm$0.01} \\
    \midrule
        Transformer (100) & {2.36$\pm$0.00} & {114.1$\pm$0.6} & {26.7$\pm$0.1} & {1.43$\pm$0.00} & {0.09$\pm$0.00} & {0.30$\pm$0.00} \\
        Transformer (50) & {2.39$\pm$0.00} & {120.1$\pm$0.6} & {26.5$\pm$0.1} & {1.48$\pm$0.00} & {0.10$\pm$0.00} & {0.31$\pm$0.00} \\
        Transformer (25) & {2.43$\pm$0.01} & {117.9$\pm$1.8} & {27.2$\pm$0.2} & {1.54$\pm$0.01} & {0.08$\pm$0.01} & {0.28$\pm$0.01} \\
        Transformer (12.5) & {2.48$\pm$0.01} & {128.1$\pm$2.3} & {27.9$\pm$0.1} & {1.62$\pm$0.01} & {0.06$\pm$0.00} & {0.26$\pm$0.01} \\
        Transformer (6.25) & {2.52$\pm$0.01} & {139.7$\pm$2.4} & {27.8$\pm$0.1} & {1.69$\pm$0.02} & {0.06$\pm$0.00} & {0.26$\pm$0.00} \\
    \midrule
        TPC (100) & {1.78$\pm$0.02} & {63.5$\pm$4.3} & {21.7$\pm$0.5} & {0.70$\pm$0.03} & {0.27$\pm$0.02} & {0.58$\pm$0.01} \\
        TPC (50) & {1.95$\pm$0.02} & {72.0$\pm$3.1} & {23.8$\pm$0.4} & {0.87$\pm$0.03} & {0.19$\pm$0.01} & {0.51$\pm$0.01} \\
        TPC (25) & {2.09$\pm$0.01} & {89.0$\pm$3.8} & {24.8$\pm$0.3} & {1.09$\pm$0.02} & {0.16$\pm$0.01} & {0.45$\pm$0.01} \\
        TPC (12.5) & {2.28$\pm$0.01} & {101.4$\pm$4.8} & {27.0$\pm$0.4} & {1.36$\pm$0.03} & {0.08$\pm$0.01} & {0.35$\pm$0.02} \\
        TPC (6.25) & {2.49$\pm$0.02} & {139.9$\pm$5.5} & {28.0$\pm$0.3} & {1.64$\pm$0.03} & {0.05$\pm$0.01} & {0.28$\pm$0.01} \\
    \bottomrule
    \end{tabular}
\end{table*}

\begin{table*}[h]
    \caption{The effect of training with the mean squared logarithmic error (MSLE) loss function when compared to mean squared error (MSE) on the eICU dataset. This is an extension to Table~\ref{tab:ablationresults} (refer to its legend for definitions of the metric acronyms, detailed of CI calculations and meaning of the colour scheme).}
    \label{tab:mseresults}
    \centering
    \begin{tabular}{p{3.2cm}|p{1.4cm}p{1.4cm}p{1.25cm}p{1.4cm}p{1.4cm}p{1.4cm}}
    \toprule
        \textbf{Model} & \textbf{MAD} & \textbf{MAPE} & \textbf{MSE} & \textbf{MSLE} & \boldmath{$R^2$} & \textbf{Kappa} \\
    \midrule
        LSTM (MSE) & {2.57$\pm$0.03} & {235.2$\pm$6.2} & {\textBF{\textcolor{blue}{24.5$\pm$0.2}}}$^{\footnotesize{**}}$ & {1.97$\pm$0.02} & {\textBF{\textcolor{blue}{0.17$\pm$0.01}}}$^{\footnotesize{**}}$ & {\textBF{\textcolor{blue}{0.28$\pm$0.01}}} \\
        LSTM (MSLE) & {\textBF{\textcolor{blue}{2.39$\pm$0.00}}}$^{\footnotesize{**}}$ & {\textBF{\textcolor{blue}{118.2$\pm$1.1}}}$^{\footnotesize{**}}$ & {26.9$\pm$0.1} & {\textBF{\textcolor{blue}{1.47$\pm$0.01}}}$^{\footnotesize{**}}$ & {0.09$\pm$0.00} & {\textBF{\textcolor{blue}{0.28$\pm$0.00}}} \\
    \midrule
        CW LSTM (MSE) & {2.56$\pm$0.01} & {218.5$\pm$4.0} & {\textBF{\textcolor{blue}{24.2$\pm$0.1}}}$^{\footnotesize{**}}$ & {1.84$\pm$0.02} & {\textBF{\textcolor{blue}{0.18$\pm$0.00}}}$^{\footnotesize{**}}$ & {\textBF{\textcolor{blue}{0.34$\pm$0.01}}}$^{\footnotesize{**}}$ \\
        CW LSTM (MSLE) & {\textBF{\textcolor{blue}{2.37$\pm$0.00}}}$^{\footnotesize{**}}$ & {\textBF{\textcolor{blue}{114.5$\pm$0.4}}}$^{\footnotesize{**}}$ & {26.6$\pm$0.1} & {\textBF{\textcolor{blue}{1.43$\pm$0.00}}}$^{\footnotesize{**}}$ & {0.10$\pm$0.00} & {0.30$\pm$0.00} \\
    \midrule
        Transformer (MSE) & {2.51$\pm$0.01} & {212.7$\pm$5.2} & {\textBF{\textcolor{blue}{24.7$\pm$0.2}}}$^{\footnotesize{**}}$ & {1.87$\pm$0.03} & {\textBF{\textcolor{blue}{0.16$\pm$0.01}}}$^{\footnotesize{**}}$ & {0.28$\pm$0.01} \\
        Transformer (MSLE) & {\textBF{\textcolor{blue}{2.36$\pm$0.00}}}$^{\footnotesize{**}}$ & {\textBF{\textcolor{blue}{114.1$\pm$0.6}}}$^{\footnotesize{**}}$ & {26.7$\pm$0.1} & {\textBF{\textcolor{blue}{1.43$\pm$0.00}}}$^{\footnotesize{**}}$ & {0.09$\pm$0.00} & {\textBF{\textcolor{blue}{0.30$\pm$0.00}}}$^{\footnotesize{**}}$ \\
    \midrule
        TPC (MSE) & {2.21$\pm$0.02} & {154.3$\pm$10.1} & {\textBF{\textcolor{blue}{21.6$\pm$0.2}}} & {1.80$\pm$0.10} & {\textBF{\textcolor{blue}{0.27$\pm$0.01}}} & {0.47$\pm$0.01} \\
        TPC (MSLE) & {\textBF{\textcolor{blue}{1.78$\pm$0.02}}}$^{\footnotesize{**}}$ & {\textBF{\textcolor{blue}{63.5$\pm$4.3}}}$^{\footnotesize{**}}$ & {\textBF{\textcolor{lightblue}{21.7$\pm$0.5}}} & {\textBF{\textcolor{blue}{0.70$\pm$0.03}}}$^{\footnotesize{**}}$ & {\textBF{\textcolor{blue}{0.27$\pm$0.02}}} & {\textBF{\textcolor{blue}{0.58$\pm$0.01}}}$^{\footnotesize{**}}$ \\
    \bottomrule
    \end{tabular}
\end{table*}

\begin{table*}[h]
  \caption{eICU multitask results. We compare the performance of each model on individual tasks (LoS or mortality prediction) to the multitask setting (both LoS and mortality). The results from Table~\ref{tab:results}a are repeated here for ease of comparison. Note that the `mean' and `median' models are only for LoS -- there is no equivalent model for mortality prediction.}
  \label{tab:eICUmultitaskresults}
    \centering
    \begin{tabular}{p{2.1cm}|p{1.75cm}p{1.75cm}|p{1.4cm}p{1.4cm}p{1.25cm}p{1.4cm}p{1.4cm}p{1.4cm}}
    \toprule
        & \multicolumn{2}{l|}{\textbf{In-Hospital Mortality}} & \multicolumn{6}{c}{\textbf{Length of Stay}} \\
        \textbf{Model} & \textbf{AUROC} & \textbf{AUPRC} & \textbf{MAD} & \textbf{MAPE} & \textbf{MSE} & \textbf{MSLE} & \boldmath{$R^2$} & \textbf{Kappa} \\
    \midrule
        Mean & -- & -- & {3.21} & {395.7} & {29.5} & {2.87} & {0.00} & {0.00} \\
        Median & -- & -- & {2.76} & {184.4} & {32.6} & {2.15} & \hspace{-0.32em}{-0.11} & {0.00} \\
    \midrule
        \multirow{3}{*}{LSTM} & \textBF{\textcolor{lightblue}{0.849$\pm$0.002}} & 0.407$\pm$0.012 & -- & -- & -- & - & - & - \\
        & -- & -- & \textBF{\textcolor{lightblue}{2.39$\pm$0.00}} & {118.2$\pm$1.1} & \textBF{\textcolor{blue}{26.9$\pm$0.1}}$^{\footnotesize{*}}$ & \textBF{\textcolor{blue}{1.47$\pm$0.01}} & \textBF{\textcolor{blue}{0.09$\pm$0.00}}$^{\footnotesize{*}}$ & \textBF{\textcolor{blue}{0.28$\pm$0.00}} \\
        & \textBF{\textcolor{blue}{0.852$\pm$0.003}} & \textBF{\textcolor{blue}{0.436$\pm$0.007}}$^{\footnotesize{**}}$ & \textBF{\textcolor{blue}{2.40$\pm$0.01}} & \textBF{\textcolor{blue}{116.5$\pm$0.8}}$^{\footnotesize{*}}$ & 27.2$\pm$0.2 & \textBF{\textcolor{blue}{1.47$\pm$0.01}} & 0.08$\pm$0.01 & \textBF{\textcolor{blue}{0.28$\pm$0.01}} \\
    \midrule
        \multirow{3}{*}{CW LSTM} & 0.855$\pm$0.001 & 0.464$\pm$0.004 & -- & -- & -- & -- & -- & -- \\
        & -- & -- & \textBF{\textcolor{blue}{2.37$\pm$0.00}} & \textBF{\textcolor{blue}{114.5$\pm$0.4}} & \textBF{\textcolor{blue}{26.6$\pm$0.1}}$^{\footnotesize{*}}$ & \textBF{\textcolor{blue}{1.43$\pm$0.00}}$^{\footnotesize{*}}$ & \textBF{\textcolor{blue}{0.10$\pm$0.00}}$^{\footnotesize{*}}$ & \textBF{\textcolor{blue}{0.30$\pm$0.00}} \\
        & \textBF{\textcolor{blue}{0.865$\pm$0.002}}$^{\footnotesize{**}}$ & \textBF{\textcolor{blue}{0.490$\pm$0.007}}$^{\footnotesize{**}}$ & \textBF{\textcolor{blue}{2.37$\pm$0.00}} & \textBF{\textcolor{lightblue}{115.0$\pm$0.7}} & 26.8$\pm$0.1 & 1.44$\pm$0.00 & 0.09$\pm$0.00 & \textBF{\textcolor{blue}{0.30$\pm$0.00}} \\
    \midrule
        \multirow{3}{*}{Transformer} & 0.851$\pm$0.002 & 0.454$\pm$0.005 & -- & -- & -- & -- & -- & -- \\
        & -- & -- & \textBF{\textcolor{blue}{2.36$\pm$0.00}} & \textBF{\textcolor{blue}{114.1$\pm$0.6}} & \textBF{\textcolor{lightblue}{26.7$\pm$0.1}} & \textBF{\textcolor{blue}{1.43$\pm$0.00}} & \textBF{\textcolor{lightblue}{0.09$\pm$0.00}} & \textBF{\textcolor{blue}{0.30$\pm$0.00}} \\
        & \textBF{\textcolor{blue}{0.858$\pm$0.001}}$^{\footnotesize{**}}$ & \textBF{\textcolor{blue}{0.475$\pm$0.004}}$^{\footnotesize{**}}$ & \textBF{\textcolor{blue}{2.36$\pm$0.00}} & \textBF{\textcolor{lightblue}{114.2$\pm$0.7}} & \textBF{\textcolor{blue}{26.6$\pm$0.1}} & \textBF{\textcolor{blue}{1.43$\pm$0.00}} & \textBF{\textcolor{blue}{0.10$\pm$0.00}} & \textBF{\textcolor{blue}{0.30$\pm$0.00}} \\
    \midrule
        \multirow{3}{*}{TPC} & \textBF{\textcolor{lightblue}{0.864$\pm$0.001}} & 0.508$\pm$0.005 & -- & -- & -- & -- & -- & -- \\
        & -- & -- & 1.78$\pm$0.02 & 63.5$\pm$3.8 & 21.8$\pm$0.5 & 0.71$\pm$0.03 & 0.26$\pm$0.02 & 0.58$\pm$0.01 \\
        & \textBF{\textcolor{blue}{0.865$\pm$0.002}} & \textBF{\textcolor{blue}{0.523$\pm$0.006}}$^{\footnotesize{**}}$ & \textBF{\textcolor{blue}{1.55$\pm$0.01}}$^{\footnotesize{**}}$ & \textBF{\textcolor{blue}{46.4$\pm$2.6}}$^{\footnotesize{**}}$ & \textBF{\textcolor{blue}{18.7$\pm$0.2}}$^{\footnotesize{**}}$ & \textBF{\textcolor{blue}{0.40$\pm$0.02}}$^{\footnotesize{**}}$ & \textBF{\textcolor{blue}{0.37$\pm$0.01}}$^{\footnotesize{**}}$ & \textBF{\textcolor{blue}{0.70$\pm$0.00}}$^{\footnotesize{**}}$ \\
    \bottomrule
    \end{tabular}
\end{table*}

\begin{table*}[h]
  \caption{MIMIC-IV multitask results.}
  \label{tab:MIMICmultitaskresults}
    \centering
    \begin{tabular}{p{2.1cm}|p{1.75cm}p{1.75cm}|p{1.4cm}p{1.4cm}p{1.25cm}p{1.4cm}p{1.4cm}p{1.4cm}}
    \toprule
        & \multicolumn{2}{l|}{\textbf{In-Hospital Mortality}} & \multicolumn{6}{c}{\textbf{Length of Stay}} \\
        \textbf{Model} & \textbf{AUROC} & \textbf{AUPRC} & \textbf{MAD} & \textbf{MAPE} & \textbf{MSE} & \textbf{MSLE} & \boldmath{$R^2$} & \textbf{Kappa} \\
    \midrule
        Mean & -- & -- & 5.24 & 474.9 & 77.7 & 2.80 & 0.00 & 0.00 \\
        Median & -- & -- & 4.60 & 216.8 & 86.8 & 2.09 & \hspace{-0.32em}{-0.12} & 0.00 \\
    \midrule
        \multirow{3}{*}{LSTM} & \textBF{\textcolor{lightblue}{0.895$\pm$0.001}} & \textBF{\textcolor{lightblue}{0.657$\pm$0.003}} & -- & -- & -- & -- & -- & -- \\
        & -- & -- &  \textBF{\textcolor{lightblue}{3.68$\pm$0.02}} & \textBF{\textcolor{lightblue}{107.2$\pm$3.1}} & \textBF{\textcolor{lightblue}{65.7$\pm$0.7}} & 1.26$\pm$0.01 & \textBF{\textcolor{lightblue}{0.15$\pm$0.01}} & \textBF{\textcolor{lightblue}{0.43$\pm$0.01}} \\
        & \textBF{\textcolor{blue}{0.896$\pm$0.002}} & \textBF{\textcolor{blue}{0.659$\pm$0.004}} &  \textBF{\textcolor{blue}{3.66$\pm$0.01}} &  \textBF{\textcolor{blue}{106.8$\pm$2.7}} &  \textBF{\textcolor{blue}{65.3$\pm$0.6}} &  \textBF{\textcolor{blue}{1.25$\pm$0.01}}$^{\footnotesize{*}}$ &  \textBF{\textcolor{blue}{0.16$\pm$0.01}} &  \textBF{\textcolor{blue}{0.44$\pm$0.00}} \\
    \midrule
        \multirow{3}{*}{CW LSTM} & \textBF{\textcolor{lightblue}{0.897$\pm$0.002}} & \textBF{\textcolor{lightblue}{0.650$\pm$0.005}} & -- & -- & -- & -- & -- & -- \\
        & -- & -- & \textBF{\textcolor{blue}{3.68$\pm$0.02}} & \textBF{\textcolor{blue}{107.0$\pm$1.8}} & \textBF{\textcolor{lightblue}{66.4$\pm$0.6}} & \textBF{\textcolor{blue}{1.23$\pm$0.01}} & \textBF{\textcolor{blue}{0.15$\pm$0.01}} & \textBF{\textcolor{lightblue}{0.43$\pm$0.00}} \\
        & \textBF{\textcolor{blue}{0.899$\pm$0.002}} & \textBF{\textcolor{blue}{0.654$\pm$0.003}} & \textBF{\textcolor{lightblue}{3.69$\pm$0.02}} & \textBF{\textcolor{lightblue}{107.2$\pm$1.6}} & \textBF{\textcolor{blue}{66.3$\pm$0.6}} & \textBF{\textcolor{blue}{1.23$\pm$0.01}} & \textBF{\textcolor{blue}{0.15$\pm$0.01}} & \textBF{\textcolor{blue}{0.44$\pm$0.00}} \\
    \midrule
        \multirow{3}{*}{Transformer} & 0.890$\pm$0.002 & 0.641$\pm$0.008 & -- & -- & -- & -- & -- & -- \\
        & -- & -- & \textBF{\textcolor{lightblue}{3.62$\pm$0.02}} & \textBF{\textcolor{lightblue}{113.8$\pm$1.8}} & \textBF{\textcolor{lightblue}{63.4$\pm$0.5}} & \textBF{\textcolor{lightblue}{1.21$\pm$0.01}} & \textBF{\textcolor{lightblue}{0.18$\pm$0.01}} & \textBF{\textcolor{blue}{0.45$\pm$0.00}} \\
        & \textBF{\textcolor{blue}{0.898$\pm$0.001}}$^{\footnotesize{**}}$ & \textBF{\textcolor{blue}{0.656$\pm$0.005}}$^{\footnotesize{*}}$ & \textBF{\textcolor{blue}{3.61$\pm$0.01}} & \textBF{\textcolor{blue}{112.3$\pm$2.0}} & \textBF{\textcolor{blue}{63.3$\pm$0.3}} & \textBF{\textcolor{blue}{1.20$\pm$0.01}} & \textBF{\textcolor{blue}{0.19$\pm$0.00}} & \textBF{\textcolor{blue}{0.45$\pm$0.00}} \\
    \midrule
        \multirow{3}{*}{TPC} & 0.905$\pm$0.001 & 0.691$\pm$0.006 & -- & -- & -- & -- & -- & -- \\
        & -- & -- & 2.39$\pm$0.03 & 47.6$\pm$1.4 & 46.3$\pm$1.3 & 0.39$\pm$0.02 & 0.40$\pm$0.02 & 0.78$\pm$0.01 \\
        & \textBF{\textcolor{blue}{0.918$\pm$0.002}}$^{\footnotesize{**}}$ & \textBF{\textcolor{blue}{0.713$\pm$0.007}}$^{\footnotesize{**}}$ & \textBF{\textcolor{blue}{2.28$\pm$0.07}}$^{\footnotesize{*}}$ & \textBF{\textcolor{blue}{32.4$\pm$1.2}}$^{\footnotesize{**}}$ & \textBF{\textcolor{blue}{42.0$\pm$1.2}}$^{\footnotesize{**}}$ & \textBF{\textcolor{blue}{0.19$\pm$0.00}}$^{\footnotesize{**}}$ & \textBF{\textcolor{blue}{0.46$\pm$0.02}}$^{\footnotesize{**}}$ & \textBF{\textcolor{blue}{0.85$\pm$0.00}}$^{\footnotesize{**}}$ \\
    \bottomrule
    \end{tabular}
\end{table*}

\begin{table*}[h]
    \caption{eICU time series features. `Time in the ICU' and `Time of day' were not part of the tables in eICU but were added later as helpful indicators to the model.}
    \label{tab:timeseries}
    \centering
    \begin{tabular}{llll}
        \toprule
        \multicolumn{4}{c}{\textbf{Source Table}} \\
        \multicolumn{3}{l}{\textbf{\textit{lab}}} & \textbf{\textit{respiratorycharting}} \\
        \midrule
        -basos & MPV & glucose & Exhaled MV \\
        -eos & O2 Sat (\%) & lactate & Exhaled TV (patient) \\
        -lymphs & PT & magnesium & LPM O2 \\
        -monos & PT - INR & pH & Mean Airway Pressure \\
        -polys & PTT & paCO2 & Peak Insp. Pressure \\
        ALT (SGPT) & RBC & paO2 & PEEP \\
        AST (SGOT) & RDW & phosphate & Plateau Pressure \\
        BUN & WBC x 1000 & platelets x 1000 & Pressure Support \\
        Base Excess & albumin & potassium & RR (patient) \\
        FiO2 & alkaline phos. & sodium & SaO2 \\
        HCO3 & anion gap & total bilirubin & TV/kg IBW \\
        Hct & bedside glucose & total protein & Tidal Volume (set) \\
        Hgb & bicarbonate & troponin - I & Total RR \\
        MCH & calcium & urinary specific gravity & Vent Rate \\
        MCHC & chloride & & \\
        MCV & creatinine & & \\
        \vspace{-0.8em}\\
        \toprule
        \textbf{\textit{nursecharting}} &  \textbf{\textit{vitalperiodic}} & \textbf{\textit{vitalaperiodic}} & \textbf{N/A} \\
        \midrule
        Bedside Glucose & cvp & noninvasivediastolic & Time in the ICU \\
        Delirium Scale/Score & heartrate & noninvasivemean & Time of day \\
        Glasgow coma score & respiration & noninvasivesystolic & \\
        Heart Rate & sao2 & & \\
        Invasive BP & st1 & & \\
        Non-Invasive BP & st2 & & \\
        O2 Admin Device & st3 & & \\
        O2 L/\% & systemicdiastolic & & \\
        O2 Saturation & systemicmean & & \\
        Pain Score/Goal & systemicsystolic & & \\
        Respiratory Rate & temperature & & \\
        Sedation Score/Goal & & & \\
        Temperature & & & \\
        \bottomrule
    \end{tabular}
\end{table*}

\begin{table*}[h]
    \caption{MIMIC-IV time series features.}
    \label{tab:timeseriesMIMIC}
    \centering
    \begin{tabular}{lll}
        \toprule
        \multicolumn{3}{c}{\textbf{Source Table}} \\
        \multicolumn{3}{l}{\textbf{\textit{chartevents}}} \\
        \midrule
        Activity / Mobility (JH-HLM) & Mean Airway Pressure & Resp Alarm - High \\
        Apnea Interval & Minute Volume & Resp Alarm - Low \\
        Arterial Blood Pressure Alarm - High & Minute Volume Alarm - High & Respiratory Rate \\
        Arterial Blood Pressure Alarm - Low & Minute Volume Alarm - Low & Respiratory Rate (Set) \\
        Arterial Blood Pressure diastolic & Non Invasive Blood Pressure diastolic & Respiratory Rate (Total) \\
        Arterial Blood Pressure mean & Non Invasive Blood Pressure mean & Respiratory Rate (spontaneous) \\
        Arterial Blood Pressure systolic & Non Invasive Blood Pressure systolic & Richmond-RAS Scale \\
        Braden Score & Non-Invasive Blood Pressure Alarm - High & Strength L Arm \\
        Current Dyspnea Assessment & Non-Invasive Blood Pressure Alarm - Low & Strength L Leg \\
        Daily Weight & O2 Flow & Strength R Arm \\
        Expiratory Ratio & O2 Saturation Pulseoxymetry Alarm - Low & Strength R Leg \\
        Fspn High & O2 saturation pulseoxymetry & Temperature Fahrenheit \\
        GCS - Eye Opening & PEEP set & Tidal Volume (observed) \\
        GCS - Motor Response & PSV Level & Tidal Volume (set) \\
        GCS - Verbal Response & Pain Level & Tidal Volume (spontaneous) \\
        Glucose finger stick (range 70-100) & Pain Level Response & Total PEEP Level \\
        Heart Rate & Paw High & Ventilator Mode \\
        Heart Rate Alarm - Low & Peak Insp. Pressure & Vti High \\
        Heart rate Alarm - High & Phosphorous &  \\
        Inspired O2 Fraction & Plateau Pressure &  \\
        \vspace{-0.8em}\\
        \toprule
        \multicolumn{2}{l}{\textbf{\textit{labevents}}} & \textbf{N/A} \\
        \midrule
        Alanine Aminotransferase (ALT) & MCHC & Time in the ICU \\
        Alkaline Phosphatase & MCV & Time of day \\
        Anion Gap & Magnesium & \\
        Asparate Aminotransferase (AST) & Oxygen Saturation & \\
        Base Excess & PT & \\
        Bicarbonate & PTT & \\
        Bilirubin, Total & Phosphate &  \\
        Calcium, Total & Platelet Count & \\
        Calculated Total CO2 & Potassium & \\
        Chloride & Potassium, Whole Blood & \\
        Creatinine & RDW & \\
        Free Calcium & RDW-SD & \\
        Glucose & Red Blood Cells & \\
        H & Sodium & \\
        Hematocrit & Sodium, Whole Blood & \\
        Hematocrit, Calculated & Temperature & \\
        Hemoglobin & Urea Nitrogen & \\
        I & White Blood Cells & \\
        INR(PT) & pCO2 & \\
        L & pH & \\
        Lactate & pO2 & \\
        MCH & & \\
        \bottomrule
    \end{tabular}
\end{table*}

\end{document}